\documentclass[10pt,journal,compsoc]{IEEEtran}
\usepackage{cite}
\usepackage{amsmath}
\usepackage{epsfig}
\usepackage{graphicx}
\usepackage{array}
\usepackage{multirow}
\usepackage{amssymb}
\usepackage{extarrows}
\usepackage{booktabs}       
\usepackage{diagbox}
\usepackage{comment}

\usepackage{times}
\usepackage{wrapfig}
\usepackage{caption}
\usepackage{subcaption}
\usepackage{verbatim}
\usepackage{color}
\usepackage{cite}
\usepackage{pifont}

\usepackage{arydshln}

\usepackage{url}

\usepackage{mathtools}
\DeclarePairedDelimiter{\floor}{\lfloor}{\rfloor}
\DeclarePairedDelimiter{\ceil}{\lceil}{\rceil}
\DeclareMathOperator*{\argmin}{arg\,min}

\newcommand{\norm}[1]{\left\lVert#1\right\rVert}

\newcommand\etal{\textit{et~al.}}

\newcommand{\eg}{\textit{e.g.}}

\newcommand{\Paragraph}[1]{\vspace{1mm} \noindent \textbf{#1.} \hspace{1mm}}

\newcommand{\Section}[1]{\vspace{0mm} \section{#1} \vspace{0mm}}
\newcommand{\SubSection}[1]{\vspace{0mm} \subsection{#1} \vspace{0mm}}
\newcommand{\SubSubSection}[1]{\vspace{0mm} \subsubsection{#1} \vspace{0mm}}

\def\eqnvspace{{\vspace{0mm}}}
\def\figvspace{{\vspace{0mm}}}
\def\tabvspace{{\vspace{-2mm}}}

\def\MeaningfulBasisFigWid{0.115\textwidth}
\def\FittingFigWid{0.095\textwidth}
\def\ShapeRepFigWid{0.089\textwidth}
\def\TexRepFigWid{0.11\textwidth}

\def\AlignFigWid{0.092\textwidth}
\def\AblTexFigHeight{1.25cm}
\def\VaryingShapeFigWid{0.12\textwidth}
\def\VaryingAlbFigWid{0.12\textwidth}
\def\ReconFigWid{0.092\textwidth}
\def\JsReconFigWid{0.075\textwidth}
\def\SelaReconFigWid{0.095\textwidth}

\def\EditingFigWid{0.09\textwidth}
\def\EditingLightFigWid{0.09\textwidth}

\def\AEFigWid{0.08\textwidth}

\begin{document}
%
\title{On Learning 3D Face Morphable Model\\from In-the-wild Images}

\author{Luan~Tran,
        and~Xiaoming~Liu,~\IEEEmembership{Member,~IEEE}
\IEEEcompsocitemizethanks{\IEEEcompsocthanksitem L. Tran and X. Liu are with the Department of Computer Science and Engineering, Michigan State University.\protect\\E-mail: tranluan@msu.edu, liuxm@cse.msu.edu 
}%
\thanks{Manuscript received Aug 28, 2018.}}

\markboth{}{}
\IEEEtitleabstractindextext{%

\begin{abstract}
As a classic statistical model of $3$D facial shape and albedo, $3$D Morphable Model ($3$DMM) is widely used in facial analysis, e.g., model fitting, image synthesis.
Conventional $3$DMM is learned from a set of $3$D face scans with associated well-controlled $2$D face images, and represented by two sets of PCA basis functions.
Due to the type and amount of training data, as well as, the linear bases, the representation power of $3$DMM can be limited.
To address these problems, this paper proposes an innovative framework to learn a nonlinear $3$DMM model from a large set of in-the-wild face images, without collecting $3$D face scans. 
Specifically, given a face image as input, a network encoder estimates the projection, lighting, shape and albedo parameters.
Two decoders serve as the nonlinear $3$DMM to map from the shape and albedo parameters to the $3$D shape and albedo, respectively.
With the projection parameter, lighting, $3$D shape, and albedo, a novel analytically-differentiable rendering layer is designed to reconstruct the original input face.
The entire network is end-to-end trainable with only weak supervision. 
We demonstrate the superior representation power of our nonlinear $3$DMM over its linear counterpart, and its contribution to face alignment, $3$D reconstruction, and face editing.

Source code and additional results can be found at our project page:~\url{http://cvlab.cse.msu.edu/project-nonlinear-3dmm.html}
\end{abstract}

}

\maketitle

\IEEEdisplaynontitleabstractindextext

%
\IEEEpeerreviewmaketitle

\IEEEraisesectionheading{\section{Introduction}
\label{sec:intro}}


\IEEEPARstart{T}{he} $3$D Morphable Model ($3$DMM) is a statistical model of $3$D facial shape and texture in a space where there are explicit correspondences~\cite{blanz1999morphable}. 
The morphable model framework provides two key benefits: first, a point-to-point correspondence between the reconstruction and all other models, enabling “morphing”,
and second, modeling underlying transformations between types of faces (male to female, neutral to smile, etc.). 
$3$DMM has been widely applied in numerous areas including, but not limited to, computer vision~\cite{blanz1999morphable, yu2017learning, tran2017extreme}, computer graphics~\cite{aldrian2013inverse, shi2014automatic, thies2015real, thies2016face2face}, human behavioral analysis~\cite{amberg2008expression, yin2017towards} and craniofacial surgery~\cite{staal2015describing}.

Traditionally, $3$DMM is learnt through {\it supervision} by performing dimension reduction, typically Principal Component Analysis (PCA), on a training set of co-captured $3$D face scans and $2$D images.
To model highly variable $3$D face shapes, a large amount of high-quality $3$D face scans is required. 
However, this requirement is expensive to fulfill as acquiring face scans is very laborious, in both data capturing and post-processing stage.
The first $3$DMM~\cite{blanz1999morphable} was built from scans of $200$ subjects with a similar ethnicity/age group. 
They were also captured in well-controlled conditions, with only neutral expressions. 
Hence, it is fragile to large variances in the face identity. 
The widely used Basel Face Model~(BFM)~\cite{paysan20093d} is also built with only $200$ subjects in neutral expressions.  
Lack of expression can be compensated using expression bases from FaceWarehouse~\cite{cao2014facewarehouse} or BD-$3$FE~\cite{yin20063d}, which are learned from the offsets to the neutral pose. 
After more than a decade, almost all existing models use no more than $300$ training scans. 
Such small training sets are far from adequate to describe the full variability of human faces~\cite{booth20163d}. 
Until recently, with a significant effort as well as a novel automated and robust model construction pipeline, Booth~\etal~\cite{booth20163d} build the first large-scale $3$DMM from scans of ${\sim}10,000$ subjects. 

\begin{figure}[t!]
\centering
\includegraphics[trim=30 0 12 0, clip, width=\linewidth]{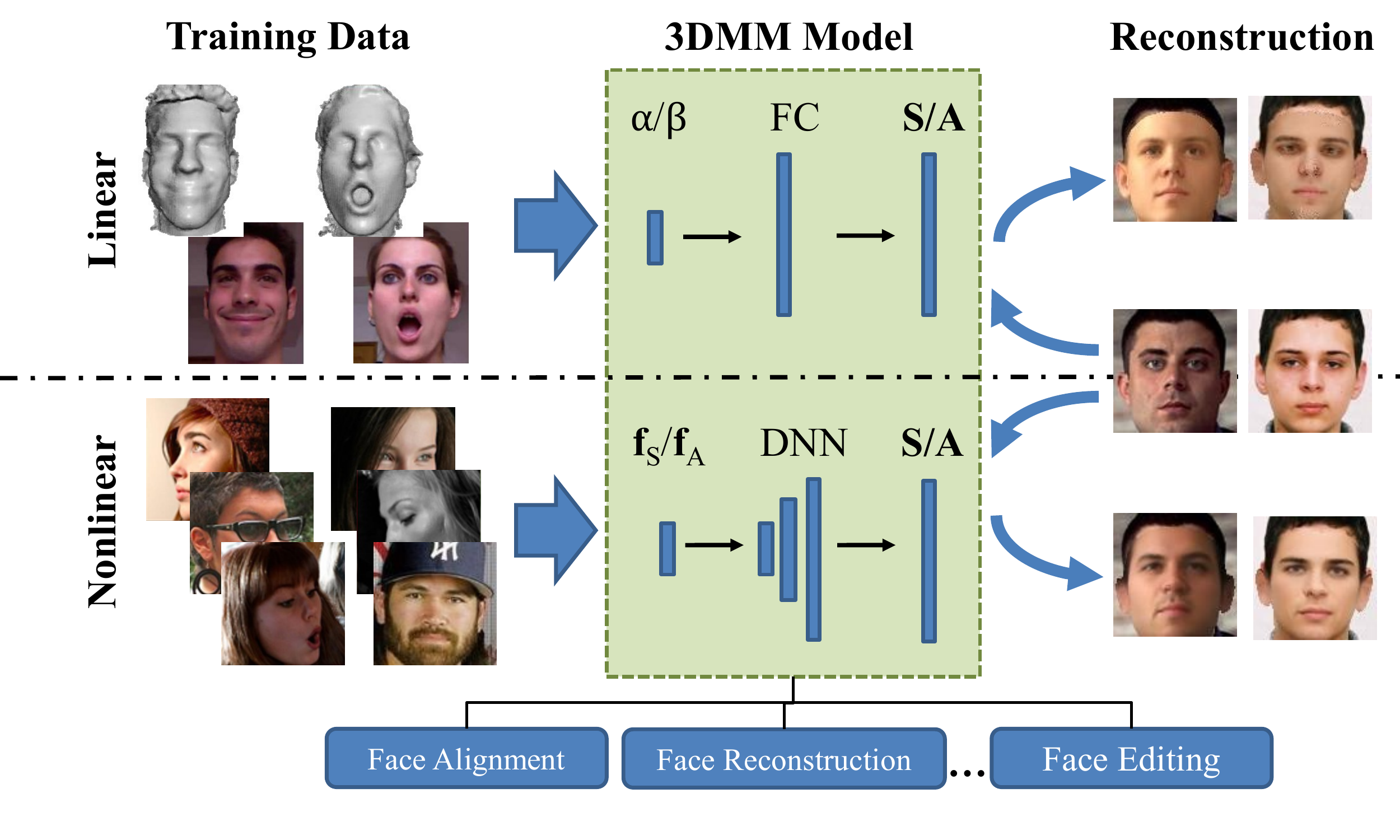}
\vspace{-4mm}
\caption{\small Conventional $3$DMM employs linear bases models for shape/albedo, which are trained with $3$D face scans and associated controlled $2$D images. We propose a nonlinear $3$DMM to model shape/albedo via deep neural networks~(DNNs). It can be trained from in-the-wild face images without $3$D scans, and also better reconstruct the original images due to the inherent nonlinearity.}
\label{fig:concept}
\figvspace 
\end{figure}

Second, the texture model of $3$DMM is normally built with a small number of $2$D face images {\it co-captured} with $3$D scans, under well-controlled conditions.
Despite there is a considerable improvement of $3$D acquisition devices in the last few years, these devices still cannot operate in arbitrary in-the-wild conditions. Therefore, all the current $3$D facial datasets have been captured in the laboratory environment.
Hence, such models are only learnt to represent the facial texture in similar, rather than in-the-wild, conditions.
This substantially limits its application scenarios. 


Finally, the representation power of $3$DMM is limited by not only the size or type of training data but also its {\it formulation}.
The facial variations are nonlinear in nature.
E.g., the variations in different facial expressions or poses are nonlinear, which violates the linear assumption of PCA-based models. 
Thus, a PCA model is unable to interpret facial variations sufficiently well.
This is especially true for facial texture. For all current $3$DMM models, their low-dimension albedo subspace faces the same problem of  lacking facial hair, e.g., beards. 
To reduce the fitting error, it compensates unexplainable texture by alternating surface normal, or shrinking the face shape~\cite{zollhoefer2018facestar}. 
Either way, linear $3$DMM-based applications often degrade their performances when handling out-of-subspace variations.

Given the barrier of $3$DMM in its data, supervision and linear bases, this paper aims to revolutionize the paradigm of learning $3$DMM by answering a fundamental question:

\vspace{0.7mm}
\begin{quote}
{\it Whether and how can we learn a nonlinear $3$D Morphable Model of face shape and albedo from a set of in-the-wild  $2$D face images, without collecting $3$D face scans?}
\end{quote}
\vspace{0.7mm}

If the answer were yes, this would be in sharp contrast to the conventional $3$DMM approach, and remedy all aforementioned limitations.
Fortunately, we have developed approaches to offer positive answers to this question.
With the recent development of deep neural networks, we view that it is the right time to undertake this new paradigm of $3$DMM learning.
Therefore, the core of this paper is regarding how to learn this new $3$DMM, what is the representation power of the model, and what is the benefit of the model to facial analysis.

We propose a novel paradigm to {\it learn a  nonlinear $3$DMM model from a large in-the-wild  $2$D face image collection, without acquiring $3$D face scans}, by leveraging the power of deep neural networks captures variations and structures in complex face data. 
As shown in Fig.~\ref{fig:concept}, starting with an observation that the linear $3$DMM formulation is equivalent to a single layer network, using a deep network architecture naturally increases the model capacity. 
Hence, we utilize two convolution neural network decoders, instead of two PCA spaces, as the shape and albedo model components, respectively.
Each decoder will take a shape or albedo parameter as input and output the dense $3$D face mesh or a face skin reflectant.
These two decoders are essentially the nonlinear $3$DMM.

Further, we learn the fitting algorithm to our nonlinear $3$DMM, which is formulated as a CNN encoder.
The encoder network takes a face image as input and generates the shape and albedo parameters, from which two decoders estimate shape and albedo.

The $3$D face and albedo would {\it perfectly} reconstruct the input face, if the fitting algorithm and $3$DMM are well learnt.
Therefore, we design a differentiable rendering layer to generate a reconstructed face by fusing the $3$D face, albedo, lighting, and the camera projection parameters estimated by the encoder. 
Finally, the end-to-end  learning scheme is constructed where the encoder and two decoders are learnt jointly to minimize the difference between the reconstructed face and the input face.
Jointly learning the $3$DMM and the model fitting encoder allows us to leverage the large collection of {\it in-the-wild} $2$D images without relying on $3$D scans.
We show significantly improved shape and facial texture representation power over the linear $3$DMM. 
Consequently, this also benefits other tasks such as $2$D face alignment, $3$D reconstruction, and face editing.

A preliminary version of this work was published in 2018 IEEE Conference on Computer Vision and Pattern Recognition~\cite{tran2018nonlinear}. We extend it in numerous ways:
1) Instead of having lighting embedded in texture, we split texture into albedo and shading. Truthfully modeling the lighting help to improve the shape modeling as it can help to guide the surface normal learning. This results in better performance in followed tasks: alignment and reconstruction, as demonstrated in our experiment section.
2) We propose to present the shape component in the $2$D UV space, which helps to reserve spatial relation among its vertices. This also allows us to use a CNN, rather than an expensive multi-layer perceptron, as the shape decoder.
3) To ensure plausible reconstruction, we employ multiple constraints to regularize the model learning.

In summary, this paper makes the following contributions:
\begin{itemize}
\item We learn a \textit{nonlinear} $3$DMM model, fully models shape, albedo and lighting, that has greater representation power than its traditional linear counterpart.

\item Both shape and albedo are represented  as $2$D images, which help to maintain spatial relations as well as leverage CNN power in image synthesis.

\item We jointly learn the model and the model fitting algorithm via \textit{weak supervision}, by leveraging a large collection of $2$D images without $3$D scans. 
The novel rendering layer enables the end-to-end training.

\item The new $3$DMM further improves performance in related tasks: face alignment, face reconstruction and face editing. 
\end{itemize}

\Section{Prior Work}
\label{sec:prior}

\begin{figure*}[t!]
\centering
\includegraphics[trim=0 0 0 0,clip, width=0.99\linewidth]{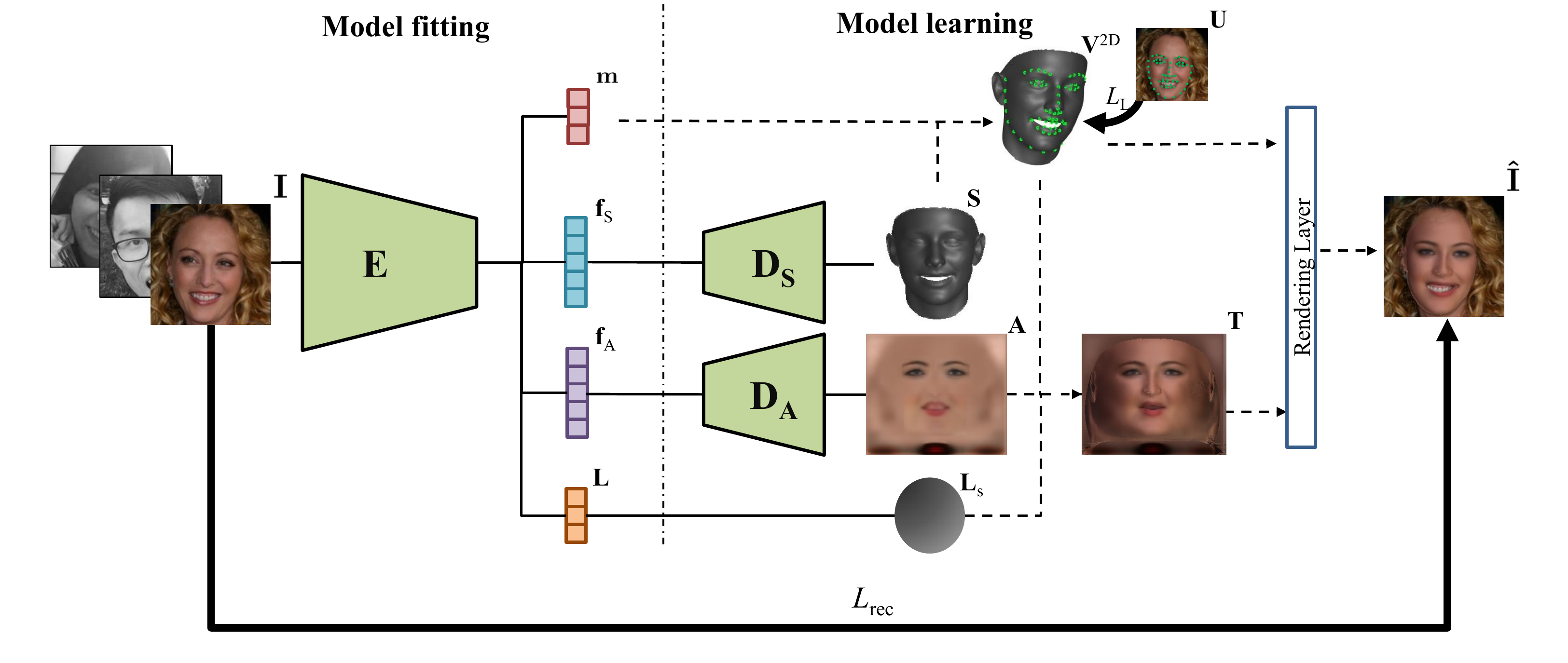}
\vspace{-3mm}
\caption{\small Jointly learning a nonlinear $3$DMM and its fitting algorithm from unconstrained $2$D in-the-wild face image collection, in a weakly supervised fashion. $\mathbf{L}_S$ is a visualization of shading on a sphere with lighting parameters $\mathbf{L}$. }
\label{fig:architecture}
\figvspace 
\end{figure*}

\Paragraph{Linear $3$DMM}
Blanz and Vetter~\cite{blanz1999morphable} propose the first generic $3$D face model learned from scan data. They define a linear subspace to represent shape and texture using principal component analysis (PCA) and show how to fit the model to data.
Since this seminal work, there has been a large amount of effort on improving $3$DMM modeling mechanism.
In~\cite{blanz1999morphable}, the dense correspondence between facial mesh is solved with a regularised form of optical flow. However, this technique is only effective in a constrained setting, where subjects share similar ethnicities and ages.
To overcome this challenge, Patel and Smith~\cite{patel20093d} employ a Thin Plate Splines (TPS) warp~\cite{bookstein1989principal} to register the meshes into a common reference frame. 
Alternatively, Paysan~\etal~\cite{paysan20093d} use a Nonrigid Iterative Closest Point~(ICP)~\cite{amberg2007optimal} to directly align $3$D scans.
In a different direction, Amberg~\etal~\cite{amberg2008expression} extended Blanz and Vetter's PCA-based model to emotive facial shapes by adopting an additional PCA modeling of the residuals from the neutral pose. This results in a single
linear model of both identity and expression variation of $3$D facial shape.
Vlasic~\etal~\cite{vlasic2005face} use a multilinear model to represent the combined effect of identity and expression variation on the facial shape.
Later, Bolkart and Wuhrer~\cite{bolkart2015groupwise} show how such a multilinear model can be estimated directly from the $3$D scans using a joint optimization over the model parameters and groupwise registration of $3$D scans.

\Paragraph{Improving Linear $3$DMM}
With PCA bases, the statistical distribution underlying $3$DMM is Gaussian. Koppen~\etal~\cite{koppen2017gaussian} argue that single-mode Gaussian can't well represent real-world distribution. They introduce the Gaussian Mixture $3$DMM that models the global population as a mixture of Gaussian subpopulations, each with its own mean, but shared covariance.
Booth~\etal~\cite{booth20173d, booth20183d} aim to improve texture of $3$DMM to go beyond controlled settings by learning “in-the-wild” feature-based texture model.
On another direction, Tran ~\etal\cite{tran2017regressing} learn to regress robust and discriminative $3$DMM representation, by leveraging multiple images from the same subject.
However, all works are still based on statistical PCA bases.
Duong ~\etal~\cite{nhan2015beyond} address the problem of linearity in face modeling by using Deep Boltzmann Machines. However, they only work with $2$D face and sparse landmarks; and hence cannot handle faces with large-pose variations or occlusion well.
Concurrent to our work, Tewari ~\etal~\cite{tewari2018self} learn a (potentially non-linear) corrective model on top of a linear model. The final model is a summation of the base linear model and the learned corrective model, which contrasts to our unified model. Furthermore, our model has an advantage of using $2$D representation of both shape and albedo, which maintains spatial relations between vertices and leverages CNN power for image synthesis. Finally, thanks for our novel rendering layer, we are able to employ perceptual, adversarial loss to improve the reconstruction quality.

\Paragraph{$2$D Face Alignment}
$2$D Face Alignment~\cite{wu2008face,liu2009discriminative} can be cast as a regression problem where $2$D landmark locations are regressed directly~\cite{dollar2010cascaded}. 
For large-pose or occluded faces, strong priors of $3$DMM face shape have been shown to be beneficial~\cite{jourabloo2015pose}. 
Hence, there is increasing attention in conducting face alignment by fitting a $3$D face model to a single $2$D image~\cite{jourabloo2016large, zhu2016face, zhu2017face, liu2016joint, mcdonagh2016joint,  jourabloo2017pose, jourabloo2017poseinvariant}. 
Among the prior works, iterative approaches with cascade of regressors tend to be preferred. 
At each cascade, there is a single~\cite{tulyakov2015regressing, jourabloo2015pose} or even two regressors~\cite{wu2015robust} used to improve its prediction. 
Recently, Jourabloo and Liu~\cite{jourabloo2017pose} propose a CNN architecture that enables the end-to-end training ability of their network cascade. 
Contrasted to aforementioned works that use a fixed $3$DMM model, our model and model fitting are learned jointly. 
This results in a more powerful model: a single-pass encoder, which is learned jointly with the model, achieves state-of-the-art face alignment performance on AFLW$2000$~\cite{zhu2016face} benchmark dataset.

\Paragraph{$3$D Face Reconstruction}
Face reconstruction creates a $3$D face model from an image collection~\cite{roth2015unconstrained, roth2017adaptive} or even with a single image~\cite{richardson20163d, sela2017unrestricted}. This long-standing problem draws a lot of interest because of its wide applications.
$3$DMM also demonstrates its strength in face reconstruction, especially in the monocular case. This problem is a highly under-constrained, as with a single image, present information about the surface is limited. Hence, $3$D face reconstruction must rely on prior knowledge like $3$DMM~\cite{roth2017adaptive}. Statistical PCA linear $3$DMM is the most commonly used approach. Besides $3$DMM fitting methods~\cite{blanz2003face, gu2008generative, zhang2006face,dou2017end, tewari2017mofa, liu2018disentangling}, recently, Richardson~\etal~\cite{richardson2017learning} design a refinement network that adds facial details on top of the $3$DMM-based geometry.
However, this approach can only learn $2.5$D depth map, which loses the correspondence property of $3$DMM. 
The follow up work by Sela~\etal\cite{sela2017unrestricted} try to overcome this weakness by learning a correspondence map. 
Despite having some impressive reconstruction results, both these methods are limited by training data synthesized from the linear $3$DMM model. Hence, they  fail to handle out-of-subspace variations, e.g., facial hair.

\Paragraph{Unsupervised learning in $3$DMM}
Collecting large-scale $3$D scans with detailed labels for learning $3$DMM is not an easy task. 
A few work try to use large-scale synthetic data as in~\cite{richardson20163d,kim2017inversefacenet}, but they don't generalize well as there still be a domain gap with real images.
Tewari~\etal~\cite{tewari2017mofa} is among the first work attempting to learn $3$DMM fitting from unlabeled images. 
They use an unsupervised loss which compares projected textured face mesh with the original image itself. The sparse landmark alignment is also used as an auxiliary loss.
Genova~\etal~\cite{genova2018unsupervised} further improve this approach by comparing reconstructed images and original input using higher-level features from a pretrained face recognition network. 
Compared to these work, our work has a different objective of learning a {\it nonlinear} $3$DMM.

\section{The Proposed Nonlinear $3$DMM}
\label{sec:alg}

In this section, we start by introducing the traditional linear $3$DMM and then present our novel nonlinear $3$DMM model.

\SubSection{Conventional Linear $3$DMM}
The $3$D Morphable Model ($3$DMM)~\cite{blanz1999morphable} and its $2$D counterpart, Active Appearance Model~\cite{cootes2001active,liu2006face,liu2010video}, provide parametric models for synthesizing faces, where faces are modeled using two components: shape and albedo (skin reflectant). 
In~\cite{blanz1999morphable}, Blanz~\etal~propose to describe the $3$D face space with PCA:
\begin{equation}
\mathbf{S} = \mathbf{\bar{S}} +  \mathbf{G} \mathbf{\alpha},
\end{equation}
where $\mathbf{S}\in\mathbb{R}^{3Q}$ is a $3$D face mesh with $Q$ vertices, $\mathbf{\bar{S}}\in\mathbb{R}^{3Q}$ is the mean shape, $\alpha\in\mathbb{R}^{l_S}$ is the shape parameter corresponding to a $3$D shape bases $\mathbf{G}$.  
The shape bases can be further split into $\mathbf{G} = [\mathbf{G}_{id}, \mathbf{G}_{exp}]$, where $\mathbf{G}_{id}$ is trained from $3$D scans with neutral expression, and $\mathbf{G}_{exp}$ is from the offsets between expression and neutral scans.

The albedo of the face $\mathbf{A}\in\mathbb{R}^{3Q}$ is defined within the mean shape $\mathbf{\bar{S}}$, which describes the R, G, B colors of $Q$  corresponding vertices.
$\mathbf{A}$ is also formulated as a linear combination of basis functions:
\begin{equation}
\mathbf{A} = \mathbf{\bar{A}} + \mathbf{R} \mathbf{\beta},
\eqnvspace
\end{equation}
where $\mathbf{\bar{A}}$ is the mean albedo, $\mathbf{R}$ is the albedo bases, and $\mathbf{\beta}\in\mathbb{R}^{l_T}$ is the albedo parameter.

The $3$DMM can be used to synthesize novel views of the face. 
Firstly, a $3$D face is projected onto the image plane with the weak perspective projection model:
\begin{equation}
\mathbf{V} = \mathbf{R}\ast\mathbf{S},
\label{eqn:rotation}
\end{equation}
\begin{equation}
g(\mathbf{S}, \mathbf{m})=\mathbf{V}^{\text{2D}} = f \ast \mathbf{Pr}\ast\mathbf{V}+\mathbf{t}_{2d} =  M(\mathbf{m}) \ast \begin{bmatrix} \mathbf{S} \\ \mathbf{1} \end{bmatrix},
\label{eqn:projection}
\end{equation}
where $g(\mathbf{S}, \mathbf{m})$ is the projection function leading to the $2$D positions $\mathbf{V}^{\text{2D}}$ of $3$D rotated vertices $\mathbf{V}$, $f$ is the scale factor, 
$\mathbf{Pr} = \begin{bmatrix} 1 & 0 & 0 \\ 0 & 1 & 0 \end{bmatrix} $ is the orthographic projection matrix,
$\mathbf{R}$ is the rotation matrix constructed from three rotation angles (pitch, yaw, roll), and $\mathbf{t}_{2d}$ is the translation vector. 
While the project matrix $M$ is of the size of $2 \times 4$, it has six degrees of freedom, which is parameterized by a $6$-dim vector $\mathbf{m}$. 
Then, the $2$D image is rendered using texture and an illumination model such as Phong reflection model~\cite{phong1975illumination} or Spherical Harmonics~\cite{ramamoorthi2001efficient}.

\begin{figure}[t!]
\centering
\includegraphics[trim=0 0 0 5,clip, width=0.85\linewidth]{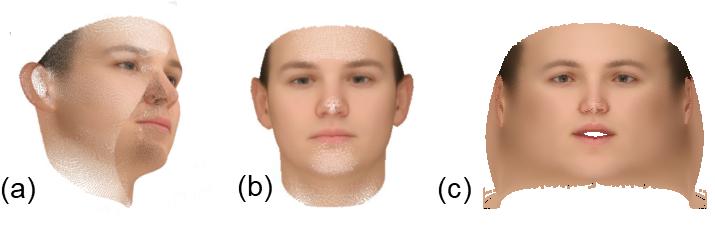}
\caption{\small Three albedo representations. (a) Albedo value per vertex, (b) Albedo as a $2$D frontal face, (c) UV space $2$D unwarped albedo.}
\label{fig:tex_representation}
\end{figure}

\SubSection{Nonlinear $3$DMM}


As mentioned in Sec.~\ref{sec:intro}, the linear $3$DMM has the problems such as requiring $3$D face scans for supervised learning, unable to leverage massive in-the-wild face images for learning, and the limited representation power due to the linear bases.
We propose to learn a nonlinear $3$DMM model using only large-scale in-the-wild $2$D face images.

\SubSubSection{Problem Formulation}

In linear $3$DMM, the factorization of each of components (shape, albedo)  can be seen as a matrix multiplication between coefficients and bases. 
From a neural network's perspective, this can be viewed as a shallow network with only {\it one fully connected layer} and no activation function. 
Naturally, to increase the model's representation power, the shallow network can be extended to a deep architecture. 
In this work, we design a novel learning scheme to joint learn a deep $3$DMM model and its inference (or fitting) algorithm. 

Specifically, as shown in Fig.~\ref{fig:architecture}, we use two deep networks to decode the shape, albedo parameters into the $3$D facial shape and albedo respectively. 
To make the framework end-to-end trainable, these parameters are estimated by an encoder network, which is essentially the fitting algorithm of our $3$DMM.
Three deep networks join forces for the ultimate goal of reconstructing the input face image, with the assistant of a physically-based rendering layer. 
Fig.~\ref{fig:architecture} visualizes the architecture of the proposed framework. Each component will be present in following sections.

\begin{figure}[t!]
\begin{center}
\small
\begin{tabular}{ c@{\hskip 1mm}c@{\hskip 1mm}c@{\hskip 1mm}c@{\hskip 2mm}c@{\hskip 2mm}}
\includegraphics[width=\AlignFigWid]{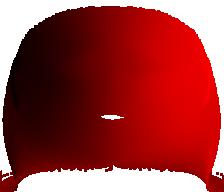} &
\includegraphics[width=\AlignFigWid]{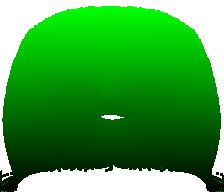} &
\includegraphics[width=\AlignFigWid]{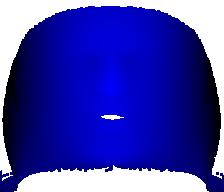} &
\includegraphics[width=\AlignFigWid]{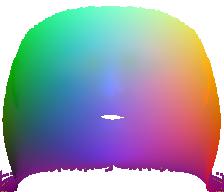}
\\
\end{tabular}
\caption{\small UV space shape representation. From left to right: individual channels for $x$, $y$ and $z$ spatial dimension and final combined shape image.  }
\label{fig:shape_representationUV}\figvspace 
\end{center}
\end{figure}

Formally, given a set of $K$ $2$D face images $\{\mathbf{I}_i \}_{i=1}^K$, we aim to learn an encoder $E$: $\mathbf{I}{\rightarrow}\mathbf{m}, \mathbf{L},\mathbf{f}_S, \mathbf{f}_A$ that estimates the projection  $\mathbf{m}$, lighting parameter $\mathbf{L}$,  shape parameters $\mathbf{f}_S\in\mathbb{R}^{l_S}$, and albedo parameter $\mathbf{f}_A\in\mathbb{R}^{l_A}$, a $3$D shape decoder $D_S$: $\mathbf{f}_S{\rightarrow} \mathbf{S}$ that decodes the shape parameter to a $3$D shape $\mathbf{S}\in\mathbb{R}^{3Q}$, and an albedo decoder $D_A$: $\mathbf{f}_A{\rightarrow} \mathbf{A}$ that decodes the albedo parameter to a realistic albedo $\mathbf{A}\in\mathbb{R}^{3Q}$, with the objective that the rendered image with $\mathbf{m}$, $\mathbf{L}$, $\mathbf{S}$, and $\mathbf{A}$ can well approximate the original image.
Mathematically, the objective function is:
\begin{gather}
\argmin_{E,D_S, D_A} \sum_{i=1}^K \norm{\hat{\mathbf{I}}_i -  \mathbf{I}_i}_1, \\
\hat{\mathbf{I}} = \mathcal{R} \left( E_m(\mathbf{I}), E_L(\mathbf{I}), D_S(E_S(\mathbf{I})), D_A(E_A(\mathbf{I})) \right), \nonumber
\end{gather}
where $\mathcal{R}(\mathbf{m},\mathbf{L}, \mathbf{S},\mathbf{A})$ is the rendering layer (Sec.~\ref{sec:rendering}).

\begin{figure*}[t!]
\centering
\includegraphics[trim=5 5 5 0,clip, width=0.85\linewidth]{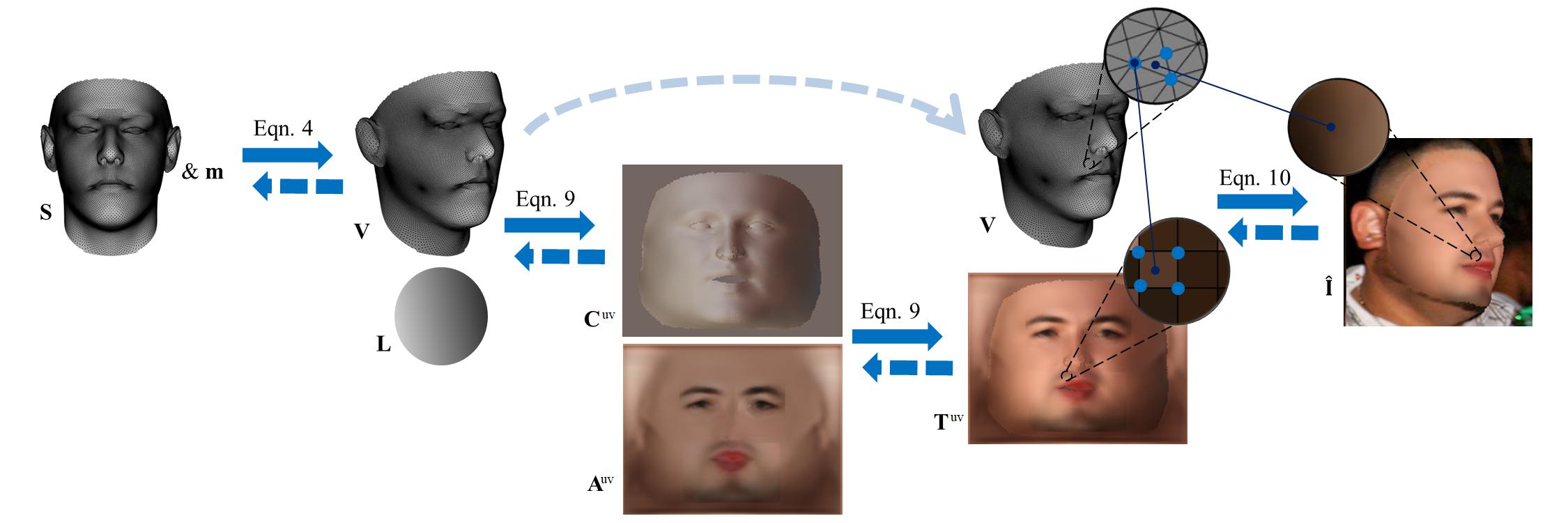}
\caption{\small Forward and backward pass of the rendering layer.}
\label{fig:render}
\figvspace 
\end{figure*}

\SubSubSection{Albedo  \&  Shape Representation}
\label{sec:shape_alb_representation}
Fig.~\ref{fig:tex_representation} illustrates three possible albedo representations. 
In traditional $3$DMM, albedo is defined per vertex (Fig.~\ref{fig:tex_representation}(a)). 
This representation is also adopted in recent work such as~\cite{tewari2017mofa, tewari2018self}.
There is an albedo intensity value corresponding to each vertex in the face mesh. 
Despite widely used, this representation has its limitations.
Since $3$D vertices are not defined on a $2$D grid, this representation is mostly parameterized as a vector, which not only loses the spatial relation of its vertices, but also prevents it to leverage the convenience of deploying CNN on $2$D albedo.
In contrast, given the rapid progress in image synthesis, it is desirable to choose a $2$D image, e.g., a frontal-view face image in Fig.~\ref{fig:tex_representation}(b), as an albedo representation. 
However, frontal faces contain little information of two sides, which would lose many albedo information for side-view faces.

In light of these consideration, we use an unwrapped $2$D texture as our texture representation (Fig.~\ref{fig:tex_representation}(c)).
Specifically, each $3$D vertex $\mathbf{v}$  is projected onto the UV space using cylindrical unwarp. 
Assuming that the face mesh has the top pointing up the $y$ axis, the projection of $\mathbf{v} = (x, y, z)$ onto the UV space $\mathbf{v}^{\text{uv}} = (u, v)$ is computed as:
\begin{equation}
 v \rightarrow \alpha_1 . \text{arctan} \left( \frac{x}{z} \right) + \beta_1, \hspace{3mm} u \rightarrow \alpha_2 . y + \beta_2,
 \label{eqn:unwarp}
\end{equation}
where $\alpha_1, \alpha_2, \beta_1, \beta_2$ are constant scale and translation scalars to place the unwrapped face into the image boundaries.
Here, per-vertex albedo $\mathbf{A}\in\mathbb{R}^{3Q}$ could be easily computed by sampling from its UV space counterpart $\mathbf{A}^{\text{uv}}\in\mathbb{R}^{U \times V}$: 
\begin{equation}
\mathbf{A}(\mathbf{v}) = \mathbf{A}^{\text{uv}}(\mathbf{v}^{\text{uv}}).
\label{eqn:uv_convert1}
\end{equation}
Usually, it involves sub-pixel sampling via bilinear interpolation: 
\begin{equation}
\mathbf{A}(\mathbf{v}) = \! \! \! \sum_{ \substack{ u' \in \{ \floor{u}, \ceil{u} \} \\ v' \in \{ \floor{v}, \ceil{v} \} }} \! \! \! \mathbf{A}^{\text{uv}}(u',v')(1{-}|u{-}u'|) (1{-}|v{-}v'|),
\label{eqn:uv_convert2}
\end{equation}
where $\mathbf{v}^{\text{uv}} = (u, v)$ is the UV space projection of $\mathbf{v}$ via Eqn.~\ref{eqn:unwarp}.

Albedo information is naturally expressed in the UV space but spatial data can be embedded in the same space as well. 
Here, a $3$D facial mesh can be represented as a $2$D image with three channels, one for each spatial dimension $x$, $y$ and $z$. 
Fig~\ref{fig:shape_representationUV} gives an example of this UV space shape representation $\mathbf{S}^{\text{uv}}\in\mathbb{R}^{U \times V}$.

Representing $3$D face shape in UV space allow us to use a CNN for shape decoder $D_S$ instead of using a multi-layer perceptron (MLP) as in our preliminary version~\cite{tran2018nonlinear}. Avoiding using wide fully-connected layers allow us to use deeper network for $D_S$, potentially model more complex shape variations. This results in better fitting results as being demonstrated in our experiment (Sec.~\ref{sec:abl_1d_2d}).

Also, it is worth to note that different from our preliminary version~\cite{tran2018nonlinear} where the reference UV space, for texture, is build upon projection of the mean shape with neutral expression; in this version, the reference shape used has the mouth open. This change helps the network to avoid learning a large gradient near the two lips' borders in the vertical direction when the mouth is open.


To regress these $2$D representation of shape and albedo, we can employ CNNs as shape and albedo networks respectively. Specifically, $D_S$, $D_A$ are CNN constructed by multiple fractionally-strided convolution layers. After each convolution is batchnorm and eLU activation, except the last convolution layers of encoder and decoders. The output layer has a $tanh$ activation to constraint the output to be in the range of $[-1, 1]$. The detailed network architecture is presented in Tab.~\ref{tab:network}.

\SubSubSection{In-Network Physically-Based Face Rendering}
\label{sec:rendering}

To reconstruct a face image from the albedo $\mathbf{A}$, shape $\mathbf{S}$, lighting parameter $\mathbf{L}$, and projection parameter $\mathbf{m}$, we define a rendering layer $\mathcal{R}(\mathbf{m},\mathbf{L},\mathbf{S},\mathbf{A})$  to render a face image from the above parameters.
This is accomplished in three steps, as shown in Fig.~\ref{fig:render}.
Firstly, the facial texture is computed using the albedo $\mathbf{A}$ and the surface normal map of the rotated shape $N(\mathbf{V}) = N(\mathbf{m}, \mathbf{S})$. 
Here, following~\cite{wang2009face}, we assume distant illumination and a purely \emph{Lambertian} surface reflectance.
Hence the incoming radiance can be approximated using spherical harmonics (SH) basis functions $H_b: \mathbb{R}^3 \rightarrow \mathbb{R}$, and controlled by coefficients $\mathbf{L}$.
Specifically, the texture in UV space $\mathbf{T}^{\text{uv}}\in\mathbb{R}^{U \times V}$ is composed of albedo $\mathbf{A}^{\text{uv}}$ and shading $\mathbf{C}^{\text{uv}}$:
\begin{equation}
\mathbf{T}^{\text{uv}} = \mathbf{A}^{\text{uv}} \odot \mathbf{C}^{\text{uv}} = \mathbf{A}^{\text{uv}} \odot \sum_{b=1}^{B^2}{ L_b H_b(N(\mathbf{m},\mathbf{S}^{\text{uv}}))},
\end{equation}
where $B$ is the number of spherical harmonics bands. We use $B=3$, which leads to $B^2=9$ coefficients in $\mathbf{L}$ for each of three color channels. 
%
Secondly, the $3$D shape/mesh $\mathbf{S}$ is projected to the image plane via Eqn.~\ref{eqn:projection}.
Finally, the $3$D mesh is then rendered using a Z-buffer renderer, where each pixel is associated with a single triangle of the mesh,
\begin{align}
\mathbf{\hat{I}}(m, n) & = \mathcal{R}(\mathbf{m}, \mathbf{L}, \mathbf{S}^{\text{uv}}, \mathbf{A}^{\text{uv}})_{m,n} \nonumber \\
 &=  \mathbf{T}^{\text{uv}}( \! \! \! \sum_{\mathbf{v}_i \in \Phi^ {\text{uv}}(g, m, n) } \! \! \! \lambda_i \mathbf{v}_i), 
\end{align}
where $\Phi(g, m, n) = \{\mathbf{v}_{1}, \mathbf{v}_2, \mathbf{v}_3 \}$ is an operation returning three vertices of the triangle that encloses the pixel $(m, n)$ after projection $g$; $\Phi^{\text{uv}}(g, m, n)$ is the same operation with resultant vertices mapped into the referenced UV space using Eqn.~\ref{eqn:unwarp}.
In order to handle occlusions, when a single pixel resides in more than one triangle, the triangle that is closest to the image plane is selected. 
The final location of each pixel is determined by interpolating the location of three vertices via barycentric coordinates $\{\lambda_{i}\}_{i=1}^3$.

There are alternative designs to our rendering layer. 
If the texture representation is defined per vertex, as in Fig.~\ref{fig:tex_representation}(a), one may warp the input image $\mathbf{I}_i$ onto the vertex space of the $3$D shape $\mathbf{S}$, whose distance to the per-vertex texture representation can form a reconstruction loss. 
This design is adopted by the recent work of~\cite{tewari2017mofa, tewari2018self}.
In comparison, our rendered image is defined on a $2$D grid while the alternative is on top of the $3$D mesh.
As a result, our rendered image can enjoy the convenience of applying the perceptual loss or adversarial loss, which is shown to be critical in improving the quality of synthetic texture.
Another design for rendering layer is image warping based on the spline interpolation, as in~\cite{cole2017face}. 
However, this warping is continuous: every pixel in the input will map to the output. 
Hence this warping operation fails in the occluded region. 
As a result, Cole~\etal~\cite{cole2017face} limit their scope to only synthesizing frontal-view faces by warping from normalized faces.

The CUDA implementation of our rendering layer is publicly available at~\url{https://github.com/tranluan/Nonlinear_Face_3DMM}.

\SubSubSection{Occlusion-aware Rendering}
\label{sec:occlusion_aware}

Very often, in-the-wild faces are occluded by glasses, hair, hands, etc. Trying to reconstruct abnormal occluded regions could make the model learning more difficult or result in an model with external occlusion baked in. Hence, we propose to use a segmentation mask to exclude occluded regions in the rendering pipeline: 
\begin{equation}
\mathbf{\hat{I}} \leftarrow \mathbf{\hat{I}}\odot \mathbf{M} + \mathbf{I} \odot (1-\mathbf{M}).
\end{equation}

\begin{figure}[t!]
\begin{center}
\small
\begin{tabular}{ @{}c@{\hskip 0.01mm}c@{\hskip .01mm}c@{\hskip .01mm}c@{\hskip .01mm}c@{\hskip .01mm}c@{\hskip .01mm}c@{\hskip .01mm}c@{}}
\includegraphics[width=\AEFigWid]{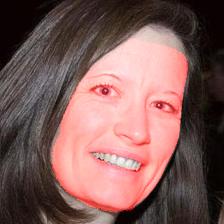} &
\includegraphics[width=\AEFigWid]{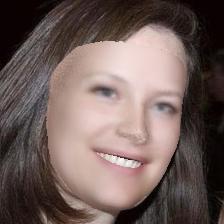} &
\includegraphics[width=\AEFigWid]{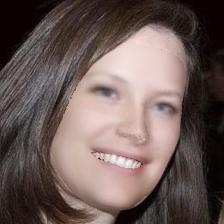} &&
\includegraphics[width=\AEFigWid]{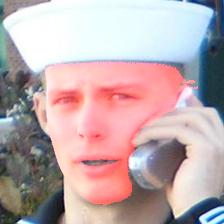} &
\includegraphics[width=\AEFigWid]{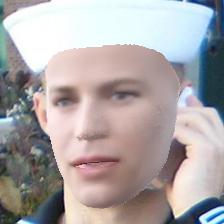} &
\includegraphics[width=\AEFigWid]{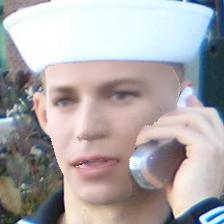} \\
\end{tabular}
\caption{\small Rendering with segmentation masks. Left to right: segmentation results, naive rendering, occulusion-aware rendering.}
\label{fig:seg}\figvspace 
\end{center}
\end{figure}

As a result, these occluded regions won't affect our optimization process.
The foreground mask $\mathbf{M}$ is estimated using the segmentation method given by Nirkin\etal~\cite{nirkin2018faceswap}.
Examples of segmentation masks and rendering results  can be found in Fig.~\ref{fig:seg}.

\begin{table}[t!]
\caption{\small The architectures of $E$, $D_A$ and $D_S$ networks.}
\tabvspace
\label{tab:network}
\begin{center}
\small
\resizebox{0.99\linewidth}{!}{
\setlength{\tabcolsep}{3pt}
\begin{tabular}{ @{}ccccccccc@{} }
\toprule
\multicolumn{3}{c}{$E$} & \hspace{2mm} 
& \multicolumn{3}{c}{$D_A/D_S$} \\
\cmidrule(r){1-3}
\cmidrule(r){5-7}
Layer & Filter/Stride & Output Size && Layer & Filter/Stride & Output Size \\ \midrule
&&&& FC & & $6{\times}7{\times}320$ \\
Conv11 & $7{\times}7/2$ & $112{\times}112{\times}32$ && FConv52& $3{\times}3/2$ & $12{\times}14{\times}160$ \\
Conv12 & $3{\times}3/1$ & $112{\times}112{\times}64$ && FConv51& $3{\times}3/1$ & $12{\times}14{\times}256$ \\\midrule
Conv21 & $3{\times}3/2$ & $56{\times}56{\times}64$ && FConv43& $3{\times}3/2$ & $24{\times}28{\times}256$&&\\ 
Conv22 & $3{\times}3/1$ & $56{\times}56{\times}64$  && FConv42& $3{\times}3/1$ & $24{\times}28{\times}128$ \\
Conv23 & $3{\times}3/1$ & $56{\times}56{\times}128$ && FConv41& $3{\times}3/1$ & $24{\times}28{\times}192$ \\
\midrule
Conv31 & $3{\times}3/2$ & $28{\times}28{\times}128$ && FConv33& $3{\times}3/2$ & $48{\times}56{\times}192$ && \\ 
Conv32 & $3{\times}3/1$ & $28{\times}28{\times}96$  && FConv32& $3{\times}3/1$ & $48{\times}56{\times}96$ \\
Conv33 & $3{\times}3/1$ & $28{\times}28{\times}192$ && FConv31& $3{\times}3/1$ & $48{\times}56{\times}128$ \\
\midrule
Conv41 & $3{\times}3/2$ & $14{\times}14{\times}192$ && FConv23& $3{\times}3/2$ & $96{\times}112{\times}128$ \\ 
Conv42 & $3{\times}3/1$ & $14{\times}14{\times}128$ && FConv22& $3{\times}3/1$ & $96{\times}112{\times}64$ \\
Conv43 & $3{\times}3/1$ & $14{\times}14{\times}256$ && FConv21& $3{\times}3/1$ & $96{\times}112{\times}64$ \\
\midrule
Conv51 & $3{\times}3/2$ & $7{\times}7{\times}256$ && FConv13& $3{\times}3/2$ & $192{\times}224{\times}64$ \\ 
Conv52 & $3{\times}3/1$ & $7{\times}7{\times}160$ && FConv12& $3{\times}3/1$ & $192{\times}224{\times}32$ \\
Conv53 & $3{\times}3/1$ & $7{\times}7{\times}(l_S{+}l_A{+}64)$ && FConv11& $3{\times}3/1$ & $192{\times}224{\times}3$ \\
\midrule
AvgPool& $7{\times}7/1$ & $1{\times}1{\times}(l_S{+}l_A{+}64)$ && \\ \midrule
FC$_{\mathbf{m}}$ & $64{\times}6$& $6$ \\ \midrule
FC$_{\mathbf{L}}$ & $64{\times}27$& $27$ \\ \bottomrule
\end{tabular}}
\end{center}
\end{table}

\SubSubSection{Model Learning}
\label{sec:network}

The entire network is end-to-end trained to reconstruct the input images, with the loss function:
\begin{equation}
L = L_{\text{rec}} +  \lambda_L L_{\text{lan}} + \lambda_{ \text{reg} } L_{ \text{reg} },
\label{eqn:overallLoss}
\end{equation}
where the reconstruction loss $L_{\text{rec}}$ enforces the rendered image $\mathbf{\hat{I}}$ to be similar to the input $\mathbf{I}$, the landmark loss $L_L$ enforces geometry constraint, and the regularization loss $L_{ \text{reg}} $ encourages plausible solutions. 

\Paragraph{Reconstruction Loss}
The main objective of the network is to reconstruct the original face via disentangle representation. Hence, we enforce the reconstructed image to be similar to the original input image:
\begin{equation}
L^i_{\text{rec}} = \frac{1}{|\mathcal{V}|} \sum_{q \in \mathcal{V} }||\mathbf{\hat{I}}(q) - \mathbf{I}(q)||_2
\label{eqn:reconLoss}
\end{equation}
where $\mathcal{V}$ is the set of all pixels in the images covered by the estimated face mesh.
There are different norms can be used to measure the closeness. 
To better handle outliers, we adopt the robust $l_{2,1}$, where the distance in the $3$D RGB color space is based on $l_2$ and the summation over all pixels enforces sparsity based on $l_1$-norm~\cite{thies2016face2face,thies2016facevr}.

To improve from blurry reconstruction results of $l_p$ losses, in our preliminary work~\cite{tran2018nonlinear}, thanks for our rendering layer, we employ adversarial loss to enhance the image realism. However, adversarial objective only encourage the reconstruction to be close to the real image distribution but not necessary the input image. Also, it's known to be not stable to optimize. Here, we propose to use a perceptual loss to enforce the closeness between images $\mathbf{\hat{I}}$ and $\mathbf{I}$, which overcomes both of adversarial loss's weaknesses.
Besides encouraging the pixels of the output image $\mathbf{\hat{I}}$ to exactly match the pixels of the input $\mathbf{I}$, we encourage them to have similar feature representations as computed by the loss network $\varphi$.
\begin{equation}
L^f_{\text{rec}} = \frac{1}{|\mathcal{C}|} \sum_{j \in \mathcal{C}} \frac{1}{W_jH_jC_j}|| \varphi_j(\mathbf{\hat{I}}) - \varphi_j(\mathbf{I})||^2_2.
\label{eqn:perceptualLoss}
\end{equation}
We choose VGG-Face\cite{parkhi15deep} as our $\varphi$ to leverage its face-related features and also because of simplicity. The loss is summed over $\mathcal{C}$, a subset of layers of $\varphi$. Here $\varphi_j(\mathbf{I})$ is the activations of the $j$-th layer of $\varphi$ when processing the image $\mathbf{I}$ with dimension $W_j \times H_j \times C_j$.
This feature reconstruction loss is one of perceptual losses widely used in different image processing tasks~\cite{johnson2016perceptual}.

The final reconstruction loss is a weighted sum of two terms:
\begin{equation}
L_{\text{rec}} = L^i_{\text{rec}} + \lambda_f L^f_{\text{rec}}.
\end{equation}

\Paragraph{Sparse Landmark Alignment}
To help achieving better model fitting, which in turn helps to improve the model learning itself, we employ the landmark alignment loss, measuring Euclidean distance between estimated and groundtruth landmarks,  as an auxiliary task,
\begin{align}
L_{\text{lan}} = & \norm {M(\mathbf{m}) \ast \begin{bmatrix} \mathbf{S}(:,\mathbf{d}) \\ \mathbf{1} \end{bmatrix} - \mathbf{U} }^2_2,
\label{eq:landmarkloss}
\end{align}
where $\mathbf{U} \in \mathbb{R}^{2{\times} 68}$ is the manually labeled $2$D landmark locations, $\mathbf{d}$ is a constant $68$-dim vector storing the indexes of $68$ $3$D vertices corresponding to the labeled $2$D landmarks.
Different from traditional face alignment work where the shape bases are fixed, our work jointly learns
the bases functions (i.e., the shape decoder $D_S$) as well. 
Minimizing the landmark loss while updating $D_S$ only moves a tiny subsets of vertices. 
If the shape $\mathbf{S}$ is represented as a vector and $D_S$ is a MLP consisting of fully connected layers, vertices are independent. Hence $L_{\text{L}}$ only adjusts $68$ vertices. 
In case $\mathbf{S}$ is represented in the UV space and $D_S$ is a CNN, local neighbor region could also be modified. 
In both cases, updating $D_S$ based on $L_{\text{L}}$ only moves a subsets of vertices, which could lead to implausible shapes. 	
Hence, when optimizing the landmark loss, we fix the decoder $D_S$ and only update the encoder.

Also, note that different from some prior work~\cite{garrido2016reconstruction}, our network only requires ground-truth landmarks during training. It is able to predict landmarks via $\mathbf{m}$ and $\mathbf{S}$ during the test time.

\Paragraph{Regularizations}
To ensure plausible reconstruction, we add a few regularization terms:

\textit{Albedo Symmetry} As the face is symmetry, we enforce the albedo symmetry constraint,
\begin{equation}
L_{\text{sym}} = \norm{ \mathbf{A}^{\text{uv}} - \text{flip}(\mathbf{A}^{\text{uv}}) }_1.
\label{eqn:alb_sym}
\end{equation}

Employing on $2$D albedo, this constraint can be easily implemented via a horizontal image flip operation $\text{flip}()$.

\textit{Albedo Constancy}
Using symmetry constraint can help to correct the global shading. 
However, symmetrical details, i.e., dimples, can still be embedded in the albedo channel. 
To further remove shading from the albedo channel, following Retinex theory which assumes
albedo to be piecewise constant, we enforce sparsity in two directions of its gradient, similar to~\cite{meka2016live, shu2017neural}:
\begin{equation}
L_{\text{const}} =  \sum_{\mathbf{v}_j^{\text{uv}} \in \mathcal{N}_i} \omega (\mathbf{v}_i^{\text{uv}}, \mathbf{v}_j^{\text{uv}}) \norm{ \mathbf{A}^{\text{uv}}(\mathbf{v}_i^{\text{uv}}) - \mathbf{A}^{\text{uv}}(\mathbf{v}_j^{\text{uv}}) }_2^p,
\label{eqn:alb_const}
\end{equation}
where $\mathcal{N}_i$ denotes a set of $4$-pixel neighborhood of pixel $\mathbf{v}_i^{\text{uv}}$.
With the assumption that pixels with the same chromaticity (i.e., $\mathbf{c}(x) = \mathbf{I}(x) / |\mathbf{I}(x)|$)  are more likely to have the same albedo, we set the constant weight $\omega (\mathbf{v}_i^{\text{uv}}, \mathbf{v}_j^{\text{uv}}) = \exp\left(-\alpha \norm{ \mathbf{c}(\mathbf{v}_i^{\text{uv}}) - \mathbf{c}(\mathbf{v}_j^{\text{uv}} ) } \right) $, where the color is referenced from the input image using the current estimated projection. Following~\cite{meka2016live}, we set $\alpha=15$ and $p=0.8$ in our experiment.

\textit{Shape Smoothness}
For shape component, we impose the smoothness by adding the Laplacian regularization on the vertex locations for the set of all vertices.
\begin{equation}
L_{\text{smooth}} = \sum_{\mathbf{v}_i^{\text{uv}} \in \mathbf{S}^{\text{uv}}} \norm{ \mathbf{S}^{\text{uv}}(\mathbf{v}_i^{\text{uv}}) - \frac{1}{|\mathcal{N}_i|}  \sum_{\mathbf{v}_j^{\text{uv}} \in \mathcal{N}_i} \mathbf{S}^{\text{uv}}(\mathbf{v}_j^{\text{uv}}) }_2.
\label{eqn:alb_smooth}
\end{equation}

\Paragraph{Intermediate Semi-Supervised Training}  
Fully unsupervised training using only the reconstruction and adversarial loss on the rendered images could lead to a degenerate solution, since the initial estimation is far from ideal to render meaningful images. 
Therefore, we introduce intermediate loss functions to guide the training in the early iterations.

With the face profiling technique, Zhu~\etal~\cite{zhu2016face} expand the 300W dataset~\cite{sagonas2016300} into $122,450$ images with fitted $3$DMM shapes $\widetilde{\mathbf{S}}$ and projection parameters $\widetilde{\mathbf{m}}$. 
Given $\widetilde{\mathbf{S}}$ and $\widetilde{\mathbf{m}}$, we create the pseudo groundtruth texture $\widetilde{\mathbf{T}}$ by referring every pixel in the UV space back to the input image, i.e., the backward of our rendering layer. 
With $\widetilde{\mathbf{m}}$, $\widetilde{\mathbf{S}}$, $\widetilde{\mathbf{T}}$, we define our intermediate loss by: 
\begin{equation}
L_0 = L_{\text{S}} + \lambda_T L_{\text{T}} + \lambda_m L_{\text{m}} + \lambda_L L_{\text{L}} + \lambda_{\text{reg}}L_{\text{reg}},
\end{equation}
where: $L_{\text{S}} = \norm{ \mathbf{S}-\mathbf{\widetilde{S}} }^2_2$, $L_{\text{T}} = \norm{ \mathbf{T}-\mathbf{\widetilde{T}} }_1$, $L_{\text{m}} = \norm{ \mathbf{m}-\mathbf{\widetilde{m}} }^2_2$.

It's also possible to provide pseudo groundtruth to the SH coefficients $\mathbf{L}$ and followed by albedo $\mathbf{A}$ using least square optimization with a constant albedo assumption, as has been done in~\cite{wang2009face,shu2017neural}. 
However, this estimation is not reliable for in-the-wild images with occlusion regions. 
Also empirically, with proposed regularizations, the model is able to explore plausible solutions for these components by itself. 
Hence, we decide to refrain from supervising $\mathbf{L}$ and $\mathbf{A}$ to simplify our pipeline.

Due to the pseudo groundtruth, using $L_0$ may run into the risk that our solution learns to mimic the linear model. 
Thus, we switch to the loss of Eqn.~\ref{eqn:overallLoss} after $L_0$ converges.
Note that the estimated groundtruth of $\widetilde{\mathbf{m}}$, $\widetilde{\mathbf{S}}$, $\widetilde{\mathbf{T}}$ and the landmarks are the only supervision used in our training, for which our learning is considered as {\it weakly} supervised.

\section{Experimental Results}
\label{sec:exp}

The experiments study three aspects of the proposed nonlinear $3$DMM, in terms of its expressiveness, representation power, and applications to facial analysis.
Using facial mesh triangle definition by Basel Face Model~(BFM)~\cite{paysan20093d}, we train our $3$DMM using 300W-LP dataset~\cite{zhu2016face}, which contains $122,450$ in-the-wild face images, in a wide pose range from $-90^{\circ}$ to $90^{\circ}$.
Images are loosely square cropped around the face and scale to $256 \times 256$. 
During training, images of size $224 \times 224$ are randomly cropped from these images to introduce translation variations. 

The model is optimized using Adam optimizer with a learning rate of $0.001$ in both training stages.
We set the following parameters: $Q=53,215$, $U=192, V=224$, $l_S=l_T=160$. $\lambda$~values are set to make losses to have similar magnitudes.

\begin{figure}[t!]
\begin{center}
\small
\setlength{\tabcolsep}{3pt}
\begin{tabular}{ @{\hskip .5 mm}c@{\hskip 1 mm}c|c@{\hskip .5 mm}c@{\hskip .5 mm}c@{\hskip .5 mm}c@{\hskip .5 mm}c}
Sym & Const &  Input & Overlay & Albedo & Shading & Texture \\ \midrule
    &       & \includegraphics[height=\AblTexFigHeight]{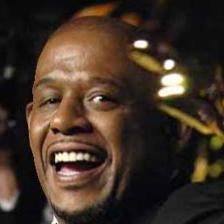} &
\includegraphics[height=\AblTexFigHeight]{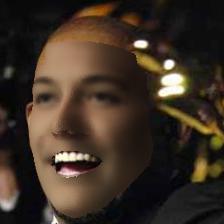} &
\includegraphics[height=\AblTexFigHeight]{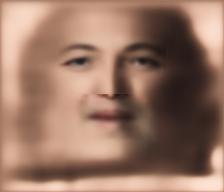} &
\includegraphics[height=\AblTexFigHeight]{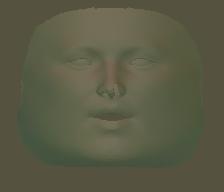} &
\includegraphics[height=\AblTexFigHeight]{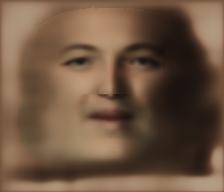}
\\
\checkmark &       & 
&
\includegraphics[height=\AblTexFigHeight]{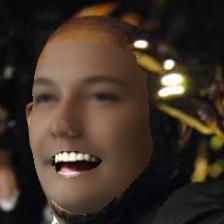} &
\includegraphics[height=\AblTexFigHeight]{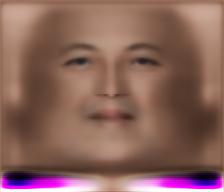} &
\includegraphics[height=\AblTexFigHeight]{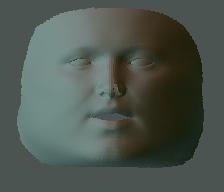} &
\includegraphics[height=\AblTexFigHeight]{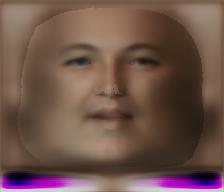}
\\
\checkmark & \checkmark &
&
\includegraphics[height=\AblTexFigHeight]{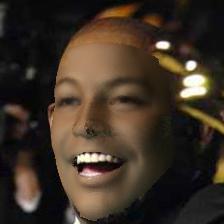} &
\includegraphics[height=\AblTexFigHeight]{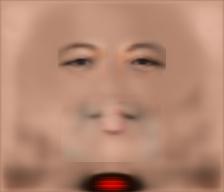} &
\includegraphics[height=\AblTexFigHeight]{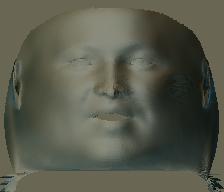} &
\includegraphics[height=\AblTexFigHeight]{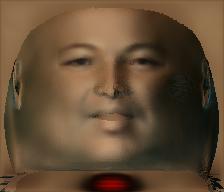}
\\
\end{tabular}
\caption{\small Effect of albedo regularizations: albedo symmetry (sym) and albedo constancy (const). When there is no regularization being used, shading is mostly baked into the albedo. Using the symmetry property helps to resolve the global lighting. Using constancy constraint futher removes shading from the albedo, which results in a better $3$D shape.}
\label{fig:abl_albedo}\figvspace
\end{center}
\end{figure}

\SubSection{Ablation Study}
\label{sec:ablation_study}

\SubSubSection{Effect of Regularization}
\label{sec:abl_regu}
\Paragraph{Albedo Regularization}
In this work, to regularize albedo learning, we employ two constraints to efficiently remove shading from albedo namely albedo symmetry and constancy. To demonstrate the effect of these regularization terms, we compare our full model with its partial variants: one without any albedo reqularization and one with the symmetry constraint only. 
Fig.~\ref{fig:abl_albedo} shows visual comparison of these models. 
Learning without any constraints results in the lighting is totally explained by the albedo, meanwhile is the shading is almost constant (Fig.~\ref{fig:abl_albedo}(a)). 
Using symmetry help to correct the global lighting. 
However, symmetric geometry details are still baked into the albedo (Fig.~\ref{fig:abl_albedo}(b)). Enforcing albedo constancy helps to further remove shading from it (Fig.~\ref{fig:abl_albedo}(c)). Combining these two regularizations helps to learn plausible albedo and lighting, which improves the shape estimation.

\begin{figure}[t!]
\begin{center}
\small
\setlength{\tabcolsep}{3pt}

\begin{tabular}{ @{}c@{\hskip 1.mm}c@{}c@{\hskip 1.mm}c@{}c@{}c@{}}
Input & Overlay & Shape &  Overlay & Shape \\
\includegraphics[width=\FittingFigWid]{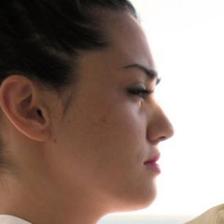} &
\includegraphics[width=\FittingFigWid]{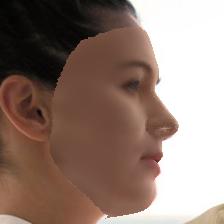} &
\includegraphics[width=\FittingFigWid]{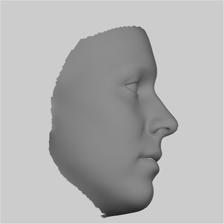} &
\includegraphics[width=\FittingFigWid]{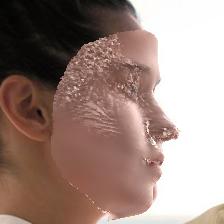} &
\includegraphics[width=\FittingFigWid]{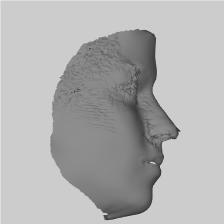} \\
\includegraphics[width=\FittingFigWid]{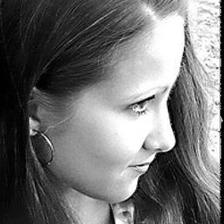} &
\includegraphics[width=\FittingFigWid]{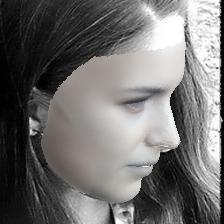} &
\includegraphics[width=\FittingFigWid]{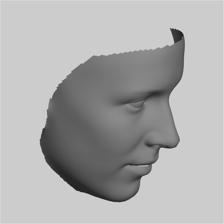} &
\includegraphics[width=\FittingFigWid]{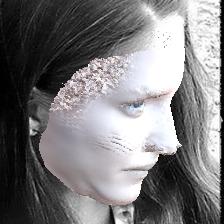} &
\includegraphics[width=\FittingFigWid]{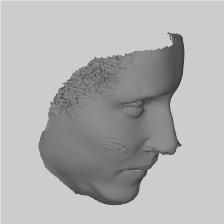}
\\
& \multicolumn{2}{c}{With smoothness} & \multicolumn{2}{c}{Without smoothness}
\end{tabular}
\caption{\small Effect of shape smoothness regularization. }
\label{fig:alb_smoothness}\figvspace
\end{center}
\end{figure}

\Paragraph{Shape Smoothness Regularization}
We also evaluate the need in shape regularization. 
Fig.~\ref{fig:alb_smoothness} shows visual comparisons between our model and its variant without the shape smoothness constraint. 
Without the smoothness term the learned shape becomes noisy especially on two sides of the face. 
The reason is that, the hair region is not completely excluded during training because of imprecise segmentation estimation.

\SubSubSection{ Modeling Lighting and  Shape Representation}
\label{sec:abl_1d_2d}

In this work, we make two major algorithmic differences with our preliminary work~\cite{tran2018nonlinear}: incorporating lighting into the model and changing the shape representation. 

Our previous work~\cite{tran2018nonlinear} models the texture directly, while this work disentangles the shading from the albedo.
As argued, modeling the lighting should have a positive impact on shape learning. 
Hence we compare our models with results from~\cite{tran2018nonlinear} in face alignment task.

Also, in our preliminary work~\cite{tran2018nonlinear}, as well as in traditional $3$DMM, shape is represented as a vector, where vertices are independent. 
Despite this shortage, this approach has been widely adopted due to its simplicity and sampling efficiency. 
In this work, we explore an alternative to this representation: represent the $3$D shape as a position map in the $2$D UV space. 
This representation has three channels: one for each spatial dimension. 
This representation maintains the spatial relation among facial mesh's vertices. 
Also, we can use CNN as the shape decoder replacing an expensive MLP. 
Here we also evaluate the performance gain by switching to this representation.

Tab.~\ref{tab:abl_1d_2d} reports the performance on the face alignment task of different variants. 
As a result, modeling lighting helps to reduce the error from $4.70$ to $4.30$. 
Using the $2$D representation, with the convenience of using CNN, the error is further reduced to $4.12$. 

\begin{table}[t!]
\caption{\small{Face alignment performance on ALFW2000.}} 
\label{tab:abl_1d_2d}
\tabvspace
\begin{center}
\begin{tabular}{ cccccccc}
\toprule
Method & Lighting & UV shape & NME \\ \midrule 
Our~\cite{tran2018nonlinear} & & & 4.70 \\
Our & \checkmark &   & 4.30 \\
Our & \checkmark & \checkmark & $\mathbf{4.12}$ \\ 
\bottomrule
\end{tabular}
\end{center}
\end{table}

\begin{figure}[t!]
\begin{center}
\small
\setlength{\tabcolsep}{3 pt}
\begin{tabular}{ @{}c@{\hskip 0.mm}c@{}c@{\hskip 1.mm}c@{}c@{}c@{}}
Input & Our & AE & Input & Our & AE \\
\includegraphics[width=\AEFigWid]{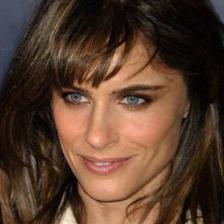} &
\includegraphics[width=\AEFigWid]{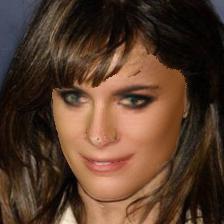} &
\includegraphics[width=\AEFigWid]{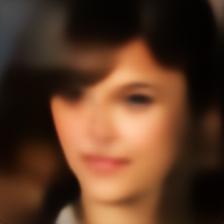} &
\includegraphics[width=\AEFigWid]{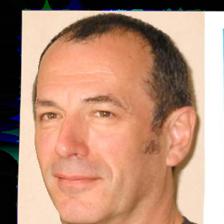} &
\includegraphics[width=\AEFigWid]{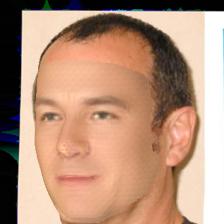} &
\includegraphics[width=\AEFigWid]{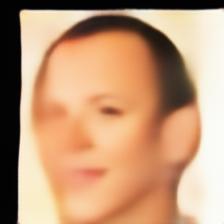}
\end{tabular}
\caption{\small Comparison to convolutional autoencoders (AE). Our approach produces results of higher quality. Also it provides access to the $3$D facial shape, albedo, lighting, and projection matrix.  }
\label{fig:autoencdoer}\figvspace
\end{center}
\end{figure}

\SubSubSection{Comparison to Autoencoders}
\label{sec:autoencoder}
We compare our model-based approach with a convolutional autoencoder in Fig.~\ref{fig:autoencdoer}. The autoencoder network has a similar depth and model size as ours. It gives blurry reconstruction results as the dataset contain large variations on face appearance, pose angle and even diversity background.
Our model-based approach obtains sharper reconstruction results and provides semantic parameters allowing access to different components including $3$D shape, albedo, lighting and projection matrix.

\begin{figure}[t!]

\begin{center}
\small
\setlength{\tabcolsep}{3pt}
\begin{tabular}{ @{}c@{}c@{}c@{}c@{}c@{}c@{}c@{}c@{}}
\includegraphics[trim=400 180 370 230,clip,width=\VaryingShapeFigWid]{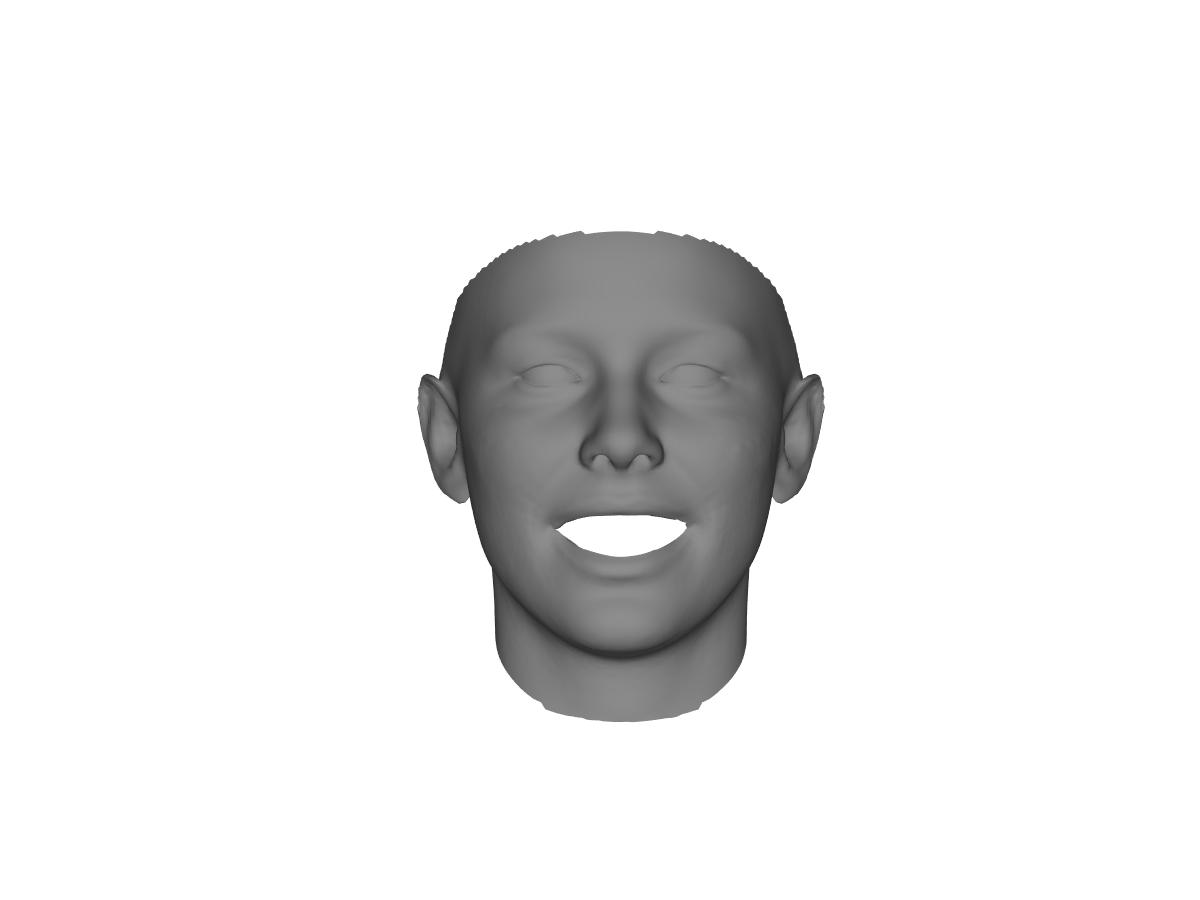} &
\includegraphics[trim=400 180 370 230,clip,width=
\includegraphics[trim=400 180 370 230,clip,width=\VaryingShapeFigWid]{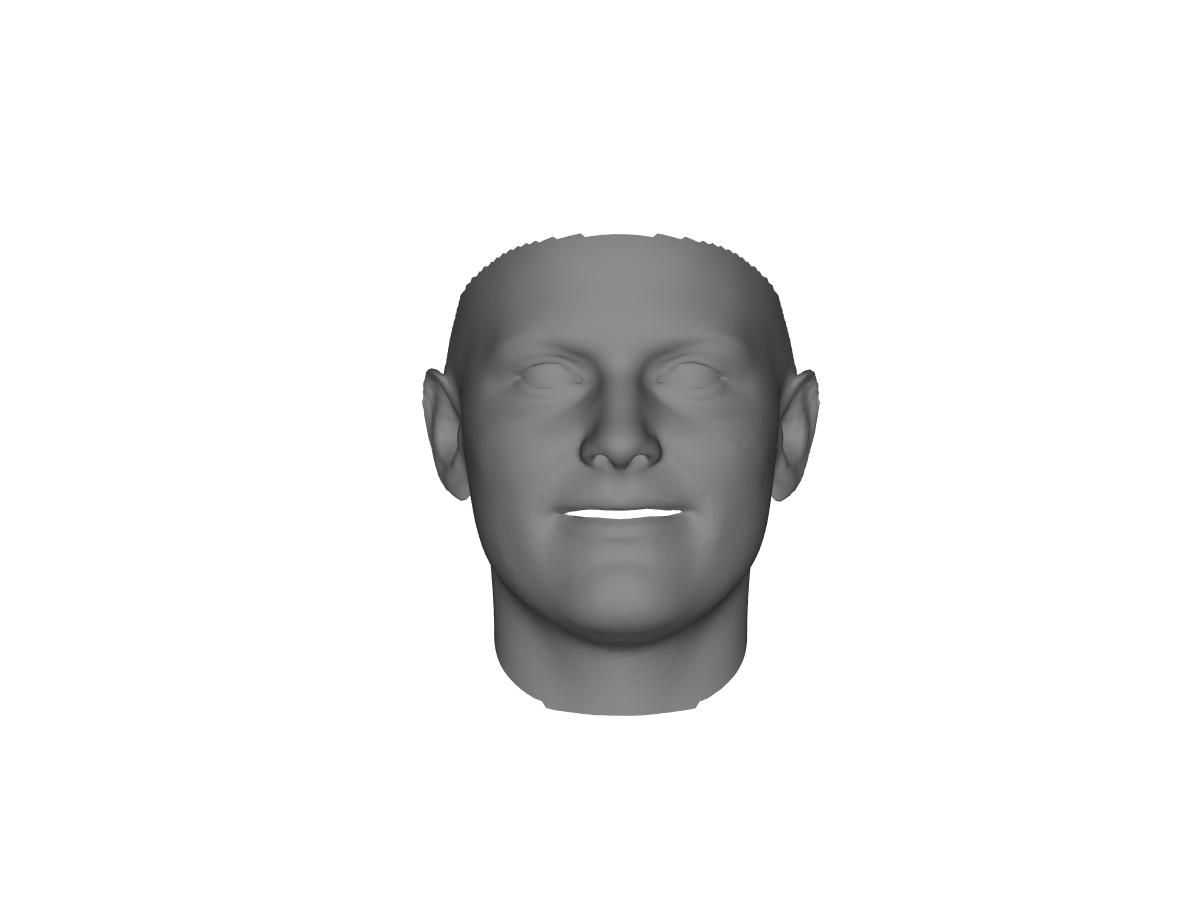} &
\includegraphics[trim=400 180 370 230,clip,width=\VaryingShapeFigWid]{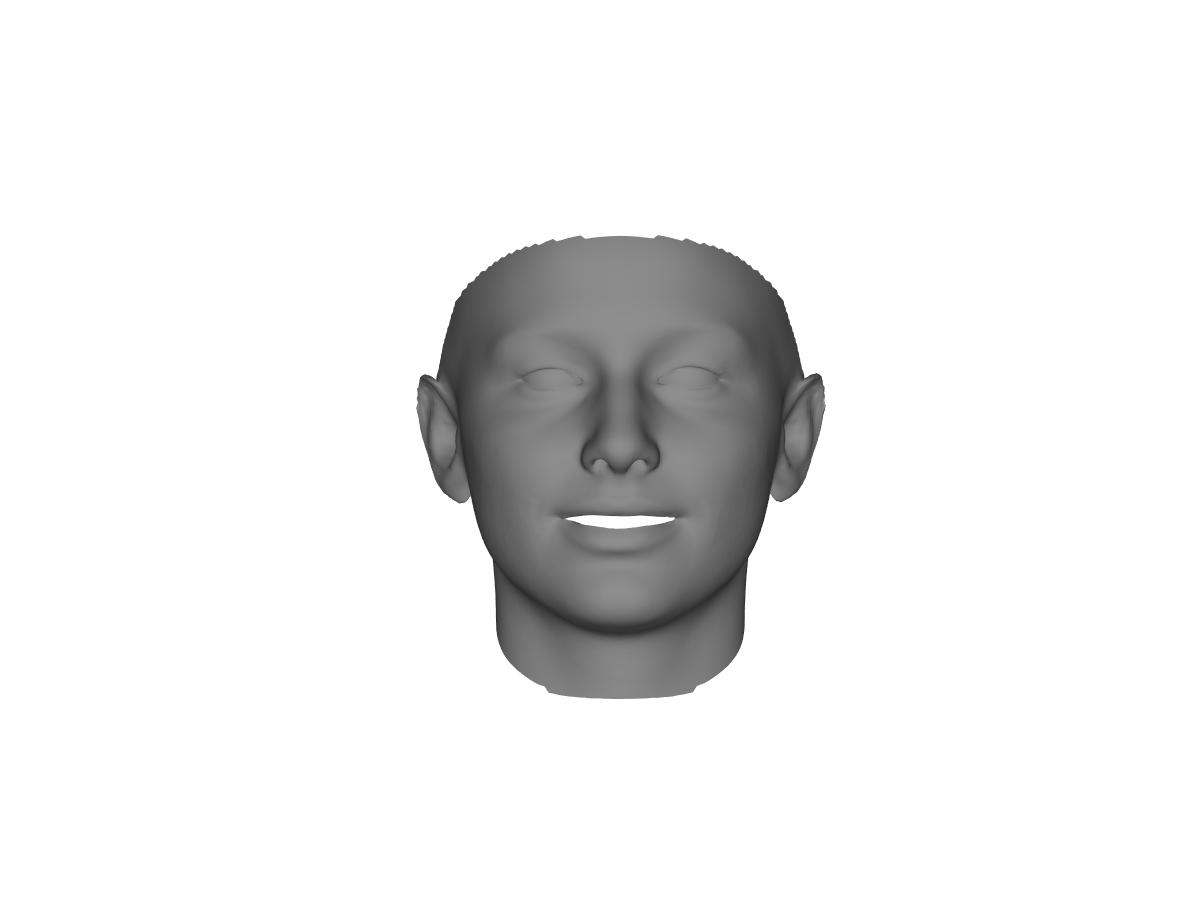} &
\includegraphics[trim=400 180 370 230,clip,width=
\includegraphics[trim=400 180 370 230,clip,width=\VaryingShapeFigWid]{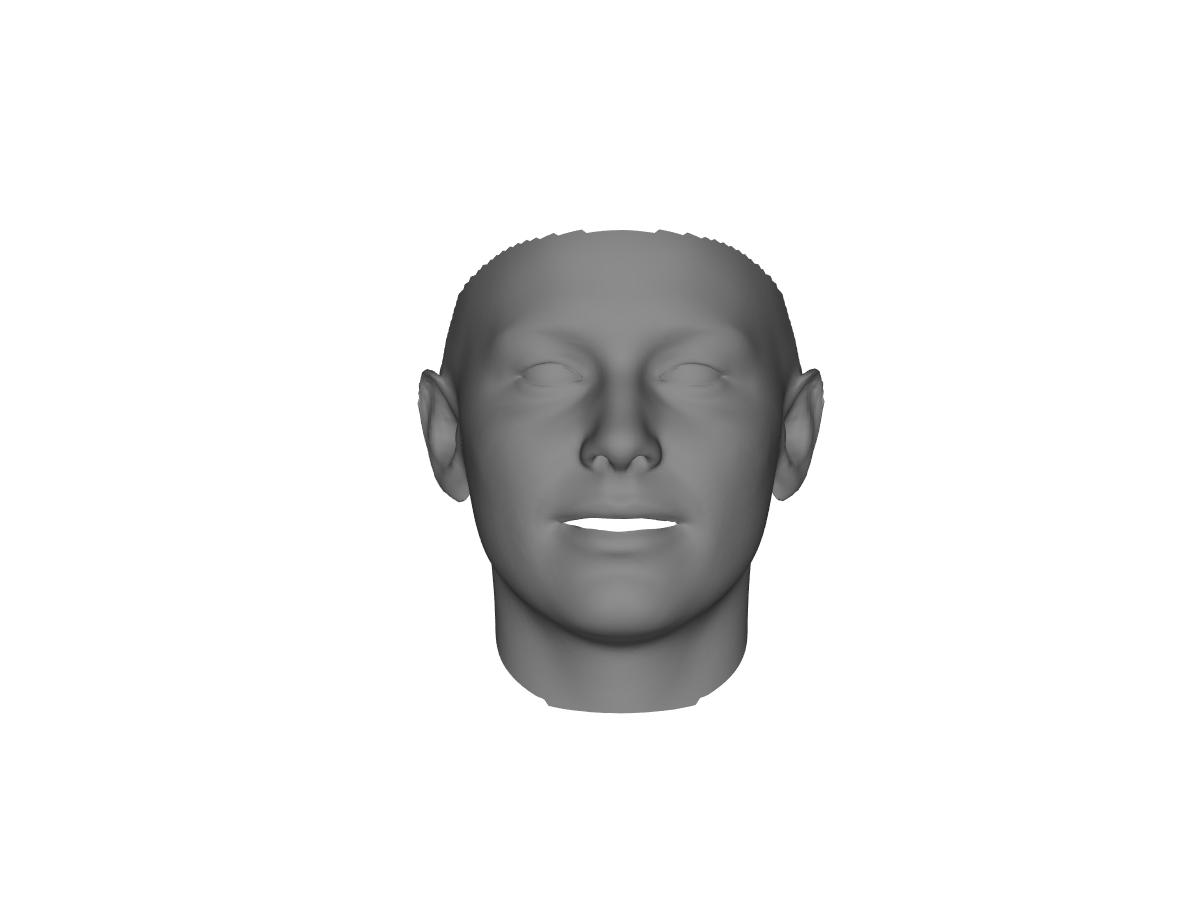} &
\\
\includegraphics[trim=400 180 370 230,clip,width=\VaryingShapeFigWid]{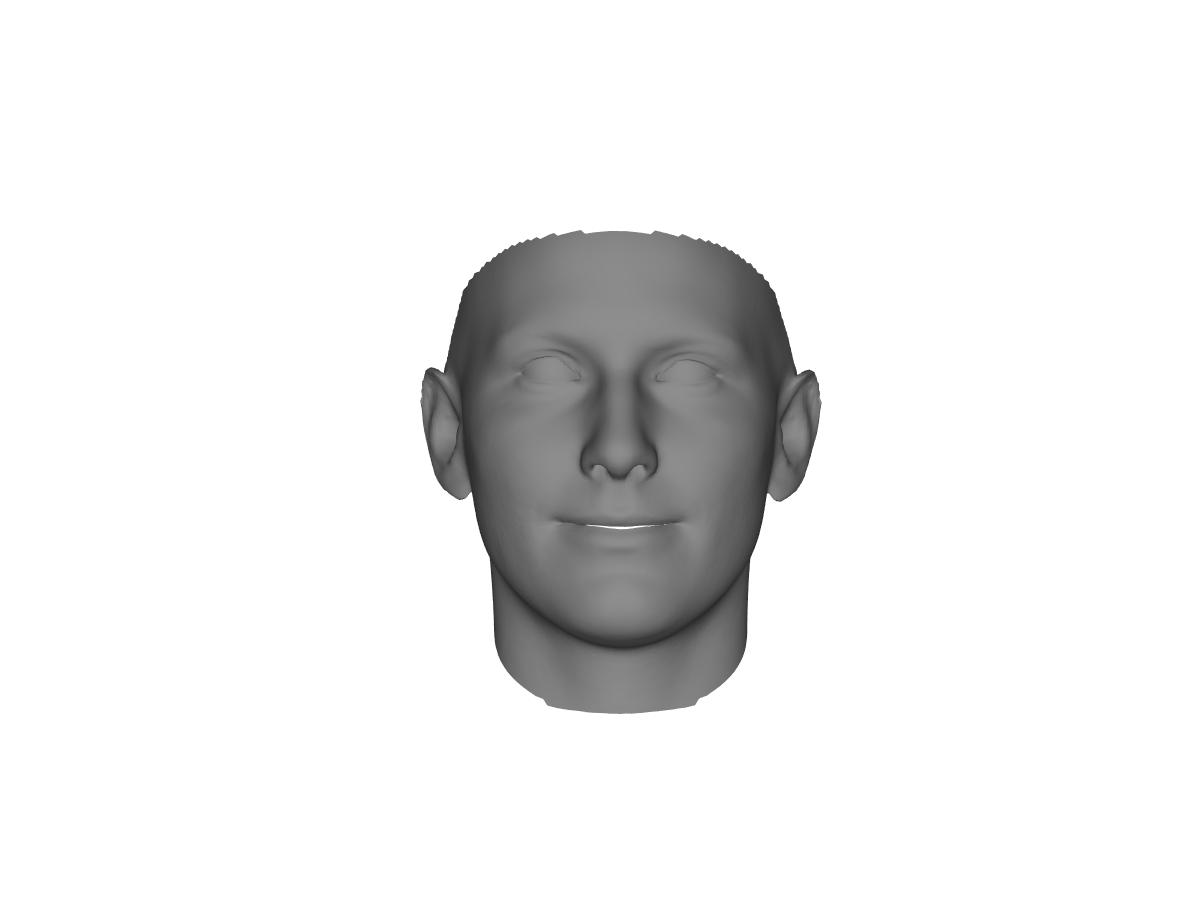} &
\includegraphics[trim=400 180 370 230,clip,width=\VaryingShapeFigWid]{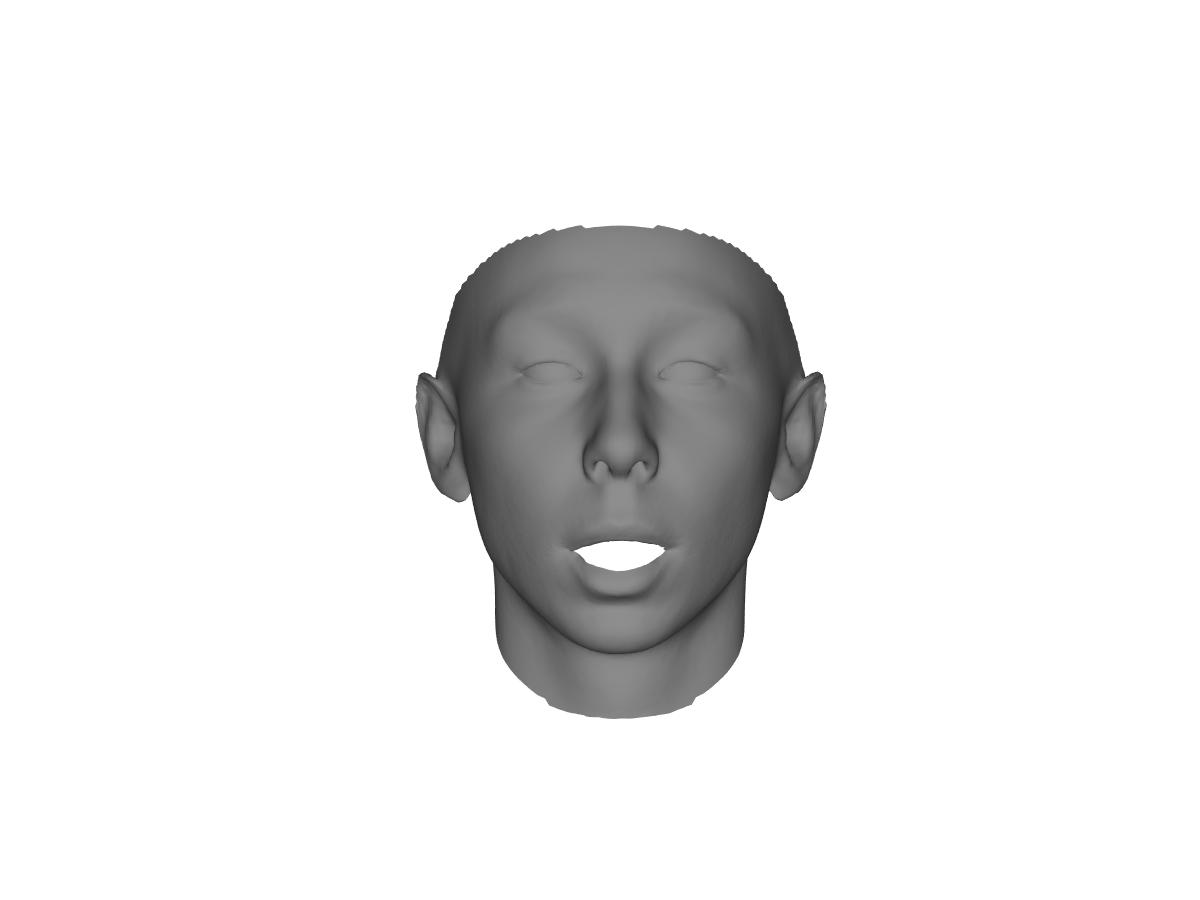} &
\includegraphics[trim=400 180 370 230,clip,width=\VaryingShapeFigWid]{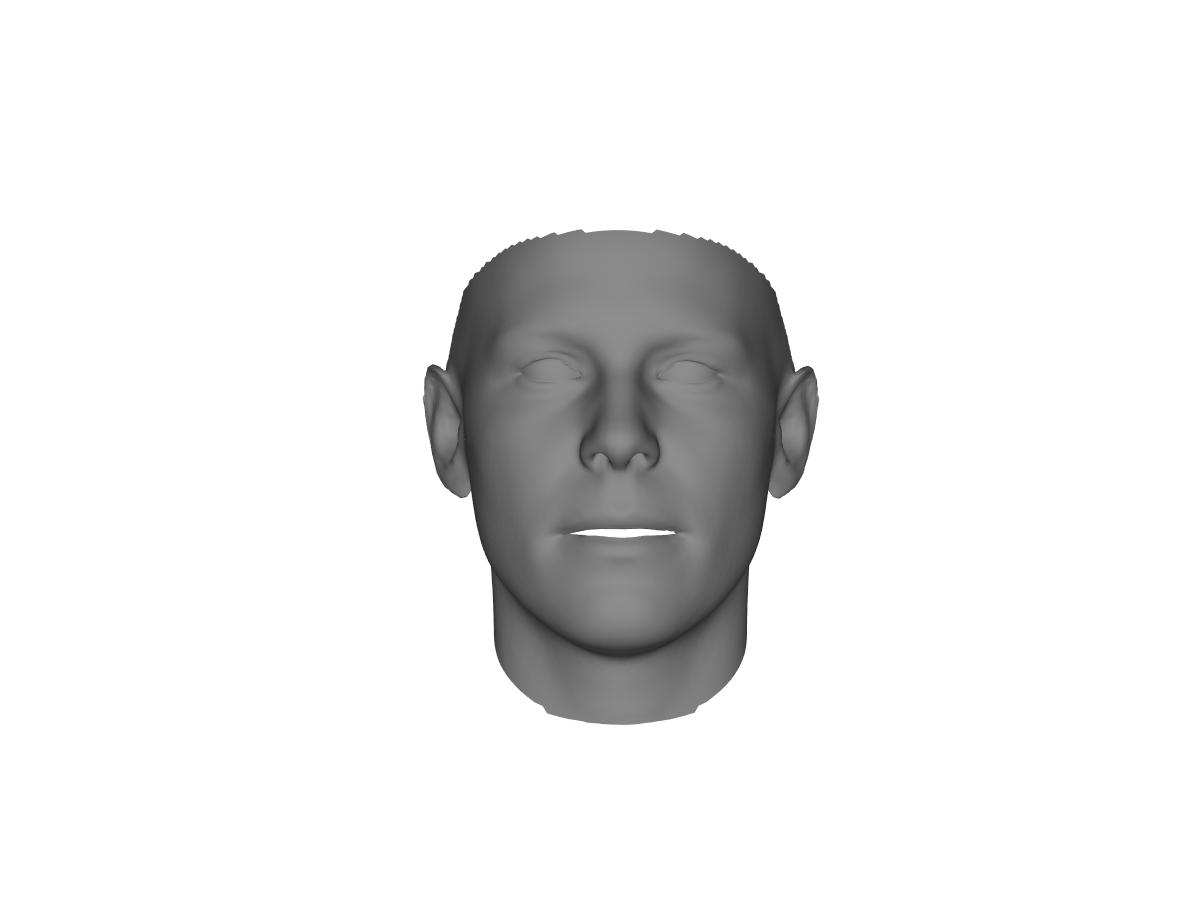} &
\includegraphics[trim=400 180 370 230,clip,width=\VaryingShapeFigWid]{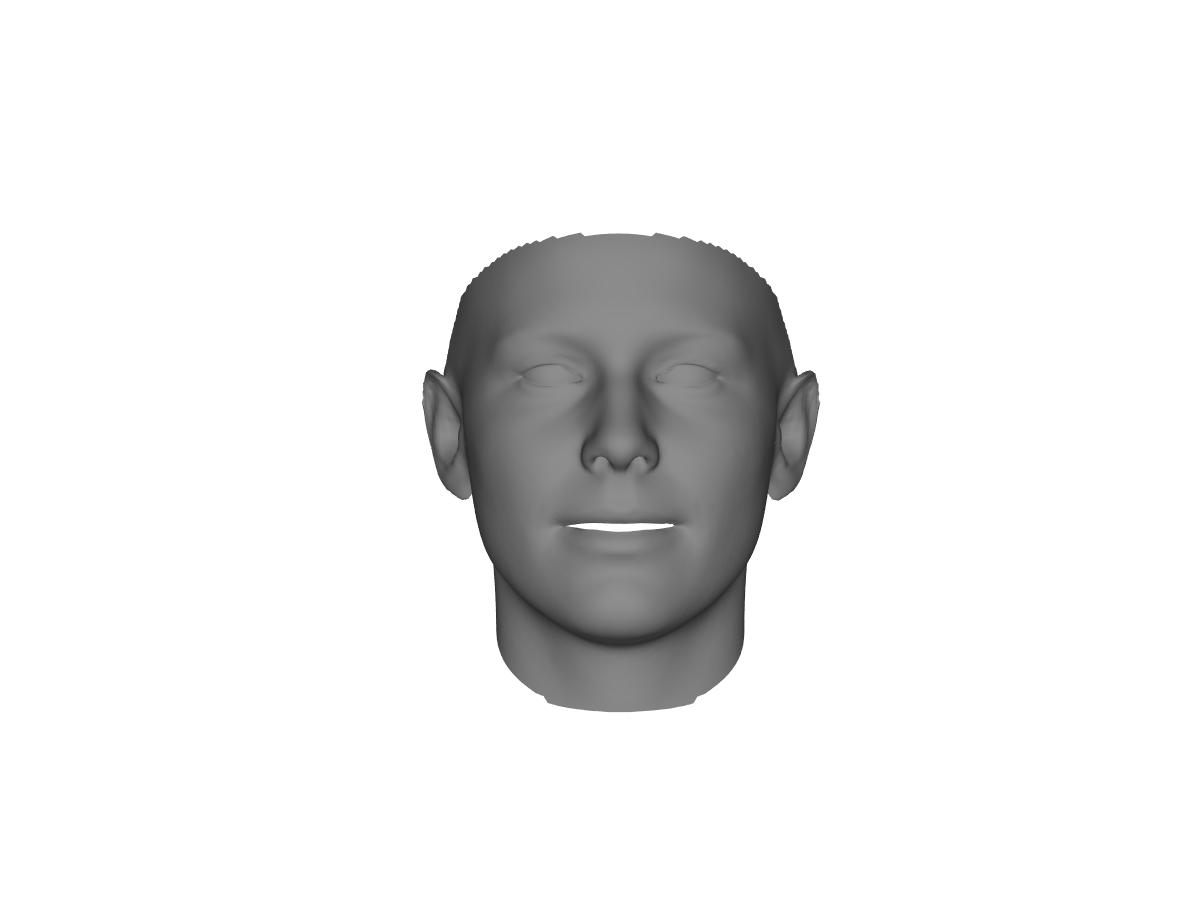} &
\end{tabular}
\caption{\small Each column shows shape changes when varying one element of $\mathbf{f}_S$, by $10$ times standard deviations, in opposite directions. Ordered by the magnitude of shape changes.}
\label{fig:varying_shape}\figvspace
\end{center}

\end{figure}

\begin{figure}[t!]
\begin{center}
\small
\setlength{\tabcolsep}{3pt}
\begin{tabular}{ @{}c@{}c@{}c@{}c@{}c@{}c@{}c@{}c@{}}
\includegraphics[width=\VaryingAlbFigWid]{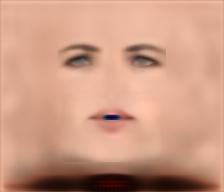} &
\includegraphics[width=\VaryingAlbFigWid]{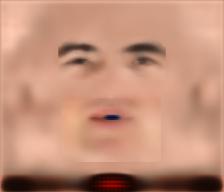} &
\includegraphics[width=\VaryingAlbFigWid]{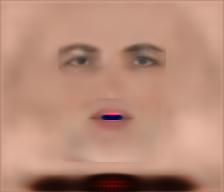} &
\includegraphics[width=\VaryingAlbFigWid]{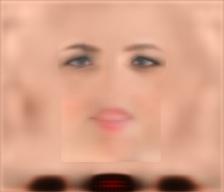} &
\\
\includegraphics[width=\VaryingAlbFigWid]{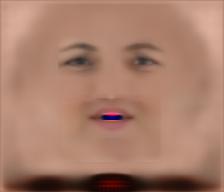} &
\includegraphics[width=\VaryingAlbFigWid]{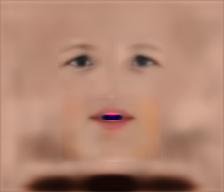} &
\includegraphics[width=\VaryingAlbFigWid]{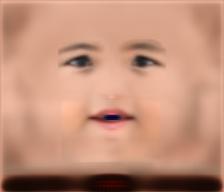} &
\includegraphics[width=\VaryingAlbFigWid]{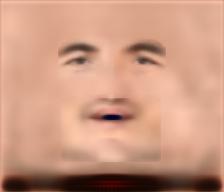} &
\end{tabular}
\caption{\small Each column shows albedo changes when varying one element of $\mathbf{f}_A$ in opposite directions.}
\label{fig:varying_tex}\figvspace
\end{center}
\end{figure}

\SubSection{Expressiveness}
\label{sec:expressiveness}
\Paragraph{Exploring feature space}
We feed the entire CelebA dataset~\cite{liu2015faceattributes} with ${\sim}200$k images to our network to obtain the empirical distribution of our shape and texture parameters. 
By varying the mean parameter along each dimension proportional to its standard deviation, we can get a sense how each element contribute to the final shape and texture.
We sort elements in the shape parameter $\mathbf{f}_S$ based on their differences to the mean $3$D shape. 
Fig.~\ref{fig:varying_shape} shows four examples of shape changes, whose differences rank No.$1$, $40$, $80$, and $120$ among $160$ elements.
Most of top changes are expression related.
Similarly, in Fig.~\ref{fig:varying_tex}, we visualize different texture changes by adjusting only one element of $\mathbf{f}_A$ off the mean parameter $\bar{\mathbf{f}}_A$.
The elements with the same $4$ ranks are selected. 

\begin{figure}[t!]
\begin{center}
\footnotesize
\setlength{\tabcolsep}{1pt}
\begin{tabular}{ ccccccccc}
 Male & Mustache & Bags Under Eyes & Old \\ [-3pt]
 \includegraphics[trim=42 5 32 7, clip, width=\MeaningfulBasisFigWid]{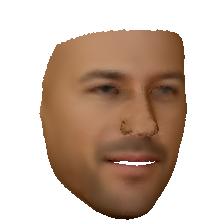} &
\includegraphics[trim=42 5 32 27, clip, width=\MeaningfulBasisFigWid]{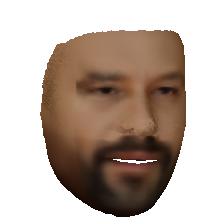} &
\includegraphics[trim=42 5 32 27, clip, width=\MeaningfulBasisFigWid]{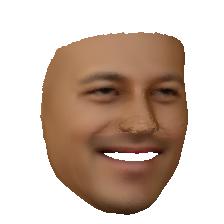} &
\includegraphics[trim=42 5 32 27, clip, width=\MeaningfulBasisFigWid]{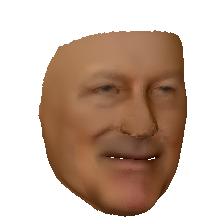} &
\\ [-3pt]
\includegraphics[trim=42 5 32 27, clip, width=\MeaningfulBasisFigWid]{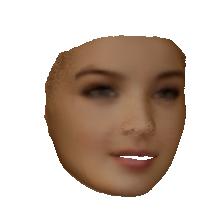} &
\includegraphics[trim=42 5 32 27, clip, width=\MeaningfulBasisFigWid]{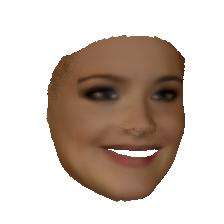} &
\includegraphics[trim=42 5 32 27, clip, width=\MeaningfulBasisFigWid]{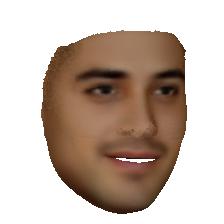} &
\includegraphics[trim=42 5 32 27, clip, width=\MeaningfulBasisFigWid]{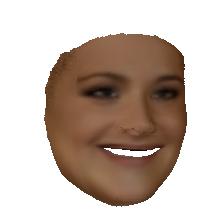} &
\\
Female & Rosy Cheeks & Bushy Eyebrows & Smiling 
\end{tabular}
\vspace{-2mm}
\caption{\small Nonlinear $3$DMM generates shape and albedo embedded with different attributes.}
\label{fig:meaningful_basis}
\figvspace 
\end{center}
\end{figure}

\Paragraph{Attribute Embedding}
To better understand different shape and albedo instances embedded in our two decoders, we dig into their attribute meaning.
For a given attribute, e.g., male, we feed images with that attribute $\{\mathbf{I}_i\}_{i=1}^n$ into our encoder $E$ to obtain two sets of parameters $\{\mathbf{f}_S^{i}\}_{i=1}^n$ and $\{\mathbf{f}_A^{i}\}_{i=1}^n$. 
These sets represent corresponding empirical distributions of the data in the low dimensional spaces. 
Computing the mean parameters $\mathbf{\bar{f}}_S, \mathbf{\bar{f}}_A$ and feed into their respective decoders, also using the mean lighting parameter, we can reconstruct the mean shape and texture with that attribute. 
Fig.~\ref{fig:meaningful_basis} visualizes the reconstructed textured $3$D mesh related to some attributes.
Differences among attributes present in both shape and texture.
Here we can observe the power of our nonlinear $3$DMM to model small details such as ``bag under eyes", or ``rosy cheeks", etc.

\begin{figure}[t!]
\begin{center}
\small
\setlength{\tabcolsep}{3pt}
\begin{tabular}{ @{}c@{\hskip 1.mm}c@{\hskip 1.mm}c@{\hskip 1.mm}c@{\hskip 1.mm}c@{}c@{}}
\multirow{2}{*}{Input} & \multirow{2}{*}{Linear} & \multicolumn{2}{c}{Nonlinear} \\ & & Grad desc & Network \\
\includegraphics[width=\TexRepFigWid]{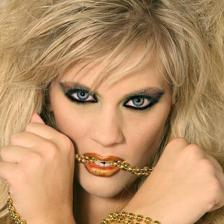} &
\includegraphics[width=\TexRepFigWid]{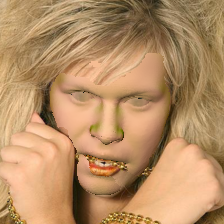} &
\includegraphics[width=\TexRepFigWid]{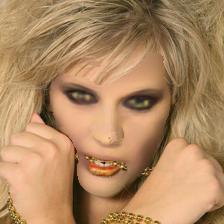} &
\includegraphics[width=\TexRepFigWid]{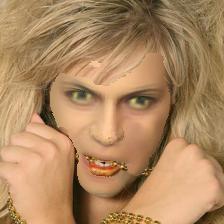} &
\\
\includegraphics[width=\TexRepFigWid]{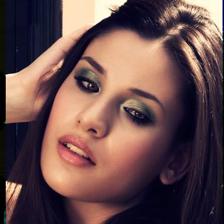} &
\includegraphics[width=\TexRepFigWid]{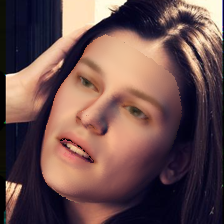} &
\includegraphics[width=\TexRepFigWid]{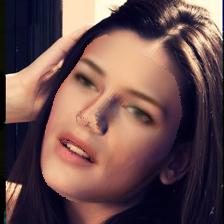} &
\includegraphics[width=\TexRepFigWid]{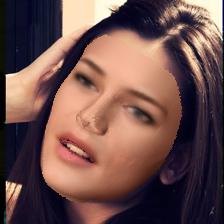} &
\\
\includegraphics[width=\TexRepFigWid]{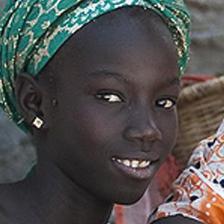} &
\includegraphics[width=\TexRepFigWid]{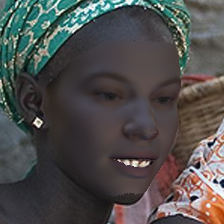} &
\includegraphics[width=\TexRepFigWid]{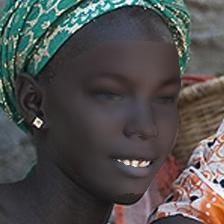} &
\includegraphics[width=\TexRepFigWid]{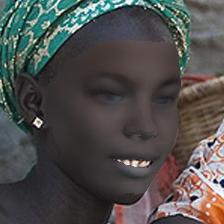} &
\\
\includegraphics[width=\TexRepFigWid]{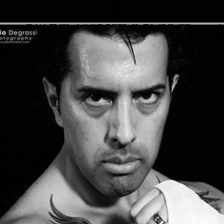} &
\includegraphics[width=\TexRepFigWid]{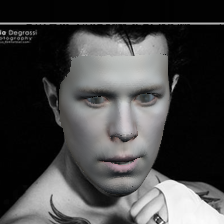} &
\includegraphics[width=\TexRepFigWid]{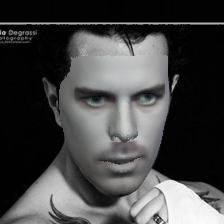} &
\includegraphics[width=\TexRepFigWid]{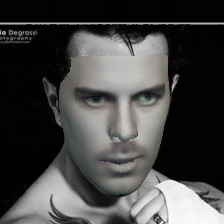} &
\end{tabular}
\caption{\small Texture representation power comparison. Our nonlinear model can better reconstruct the facial texture.}
\label{fig:tex_representationpower}\figvspace
\end{center}
\end{figure}


\SubSection{Representation Power}

We compare the representation power of the proposed nonlinear $3$DMM vs.~traditional linear $3$DMM.

\Paragraph{Albedo}
Given a face image, assuming we know the groundtruth shape and projection parameters, we can unwarp the texture into the UV space, as we generate ``pseudo groundtruth" texture in the weakly supervision step. 
With the groundtruth texture, by using gradient descent, we can jointly estimate, a lighting parameter $\mathbf{L}$ and an albedo parameter  $\mathbf{f}_A$ whose decoded texture matches with the groundtruth. 
Alternatively, we can minimize the reconstruction error in the image space, through the rendering layer with the groundtruth $\mathbf{S}$ and $\mathbf{m}$. 
Empirically, two methods give similar performances but we choose the first option as it involves only one warping step, instead of doing rendering in every optimization iteration.
For the linear model, we use albedo bases of Basel Face Model (BFM)~\cite{paysan20093d}. 
As in Fig.~\ref{fig:tex_representationpower}, our nonlinear texture is closer to the groundtruth than the linear model. 
This is expected since the linear model is trained with controlled images. 
Quantitatively, our nonlinear model has significantly lower averaged $L_1$ reconstruction error than the linear model ($0.053$ vs.~$0.062$).

\begin{figure}[t!]
\begin{center}
\small
\setlength{\tabcolsep}{3pt}
\begin{tabular}{ @{}c@{}c@{}c@{}c@{}cc}
$3$D Scan & \multicolumn{2}{c}{Nonlinear} & \multicolumn{2}{c}{Linear} \\
\includegraphics[trim =  520 170 520 150, clip, width=\ShapeRepFigWid]{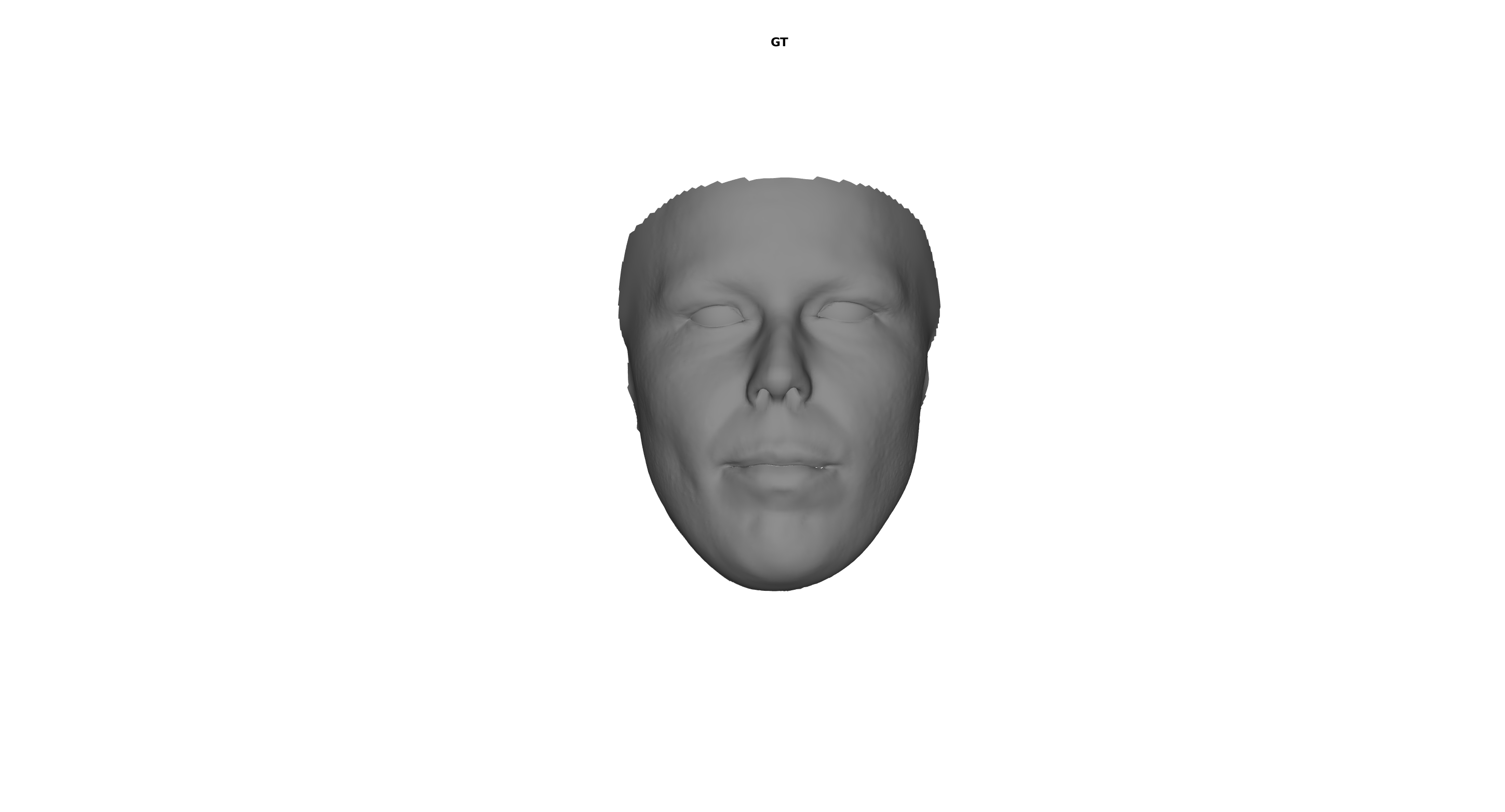} &
\includegraphics[trim =  520 170 520 150, clip, width=\ShapeRepFigWid]{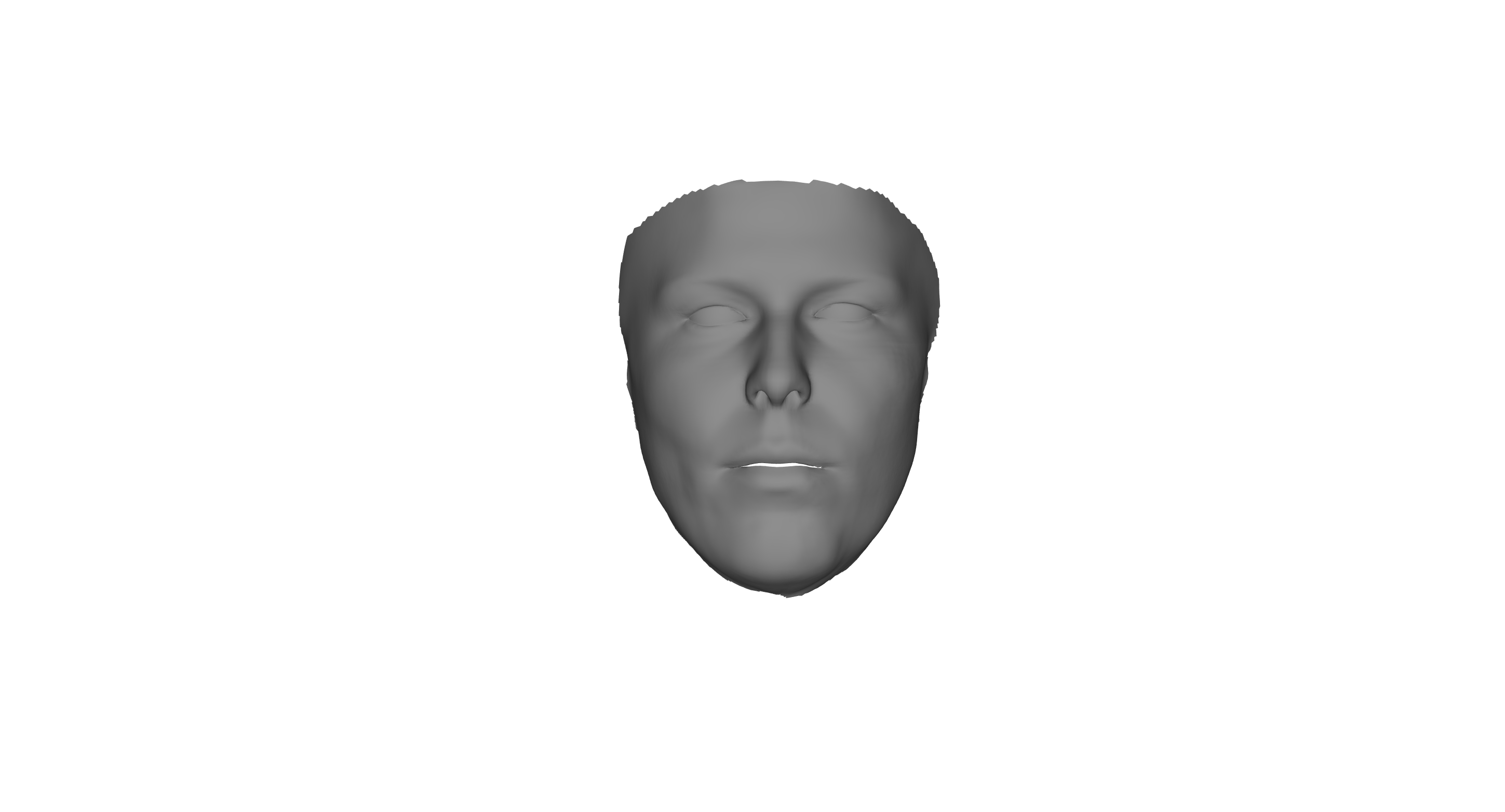} &
\includegraphics[trim =  520 170 520 150, clip, width=\ShapeRepFigWid]{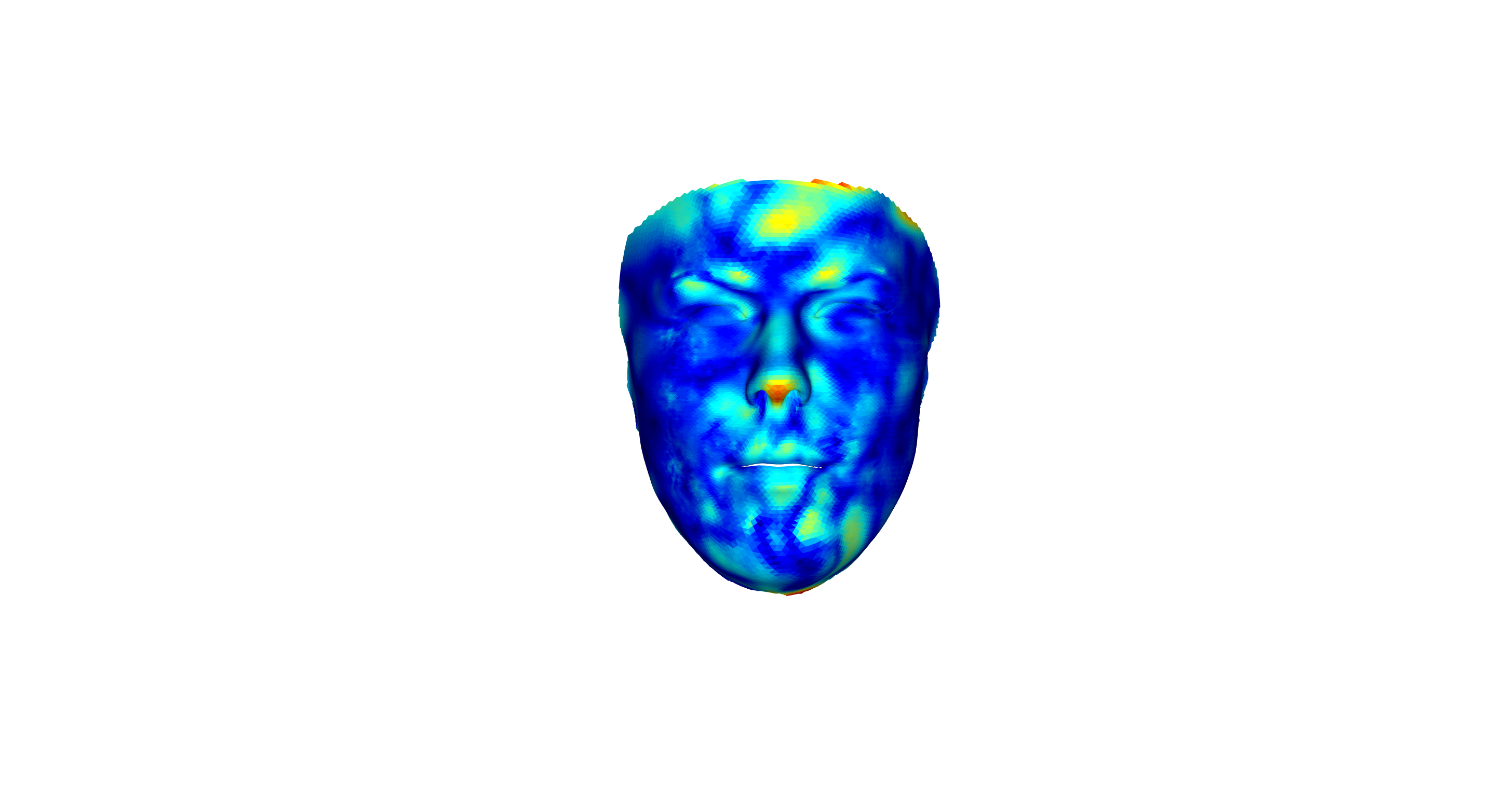} &
\includegraphics[trim =  186 100 186 70, clip, width=\ShapeRepFigWid]{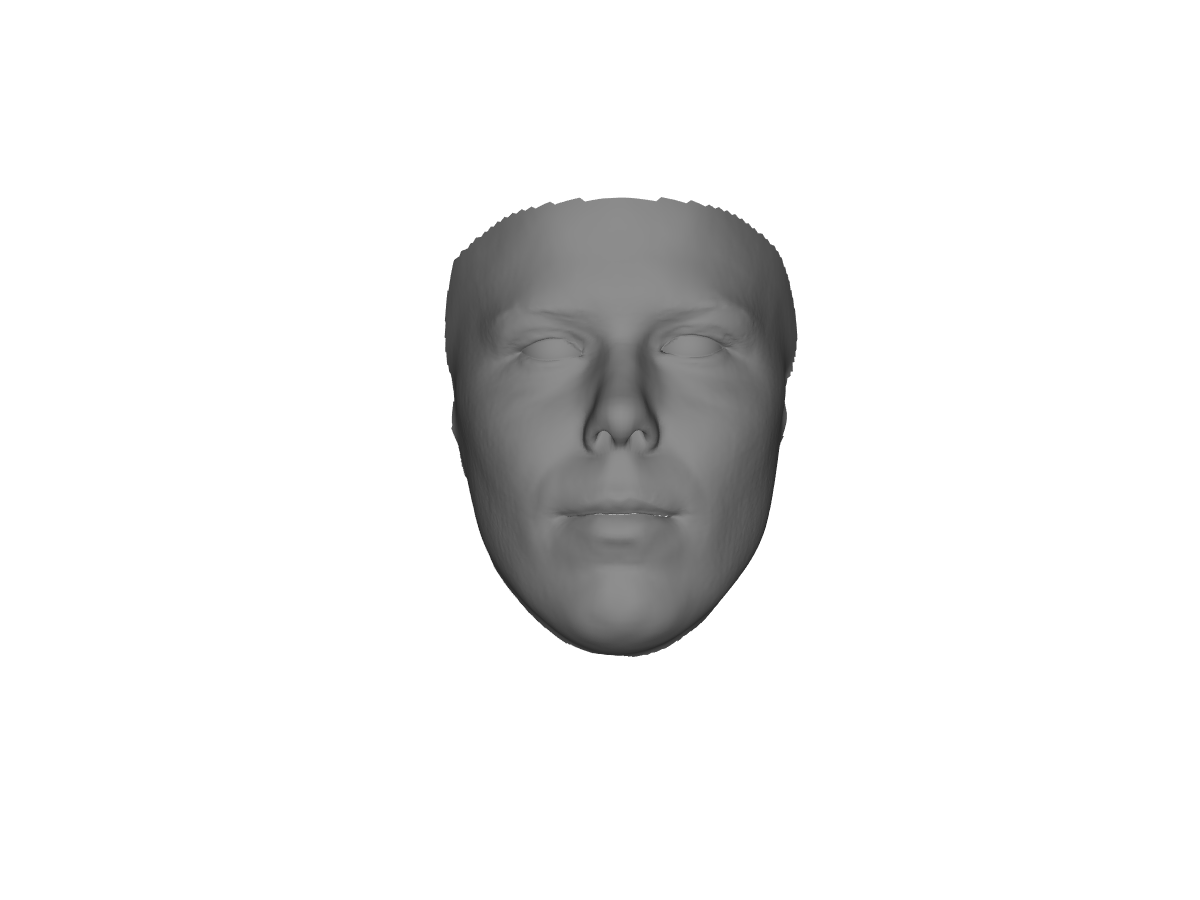}  &
\includegraphics[trim =  520 170 520 150, clip, width=\ShapeRepFigWid]{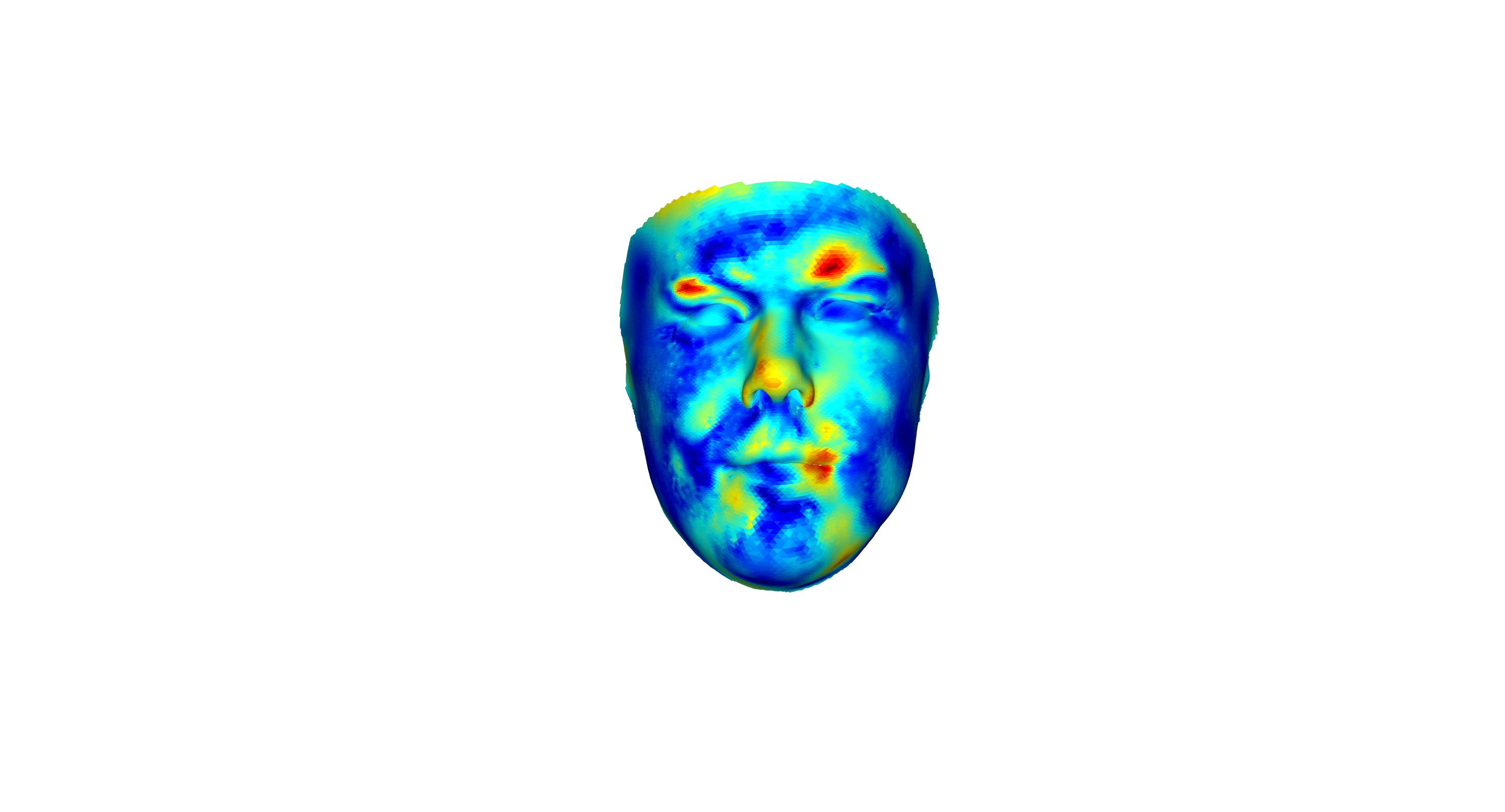}  \vspace{-1mm} 
&
\includegraphics[trim = 450 300 100 300, clip, width=0.035\textwidth]{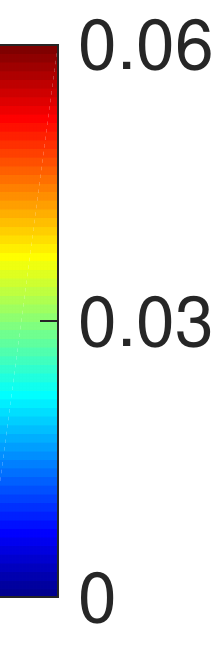}
\end{tabular}
\caption{\small Shape representation power comparison on Basel scans. Our nonlinear model is able to reconstruct input $3$D scans with smaller errors than the linear model ($l_S = 160$ for both models). The error map shows the normalized per-vertex errors. }
\label{fig:shape_representation}\figvspace
\end{center}
\end{figure}

\begin{figure}[t!]
\begin{center}
\small
\includegraphics[trim = 10 10 10 10, clip, width=\linewidth]{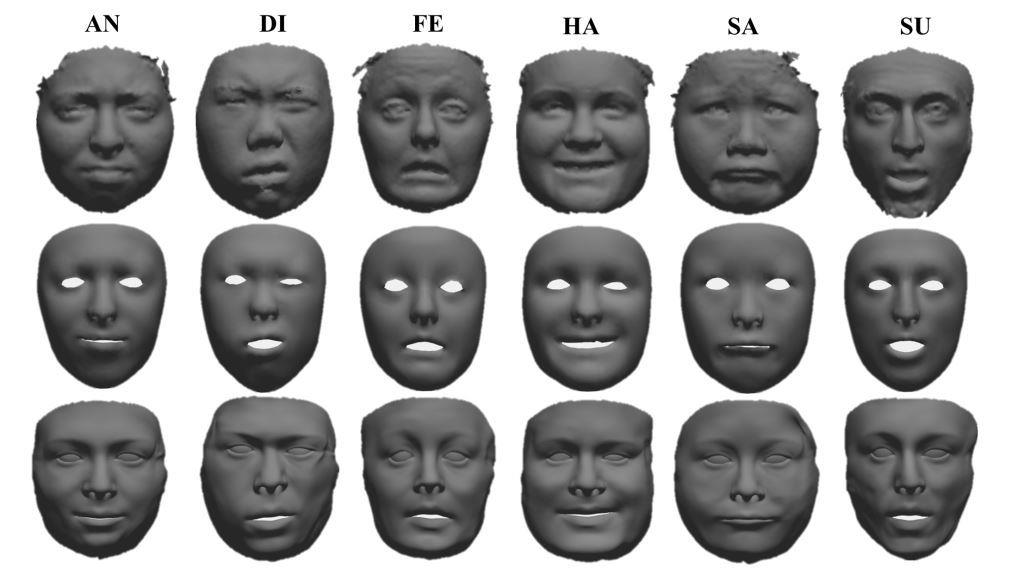}
\caption{\small Shape representation power comparison on BU-3DFE. First row: 3D scans with expressions: angry~(AN), disgust~(DI), fear~(FE), happy~(HA), sad~(SA), and surprise~(SU); second row: FaceWarehouse bilinear model~\cite{cao2014facewarehouse}'s reconstructions, third row: our reconstructions.}
\label{fig:shape_representation_BU3DFE}\figvspace
\end{center}
\end{figure}

\begin{figure*}[t!]
\begin{center}
\small
\setlength{\tabcolsep}{3pt}
\begin{tabular}{ @{}c@{\hskip 1.mm}c@{}c@{}c@{}c@{}c@{\hskip 3mm}c@{\hskip 1.mm}c@{}c@{}c@{}c@{}c@{}}
Input & Overlay & Albedo & Shape & Shading && Input & Overlay & Albedo & Shape & Shading \\

\includegraphics[width=\FittingFigWid]{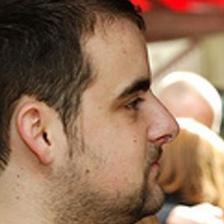} &
\includegraphics[width=\FittingFigWid]{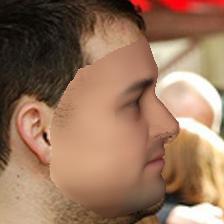} &
\includegraphics[width=\FittingFigWid]{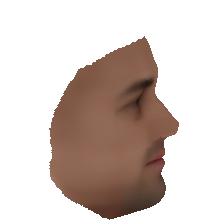} &
\includegraphics[width=\FittingFigWid]{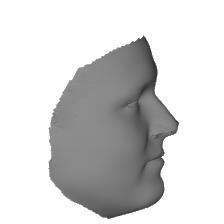} &
\includegraphics[width=\FittingFigWid]{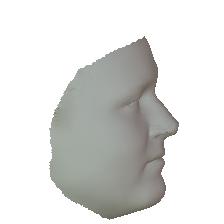} &
&
\includegraphics[width=\FittingFigWid]{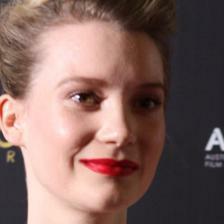} &
\includegraphics[width=\FittingFigWid]{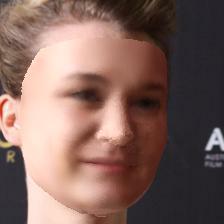} &
\includegraphics[width=\FittingFigWid]{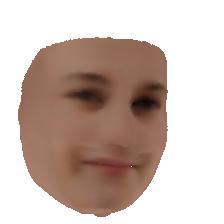} &
\includegraphics[width=\FittingFigWid]{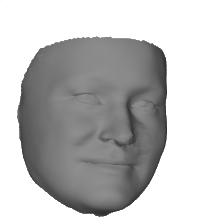} &
\includegraphics[width=\FittingFigWid]{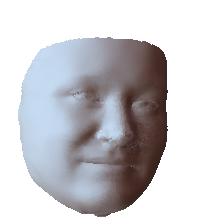} &
\\

\includegraphics[width=\FittingFigWid]{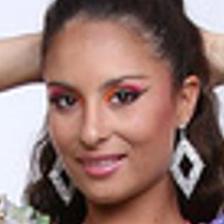} &
\includegraphics[width=\FittingFigWid]{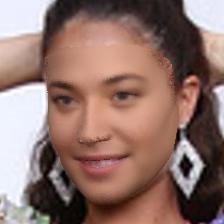} &
\includegraphics[width=\FittingFigWid]{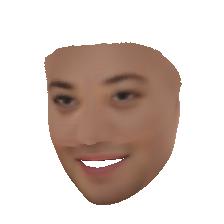} &
\includegraphics[width=\FittingFigWid]{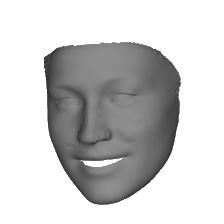} &
\includegraphics[width=\FittingFigWid]{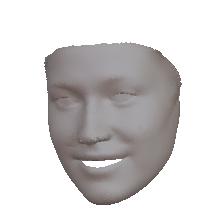} &
&
\includegraphics[width=\FittingFigWid]{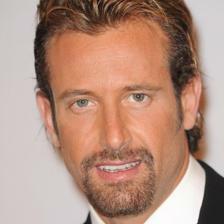} &
\includegraphics[width=\FittingFigWid]{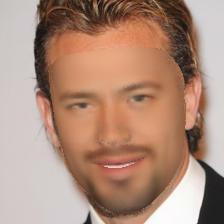} &
\includegraphics[width=\FittingFigWid]{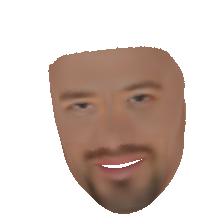} &
\includegraphics[width=\FittingFigWid]{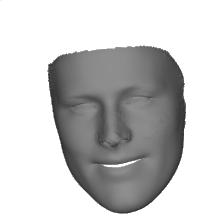} &
\includegraphics[width=\FittingFigWid]{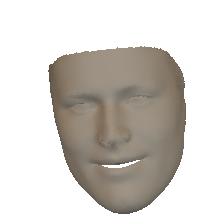} &
\\

\includegraphics[width=\FittingFigWid]{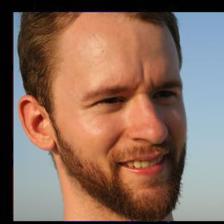} &
\includegraphics[width=\FittingFigWid]{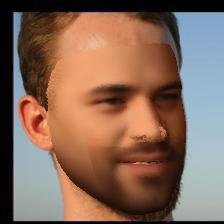} &
\includegraphics[width=\FittingFigWid]{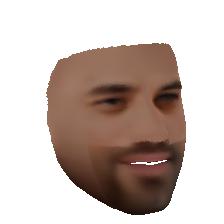} &
\includegraphics[width=\FittingFigWid]{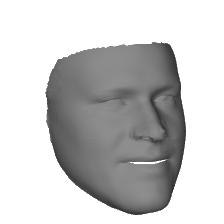} &
\includegraphics[width=\FittingFigWid]{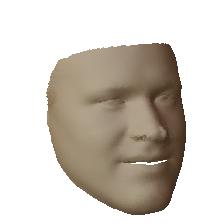} &
&
\includegraphics[width=\FittingFigWid]{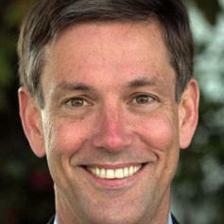} &
\includegraphics[width=\FittingFigWid]{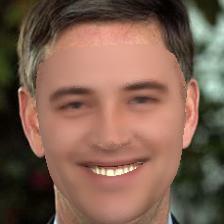} &
\includegraphics[width=\FittingFigWid]{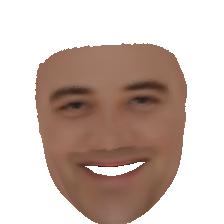} &
\includegraphics[width=\FittingFigWid]{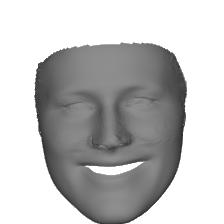} &
\includegraphics[width=\FittingFigWid]{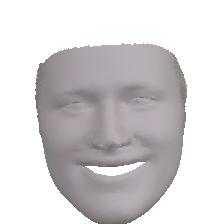} &
\\

\includegraphics[width=\FittingFigWid]{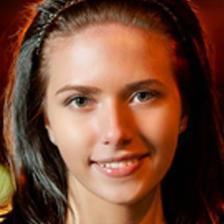} &
\includegraphics[width=\FittingFigWid]{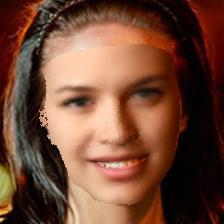} &
\includegraphics[width=\FittingFigWid]{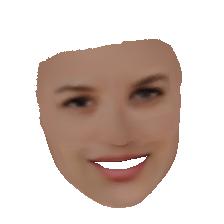} &
\includegraphics[width=\FittingFigWid]{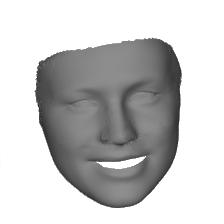} &
\includegraphics[width=\FittingFigWid]{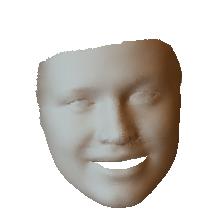} &
&
\includegraphics[width=\FittingFigWid]{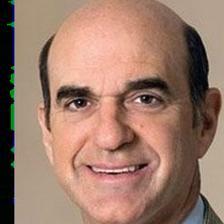} &
\includegraphics[width=\FittingFigWid]{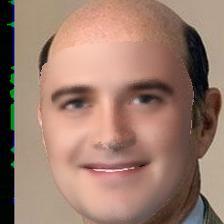} &
\includegraphics[width=\FittingFigWid]{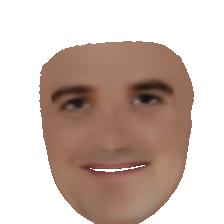} &
\includegraphics[width=\FittingFigWid]{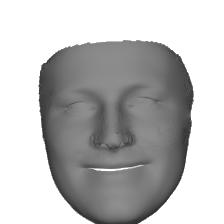} &
\includegraphics[width=\FittingFigWid]{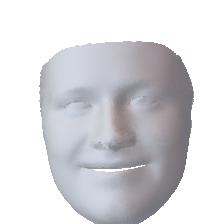} &
\\


\includegraphics[width=\FittingFigWid]{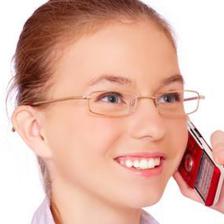} &
\includegraphics[width=\FittingFigWid]{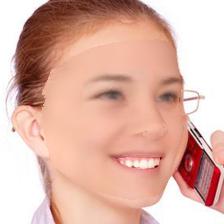} &
\includegraphics[width=\FittingFigWid]{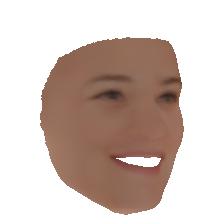} &
\includegraphics[width=\FittingFigWid]{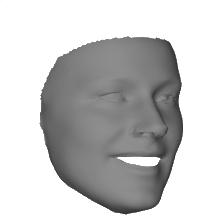} &
\includegraphics[width=\FittingFigWid]{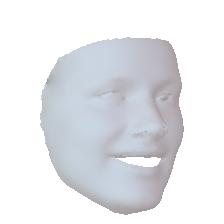} &
&
\includegraphics[width=\FittingFigWid]{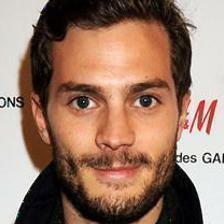} &
\includegraphics[width=\FittingFigWid]{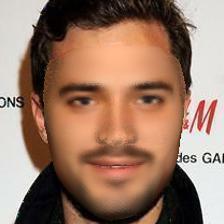} &
\includegraphics[width=\FittingFigWid]{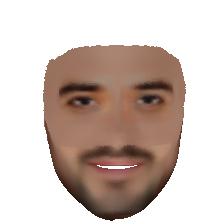} &
\includegraphics[width=\FittingFigWid]{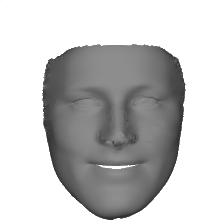} &
\includegraphics[width=\FittingFigWid]{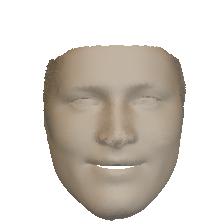} &

\end{tabular}
\caption{\small $3$DMM fits to faces with diverse skin color, pose, expression, lighting, facial hair, and faithfully recovers these cues. Left half shows results from AFLW2000 dataset, right half shows results from CelebA.}
\label{fig:3dmm_fitting}\figvspace \vspace{-1mm}

\end{center}
\end{figure*}

\begin{table}[t!]
\footnotesize
\caption{\small{$3$D scan reconstruction comparison (NME).}} 
\label{tab:shape_representation_number}
\tabvspace
\begin{center}
\begin{tabular}{ lccccccc}
\toprule 
$l_S$     & $40$ & $80$ & $160$ \\ \midrule
Linear    & $0.0321$ & $0.0279$ & $0.0241$ \\
Nonlinear\cite{tran2018nonlinear} & $0.0277$ & $0.0236$ & $0.0196$ \\
Nonlinear & $0.0268$ & $0.0214$ & $\mathbf{0.0146}$ \\

\bottomrule
\end{tabular}
\end{center}
\figvspace
\end{table}

\Paragraph{3D Shape}
We also compare the power of nonlinear and linear $3$DMMs in representing real-world $3$D scans. 
We compare with BFM~\cite{paysan20093d}, the most commonly used $3$DMM at present. 
We use ten $3$D face scans provided by~\cite{paysan20093d}, which are not included in the training set of BFM.
As these face meshes are already registered using the same triangle definition with BFM,  no registration is necessary.  
Given the groundtruth shape, by using gradient descent, we can estimate a shape parameter whose decoded shape matches the groundtruth. 
We define matching criterion on both vertex distances and surface normal direction. 
This empirically improves fidelity of final results compared to only optimizing vertex distances.
Fig.~\ref{fig:shape_representation} shows the visual quality of two models' reconstruction.
Our reconstructions closely match the face shapes details. 
%
To quantify the difference, we use NME, averaged per-vertex errors between the recovered and groundtruth shapes, normalized by inter-ocular distances. 
Our nonlinear model has a significantly smaller reconstruction error than the linear model, $0.0146$ vs.~$0.0241$ (Tab.~\ref{tab:shape_representation_number}). 

Besides, to evaluate our model power to represent different facial expressions, we make a similar comparison on 3D scans with different expression from BU-3DFE dataset~\cite{yin20063d}. Here we compare with Facewarehouse bilinear model~\cite{cao2014facewarehouse}, which is directly learn from $3$D scans with various facial expressions. Due to the difference in mesh topology, here we try to optimize the Chamfer distance~\cite{fan2017point} between the ground-truth shape and our estimation. Fig.~\ref{fig:shape_representation_BU3DFE} show qualitatively comparisons on scans with different expressions. Our model has comparable performance on capturing the facial expression, while being better on resembling facial details. This is reflected on the averaged Chamfer distance of all $2500$ scans in the dataset ($0.00083$ v.s $0.00197$ for our model and FaceWarehouse model respectively).

\SubSection{Applications}
Having shown the capability of our nonlinear $3$DMM (i.e., two decoders), now we demonstrate the applications of our entire network, which has the additional encoder.
Many applications of $3$DMM are centered on its ability to fit to $2$D face images.
Similar to linear $3$DMM, our nonlinear $3$DMM can be utilized for model fitting, which decomposes a $2$D face into its shape, albedo and lighting.
Fig.~\ref{fig:3dmm_fitting} visualizes our $3$DMM fitting results on AFLW2000 and CelebA dataset. 
Our encoder estimates the shape $\mathbf{S}$, albedo $\mathbf{A}$ as well as lighting $\mathbf{L}$ and projection parameter $\mathbf{m}$. 
We can recover personal facial characteristic in both shape and albedo. 
Our albedo can present facial hair, which is normally hard to be recovered by linear $3$DMM.

\SubSubSection{Face Alignment}

\begin{figure}[t!]
\begin{center}
\small
\setlength{\tabcolsep}{3pt}
\begin{tabular}{ @{\hskip 0.5mm}c@{\hskip 0.5mm}c@{\hskip 0.5mm}c@{\hskip 0.5mm}c@{\hskip 0.5mm}c@{\hskip 0.5mm}c@{\hskip 0.5mm}c@{\hskip 1.5mm}c@{\hskip 1.5mm}c@{\hskip 1.5mm}c@{}}
\\
\includegraphics[width=\AlignFigWid]{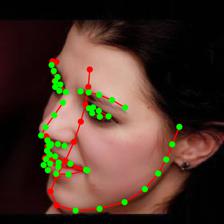} &
\includegraphics[width=\AlignFigWid]{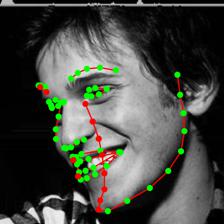} &
\includegraphics[width=\AlignFigWid]{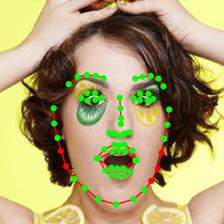} &
\includegraphics[width=\AlignFigWid]{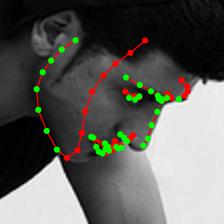} &
\includegraphics[width=\AlignFigWid]{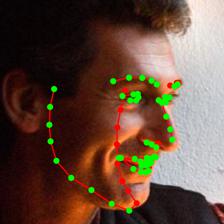} &
\\
\includegraphics[width=\AlignFigWid]{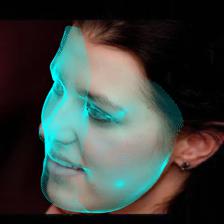} &
\includegraphics[width=\AlignFigWid]{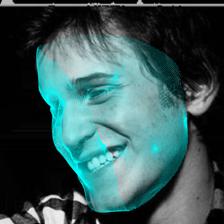} &
\includegraphics[width=\AlignFigWid]{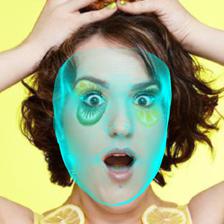} &
\includegraphics[width=\AlignFigWid]{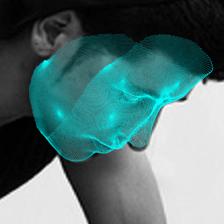} &
\includegraphics[width=\AlignFigWid]{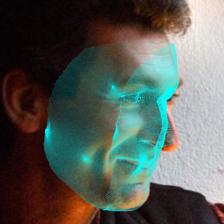} &
\end{tabular}
\vspace{-2mm}
\caption{\small Our face alignment results. Invisible landmarks are marked as red. We can well handle extreme pose, lighting and expression.}
\label{fig:2d_align}\figvspace
\end{center}
\end{figure}

Face alignment is a critical step for many facial analysis tasks such as face recognition~\cite{tran2017disentangled,tran2018representation}. 
With enhancement in the modeling, we hope to improve this task (Fig.~\ref{fig:2d_align}). 
We compare face alignment performance with state-of-the-art methods,
$3$DDFA~\cite{zhu2016face}, DeFA~\cite{liu2017dense}, $3$D-FAN~\cite{bulat2017far} and PRN~\cite{feng2018joint}, on AFLW2000 dataset on both $3$D and $3$D settings.

\begin{figure}[t!]
\begin{center}
\begin{tabular}{@{}c@{\hskip 1mm}c@{}}
\includegraphics[trim=90 215 115 220, clip, width=0.47\linewidth]{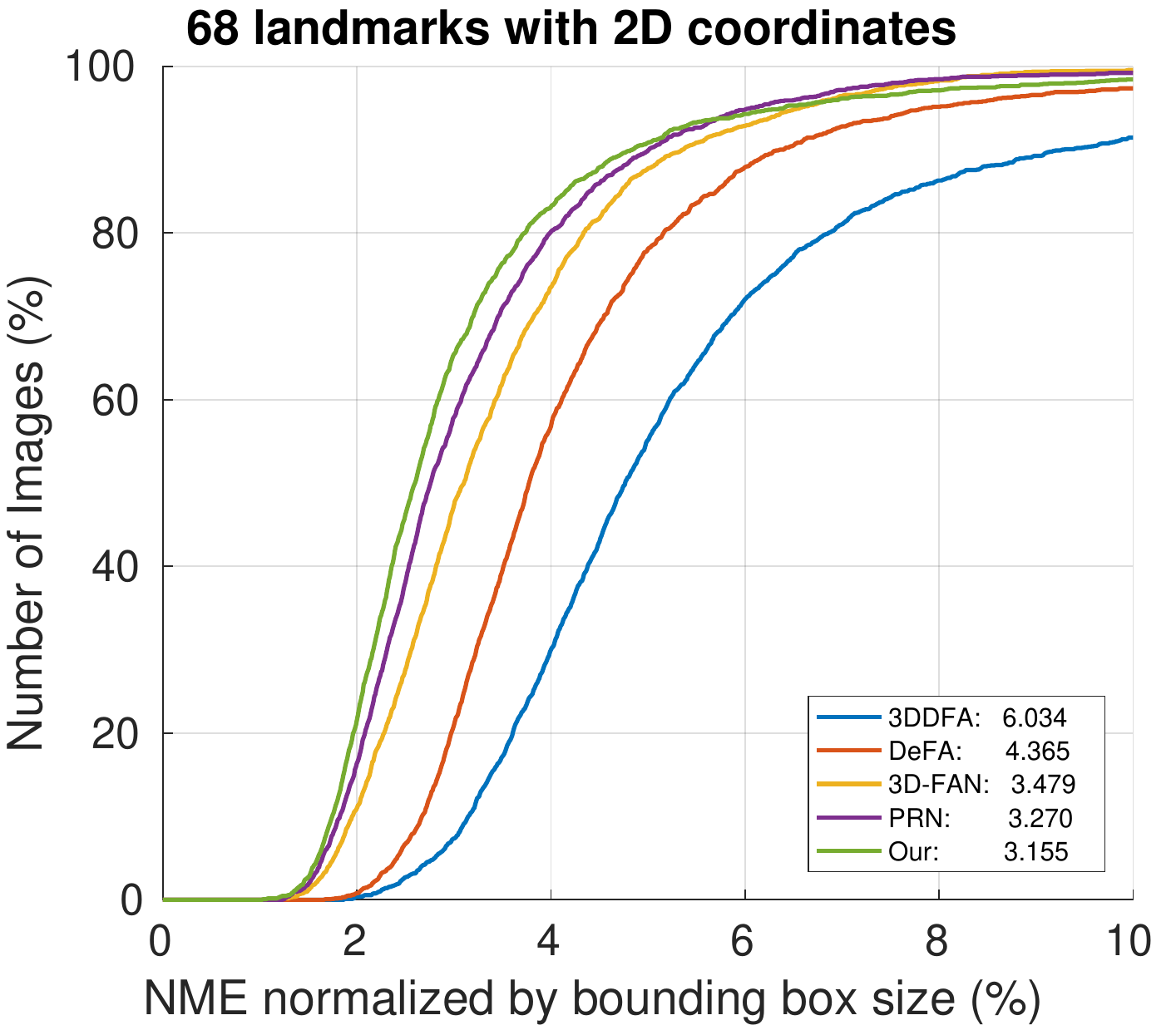} & 
\includegraphics[trim=90 215 115 220, clip, width=0.47\linewidth]{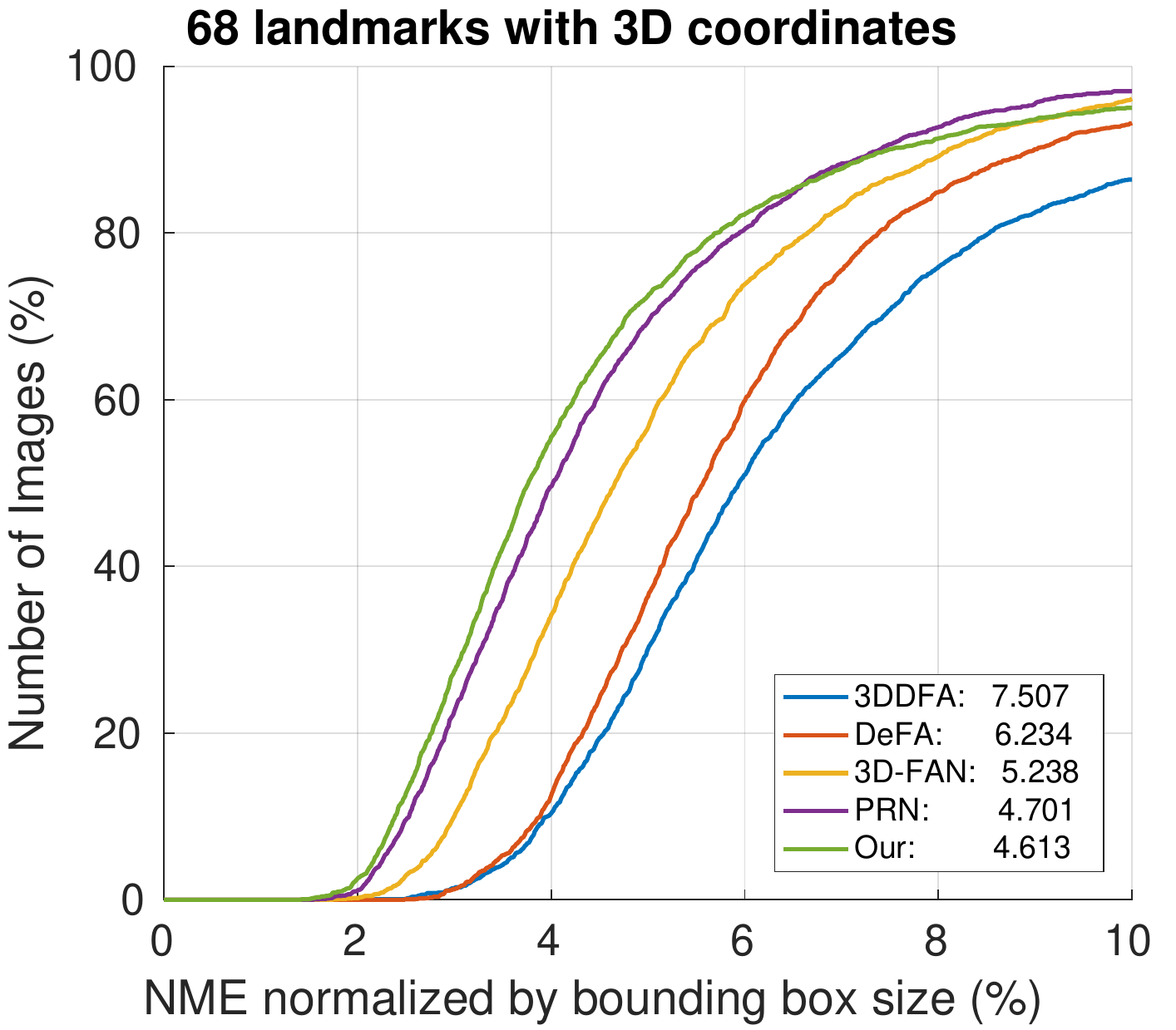}
\end{tabular}
\vspace{-2mm}
\caption{\small Face alignment Cumulative Errors Distribution (CED) curves on AFLW2000-3D on $2$D (left) and $3$D landmarks (right). NMEs are shown in legend boxes.}
\label{fig:face_align_CED}\figvspace \vspace{-1mm}
\end{center}
\end{figure}

The accuracy is evaluated using Normalized Mean Error (NME) as the evaluation metric with bounding box size as the normalization factor~\cite{bulat2017far}.
For fair comparison with these methods in term of computational complexity, for this comparison we use ResNet18~\cite{he2016deep} as our encoder. 
Here, $3$DDFA and DeFA use the linear $3$DMM model (BFM). Even though being trained with larger training corpus (DeFA) or having a cascade of CNNs iteratively refines the estimation ($3$DDFA),  these methods are still significantly outperformed by our nonlinear model~(Fig.~\ref{fig:face_align_CED}).
Meanwhile, $3$D-FAN and PRN achieve competitive performances by by-passing the linear $3$DMM model.
$3$D-FAN uses the heat map representation. PRN uses the position map representation which shares a similar spirit to our UV representation. 
Not only outperforms these methods in term of regressing landmark locations~(Fig.~\ref{fig:face_align_CED}), our model also directly provides head pose as well as the facial albedo and environment lighting information.

%
%

\SubSubSection{3D Face Reconstruction}

\Paragraph{Qualitative Comparisons}

We compare our approach to recent representative face reconstruction work: $3$DMM fitting networks learned in unsupervised (Tewari~\etal~\cite{tewari2017mofa, tewari2018self}) or supervised fashion (Sela~\etal~\cite{sela2017unrestricted}) and also a non-$3$DMM approach (Jackson~\etal~\cite{jackson2017large}).

\begin{figure}[t!]
\begin{center}
\small
\setlength{\tabcolsep}{3pt}
\begin{tabular}{c@{\hskip 1.5mm}c@{\hskip 0.5mm}c@{\hskip 1.5mm}c@{\hskip 0.5mm}c}
Input & \multicolumn{2}{c}{Our}  & \multicolumn{2}{c}{Tewari17} \\
\includegraphics[width=\ReconFigWid]{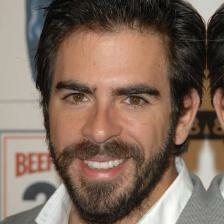} &
\includegraphics[width=\ReconFigWid]{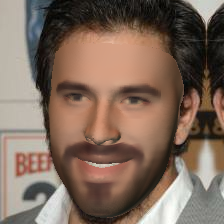} &
\includegraphics[width=\ReconFigWid]{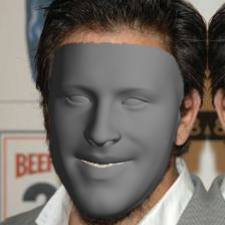} &
\includegraphics[width=\ReconFigWid]{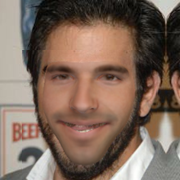} &
\includegraphics[width=\ReconFigWid]{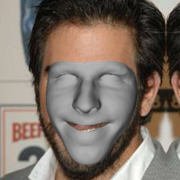}
\\
\includegraphics[width=\ReconFigWid]{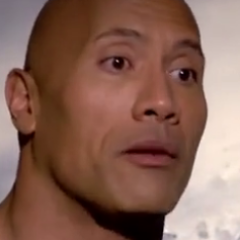} &
\includegraphics[width=\ReconFigWid]{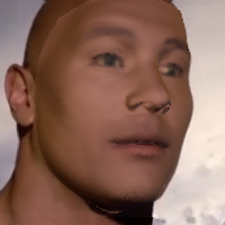} &
\includegraphics[width=\ReconFigWid]{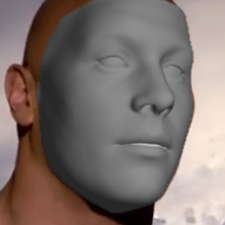} &
\includegraphics[width=\ReconFigWid]{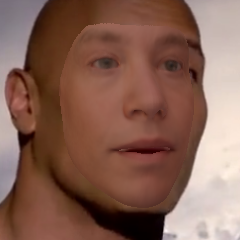} &
\includegraphics[width=\ReconFigWid]{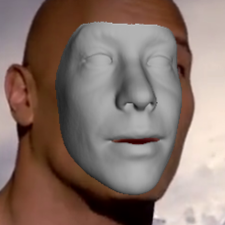}
\end{tabular}
\vspace{-2mm}
\caption{\small $3$D reconstruction results comparison to Tewari~\etal~\cite{tewari2017mofa}. Their reconstructed shapes suffer from the surface shrinkage when dealing with challenging texture or shape outside the linear model subspace. They can't handle large pose variation well either. Meanwhile, our nonlinear model is more robust to these variations.}
\label{fig:3drecon_tewari17}\figvspace
\vspace{-2mm}
\end{center}
\end{figure}

MoFA, the monocular reconstruction work by Tewari~\etal~\cite{tewari2017mofa}, is relevant to us as they also learn to fit $3$DMM in an unsupervised fashion. 
Even being trained on in-the-wild images, their method is still limited to the linear bases. 
Hence there reconstructions suffer the surface shrinkage when dealing with challenging texture, i.e., facial hair (Fig.~\ref{fig:3drecon_tewari17}). 
Our network faithfully models these in-the-wild texture, which leads to better $3$D shape reconstruction. 

\begin{figure}[t!]
\begin{center}
\small
\setlength{\tabcolsep}{3pt}
\begin{tabular}{c@{\hskip 1.5mm}c@{\hskip 0.5mm}c@{\hskip 1.5mm}c@{\hskip 0.5mm}c}
Input & \multicolumn{2}{c}{Our}  & \multicolumn{2}{c}{Tewari18} \\
\includegraphics[width=\ReconFigWid]{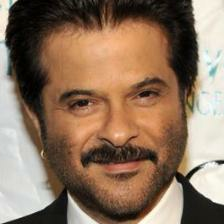} &
\includegraphics[width=\ReconFigWid]{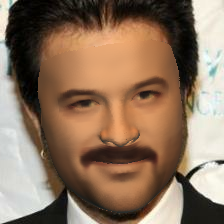} &
\includegraphics[width=\ReconFigWid]{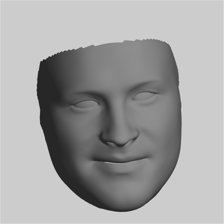} &
\includegraphics[width=\ReconFigWid]{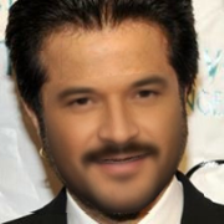} &
\includegraphics[width=\ReconFigWid]{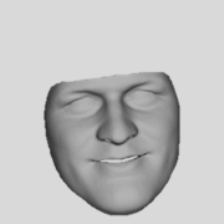}
\\
\includegraphics[width=\ReconFigWid]{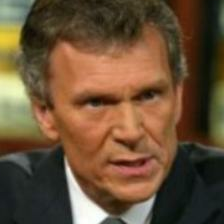} &
\includegraphics[width=\ReconFigWid]{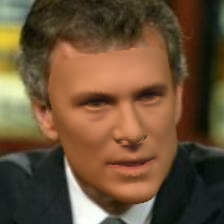} &
\includegraphics[width=\ReconFigWid]{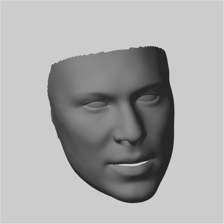} &
\includegraphics[width=\ReconFigWid]{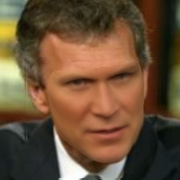} &
\includegraphics[width=\ReconFigWid]{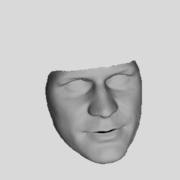}
\end{tabular}
\vspace{-2mm}
\caption{\small $3$D reconstruction comparison to Tewari~\etal~\cite{tewari2018self}. Our model better reconstructs inputs in both texture (facial hair direction in the top image) and shape (nasolabial folds in the bottom image).}
\label{fig:3drecon_tewari18}\figvspace
\vspace{-2mm}
\end{center}
\end{figure}

Concurrently, Tewari~\etal~\cite{tewari2018self} try to improve the linear $3$DMM representation power by learning a corrective space on top of a traditional linear model. Despite sharing similar spirit, our unified model exploits spatial relation between neighbor vertices and uses CNNs as shape/albedo decoders, which is more efficient than MLPs.
%
As a result, our reconstructions more closely match the input images in both texture and shape (Fig.~\ref{fig:3drecon_tewari18}).

\begin{figure*}[t!]
\begin{center}
\small
\setlength{\tabcolsep}{3pt}
\begin{tabular}{c@{\hskip 2mm}c@{\hskip 0.5mm}c@{\hskip 0.5mm}c@{\hskip 2mm}c@{\hskip 0.5mm}c@{\hskip 0.5mm}c@{\hskip 0.5mm}c@{}}
Input & \multicolumn{3}{c}{Our}  & \multicolumn{3}{c}{Sela17}\\
\includegraphics[width=\SelaReconFigWid]{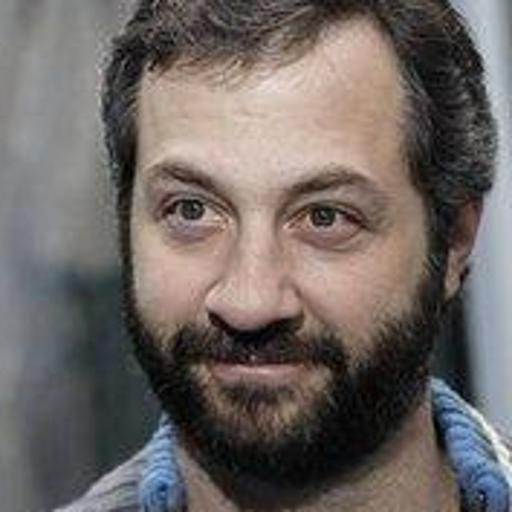} &
\includegraphics[width=\SelaReconFigWid]{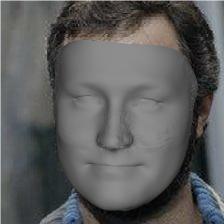} &
\includegraphics[width=\SelaReconFigWid]{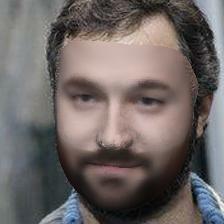} &
\includegraphics[width=\SelaReconFigWid]{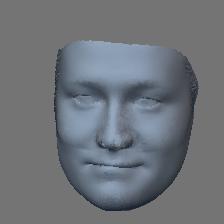} &
\includegraphics[width=\SelaReconFigWid]{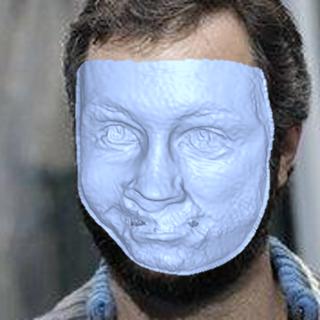} &
\includegraphics[width=\SelaReconFigWid]{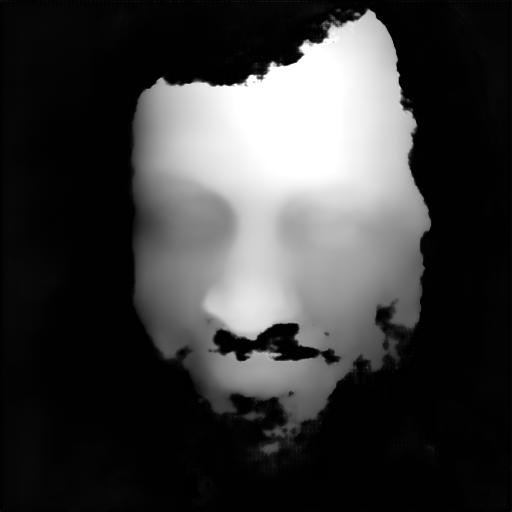} &
\includegraphics[width=\SelaReconFigWid]{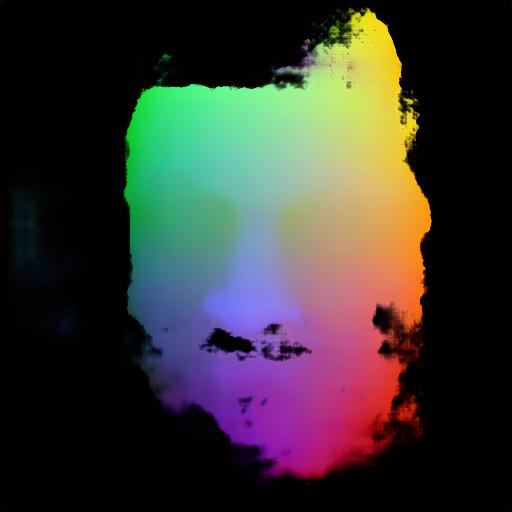}  \\
\includegraphics[width=\SelaReconFigWid]{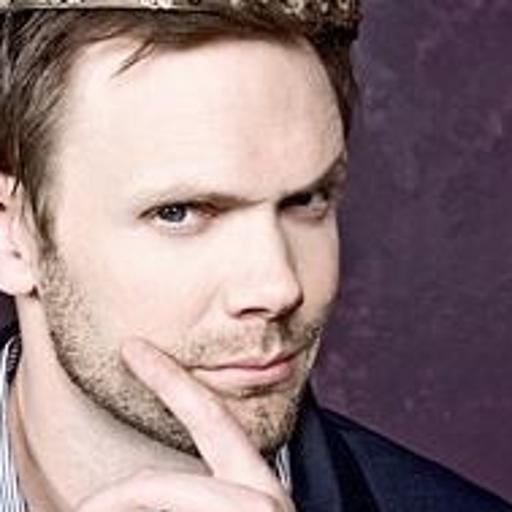} &
\includegraphics[width=\SelaReconFigWid]{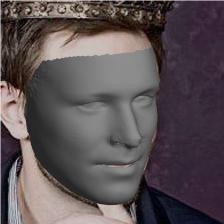} &
\includegraphics[width=\SelaReconFigWid]{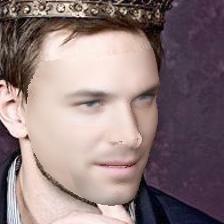} &
\includegraphics[width=\SelaReconFigWid]{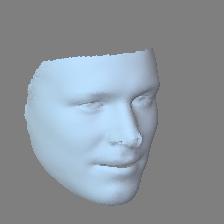} &
\includegraphics[width=\SelaReconFigWid]{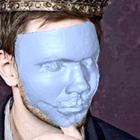} &
\includegraphics[width=\SelaReconFigWid]{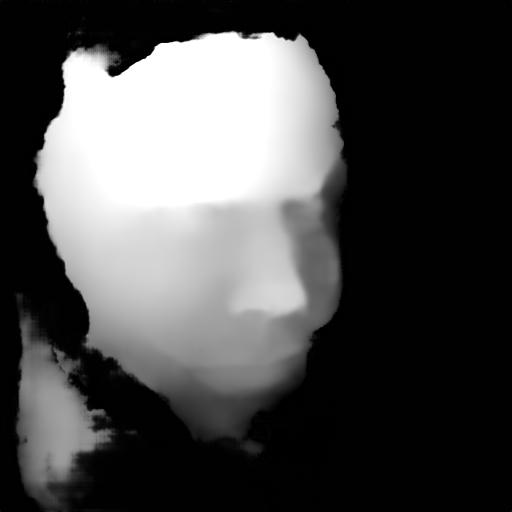} &
\includegraphics[width=\SelaReconFigWid]{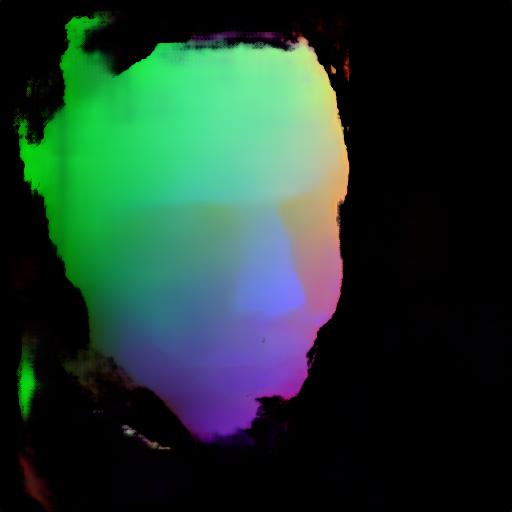}  \\
\end{tabular}
\vspace{-2mm}

\caption{\small $3$D reconstruction results comparison to Sela~\etal~\cite{richardson2017learning}. Besides showing the shape, we also show their estimated depth and correspondence map. Facial hair or occlusion can cause serious problems in their output maps.}
\label{fig:3drecon_sela}\figvspace
\vspace{-2mm}
\end{center}
\end{figure*}

The high-quality $3$D reconstruction work by Richardson~\etal\cite{richardson20163d, richardson2017learning}, Sela~\etal~\cite{sela2017unrestricted} obtain impressive results on adding fine-level details to the face shape when images are within the span of the used synthetic training corpus or the employed $3$DMM model. 
However, their performance significantly degrades when dealing with variations not in its training data span, e.g., facial hair. 
Our approach is not only robust to facial hair and make-up, but also automatically learns to reconstruct such variations based on the jointly learned model. 
We provide comparisons with them in Fig.~\ref{fig:3drecon_sela}, using the code provided by the author.

The current state-of-art 
method by Sela~\etal~\cite{sela2017unrestricted} consisting of three steps: an image-to-image network estimating a depth map and a correspondence map, non-rigid registration and a fine detail reconstruction. Their image-to-image network is trained on synthetic data generated by the linear model. Besides domain gap between synthetic and real images, this network faces a more serious problem of lacking facial hair in the low-dimension texture subspace of the linear model. This network's output tends to ignore these unexplainable region~(Fig.~\ref{fig:3drecon_sela}), which leads to failure in later steps.
Our network is more robust in handing these in-the-wild variations. 
Furthermore, our approach is orthogonal to  
Sela~\etal~\cite{sela2017unrestricted}'s fine detail reconstruction module or Richardson~\etal\cite{richardson2017learning}'s finenet. 
Employing these refinement on top of our fitting could lead to promising further improvement.

\begin{figure}[t!]
\begin{center}
\small
\setlength{\tabcolsep}{3pt}
\begin{tabular}{c@{\hskip 1.mm}c@{\hskip 0.5mm}c@{\hskip 1.5mm}c@{\hskip 1.mm}c@{\hskip 0.5mm}c@{\hskip 0.5mm}c@{}}
Input & Our  & VRN & Input & Our  & VRN \\

\includegraphics[width=\JsReconFigWid]{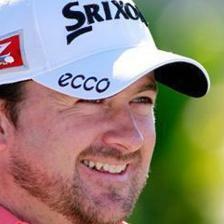} &
\includegraphics[width=\JsReconFigWid]{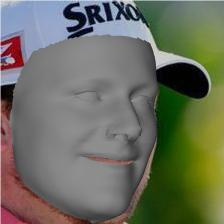} &
\includegraphics[trim=12 12 12 12, clip, width=\JsReconFigWid]{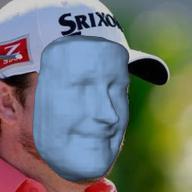} &

\includegraphics[width=\JsReconFigWid]{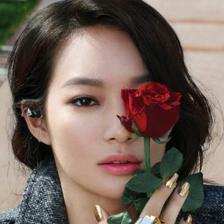} &
\includegraphics[width=\JsReconFigWid]{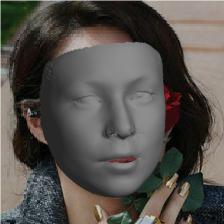} &
\includegraphics[trim=12 12 12 12, clip, width=\JsReconFigWid]{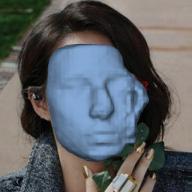} 
 \\
\includegraphics[width=\JsReconFigWid]{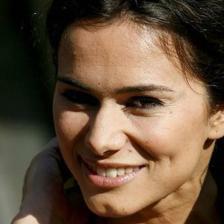} &
\includegraphics[width=\JsReconFigWid]{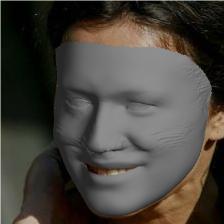} &
\includegraphics[trim=12 12 12 12, clip, width=\JsReconFigWid]{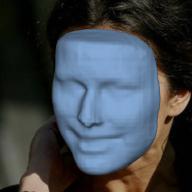} &

\includegraphics[width=\JsReconFigWid]{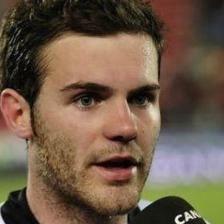} &
\includegraphics[width=\JsReconFigWid]{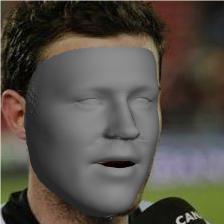} &
\includegraphics[trim=12 12 12 12, clip,width=\JsReconFigWid]{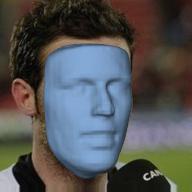}
\end{tabular}
\vspace{-2mm}
\caption{\small $3$D reconstruction results comparison to VRN by Jackson~\etal~\cite{jackson2017large} on CelebA. Volumetric shape representation results in non-smooth $3$D shape and loses shapes correspondence.}
\label{fig:3drecon_jackson}\figvspace
\end{center}
\end{figure}

We also compare our approach with a non-$3$DMM apporach VRN by Jackson~\etal~\cite{jackson2017large}. To avoid using low-dimension subspace of the linear $3$DMM, it directly regresses a $3$D shape volumetric representation via an encoder-decoder network with skip connection. This potentially helps the network to explore a larger solution space than the linear model, however with a cost of losing correspondence between facial meshes. 
Fig.~\ref{fig:3drecon_jackson} shows $3$D reconstruction visual comparison between VRN and ours. 
In general, VRN robustly handles in-the-wild texture variations. 
However, because of the volumetric shape representation, the surface is not smooth and is partially limited to present medium-level details as ours. 
Also, our model further provides projection matrix, lighting and albedo, which is applicable for more applications.

\begin{figure}[t!]
\centering
\includegraphics[trim=0 0 25 0,clip,width=\linewidth]{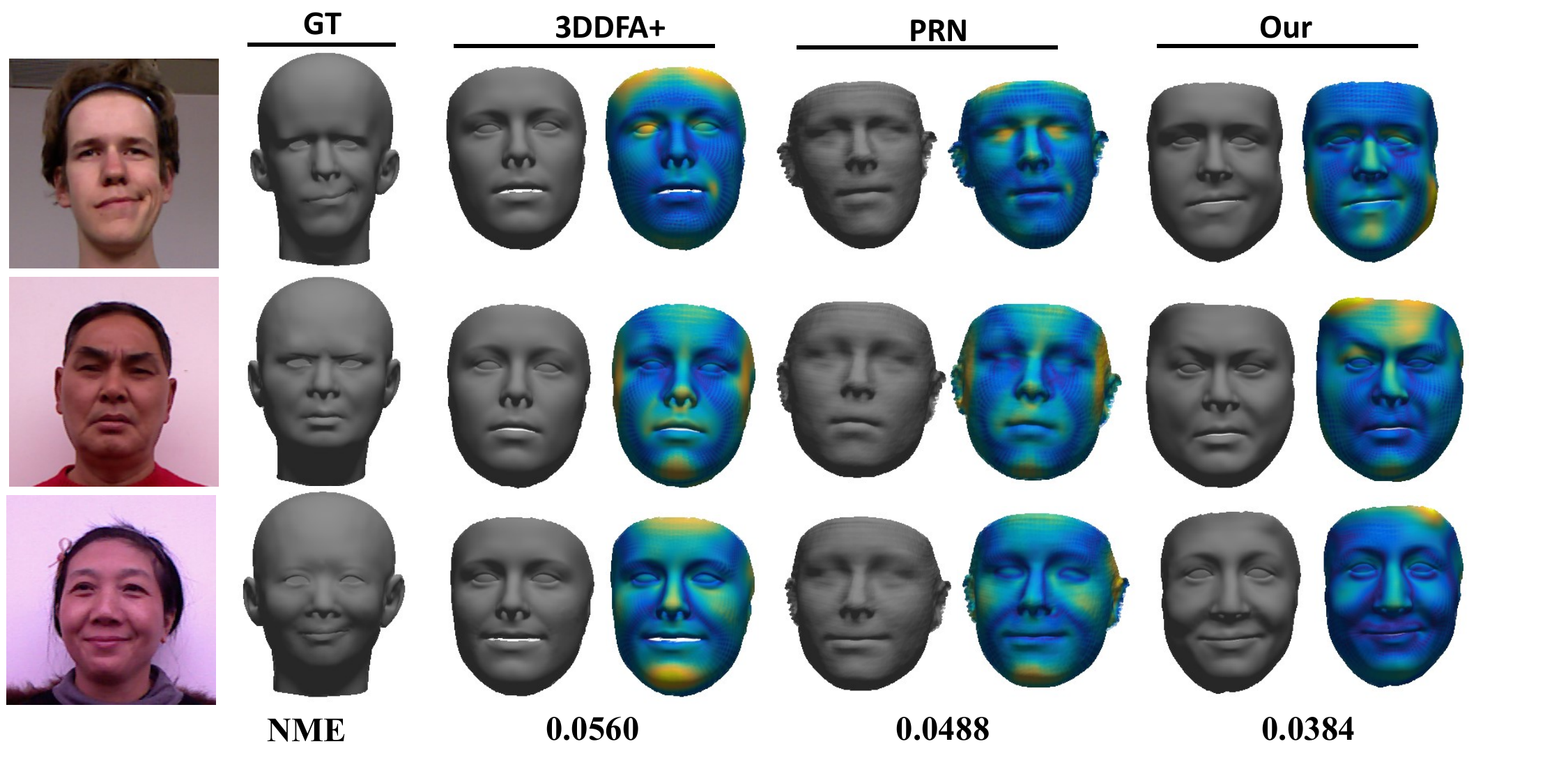}
\vspace{-5mm}
\caption{\small $3$D reconstruction quantitative evaluation on FaceWarehouse. We obtain a lower error compared to PRN~\cite{feng2018joint} and ~3DDFA+~\cite{zhu2017face}.
}
\label{fig:3d_rescon_quan}
\figvspace 
\end{figure}

\Paragraph{Quantitative Comparisons}

To quantitatively compare our method with prior works, we evaluate monocular $3$D reconstruction performance on FaceWarehouse~\cite{cao2014facewarehouse} and Florence dataset~\cite{bagdanov2011florence}, in which groundtruth 3D shape is available. Due to the diffrence in mesh topology,  ICP~\cite{amberg2007optimal} is used to establish correspondence between estimated shapes and ground truth point clouds. Similar to previous experiments, NME (averaged per-vertex errors normalized by inter-ocular distances) is used as the comparison metric.

\textbf{FaceWarehouse.} We compare our method with prior works with available pretrained models on all $19$ expressions of $150$ subjects of FaceWarehouse database~\cite{cao2014facewarehouse}. Visual and quantitative comparisons are shown in Fig.~\ref{fig:3d_rescon_quan}. 
Our model can faithfully resemble the input expression and significantly surpass all other regression methods (PRN~\cite{feng2018joint} and ~3DDFA+~\cite{zhu2017face}) in term of dense face alignment.


\begin{figure}[t!]
\begin{center}
\small
\begin{tabular}{@{}cc@{\hskip 1mm}c@{}}
\includegraphics[trim=110 250 130 240, clip, width=0.235\textwidth]{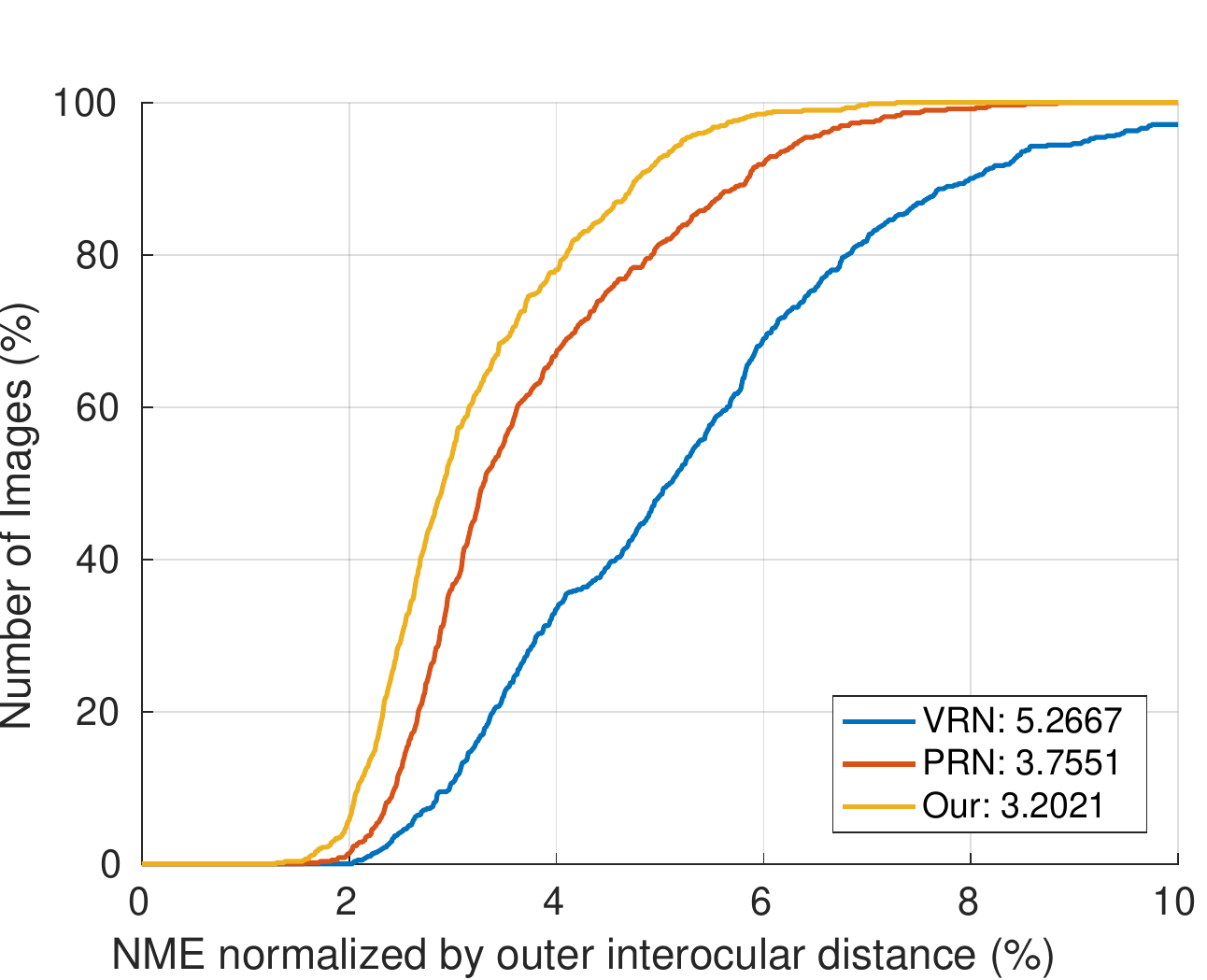} &
\includegraphics[trim=110 235 130 240, clip,width=0.215\textwidth]{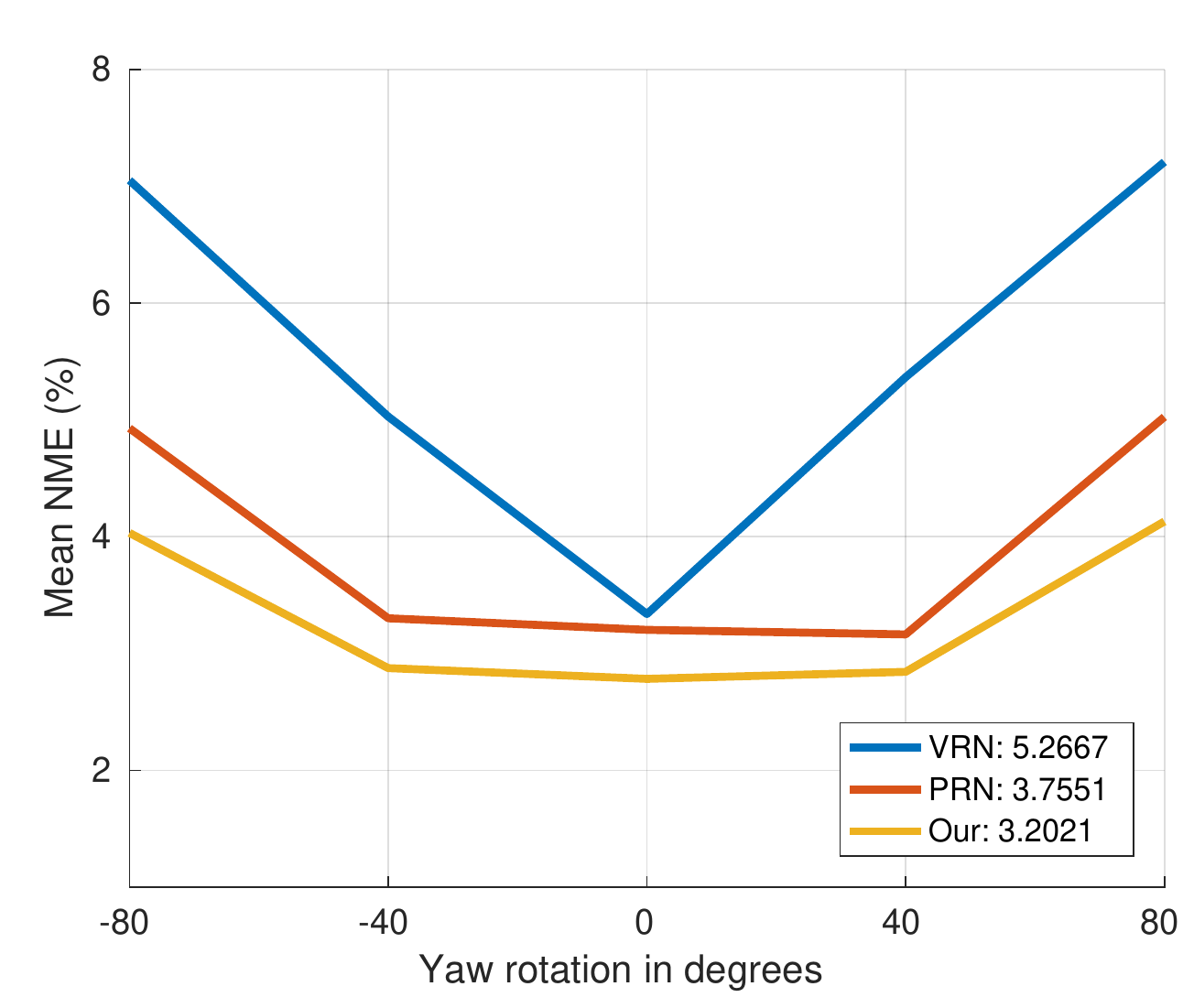} \\
(a) CED Curves & (b) Pose-specific NME
\end{tabular}
\vspace{-2mm}
\caption{\small 3D face reconstruction results on the Florence dataset~\cite{bagdanov2011florence}. The NME of each method is showed in the legend}
\label{fig:florence}\figvspace \vspace{-1mm}
\end{center}
\end{figure}

\textbf{Florence.} Using the experimental setting proposed in~\cite{jackson2017large}, we also quantitatively compared our approach with state-of-the-art methods (\eg VRN \cite{jackson2017large} and PRN~\cite{feng2018joint}) on the Florence dataset~\cite{bagdanov2011florence}. 
Each subject is rendered with multiple poses: pitch rotations of $-15^{\circ}$, $20^{\circ}$ and $25^{\circ}$ and raw rotations between $-80^{\circ}$ and $80^{\circ}$. 
Our model consistently outperforms other methods across different view angles (Fig.~\ref{fig:florence}).

\SubSubSection{Face editing}
Decomposing face image into individual components give us ability to edit the face by manipulating any component. 
Here we show two examples of face editing using our model.

\begin{figure}[t!]
\begin{center}
\small
\setlength{\tabcolsep}{3pt}
\begin{tabular}{ @{\hskip .5mm}c@{\hskip 1.5mm}c@{\hskip .5mm}c@{\hskip .5mm}c@{\hskip .5mm}c@{}c@{\hskip .5mm}}
\includegraphics[trim=5 5 10 5, clip, width=\EditingLightFigWid]{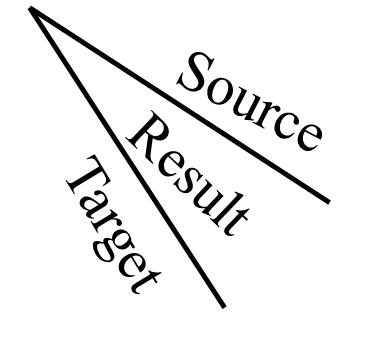} &
\includegraphics[width=\EditingLightFigWid]{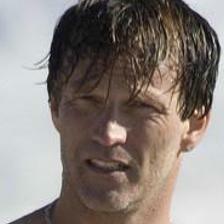} &
\includegraphics[width=\EditingLightFigWid]{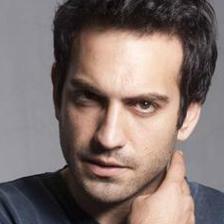} &
\includegraphics[width=\EditingLightFigWid]{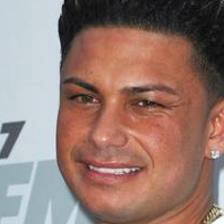} &
\includegraphics[width=\EditingLightFigWid]{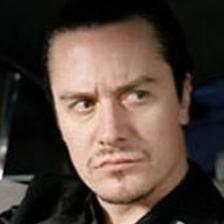} &
\\ 

\includegraphics[width=\EditingLightFigWid]{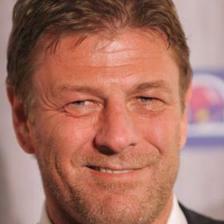} &
\includegraphics[width=\EditingLightFigWid]{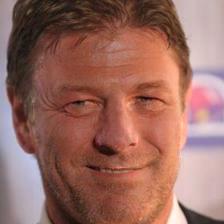} &
\includegraphics[width=\EditingLightFigWid]{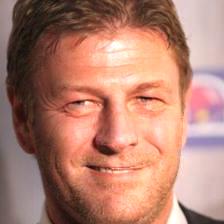} &
\includegraphics[width=\EditingLightFigWid]{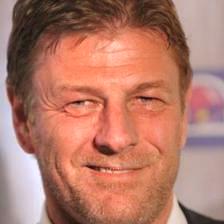} &
\includegraphics[width=\EditingLightFigWid]{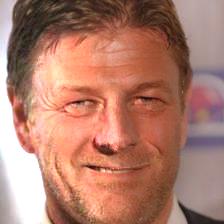} 
\\
\includegraphics[width=\EditingLightFigWid]{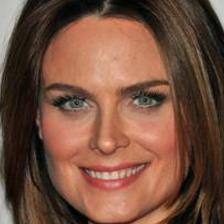} &
\includegraphics[width=\EditingLightFigWid]{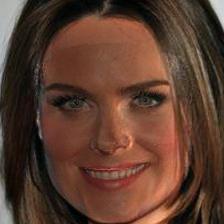} &
\includegraphics[width=\EditingLightFigWid]{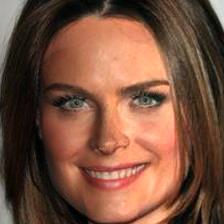} &
\includegraphics[width=\EditingLightFigWid]{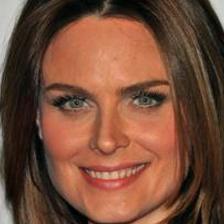} &
\includegraphics[width=\EditingLightFigWid]{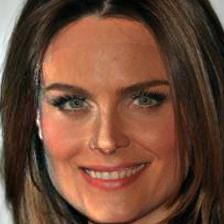} 
\\
\includegraphics[width=\EditingLightFigWid]{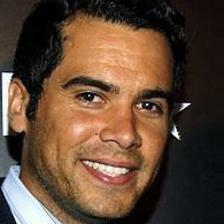} &
\includegraphics[width=\EditingLightFigWid]{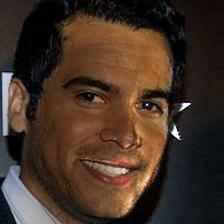} &
\includegraphics[width=\EditingLightFigWid]{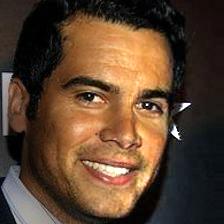} &
\includegraphics[width=\EditingLightFigWid]{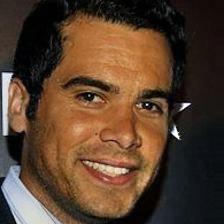} &
\includegraphics[width=\EditingLightFigWid]{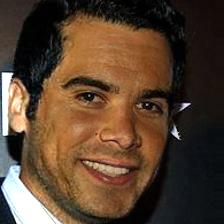} 
\\
\includegraphics[width=\EditingLightFigWid]{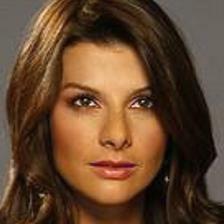} &
\includegraphics[width=\EditingLightFigWid]{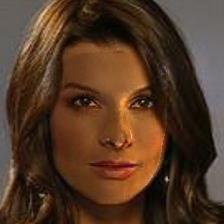} &
\includegraphics[width=\EditingLightFigWid]{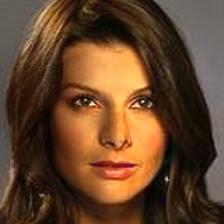} &
\includegraphics[width=\EditingLightFigWid]{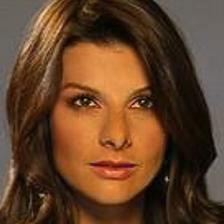} &
\includegraphics[width=\EditingLightFigWid]{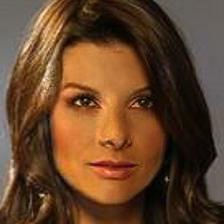} 
\\
\hdashline
\includegraphics[trim=0 0 0 -15, clip, width=\EditingLightFigWid]{img/Editing/Lighting/img_10_i.jpg} &
\includegraphics[width=\EditingLightFigWid]{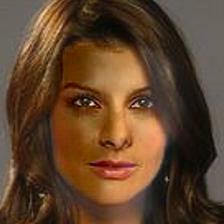} &
\includegraphics[width=\EditingLightFigWid]{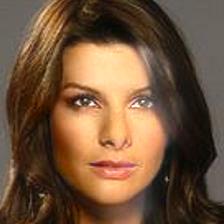} &
\includegraphics[width=\EditingLightFigWid]{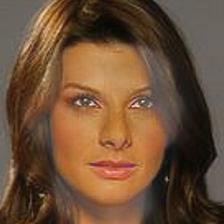} &
\includegraphics[width=\EditingLightFigWid]{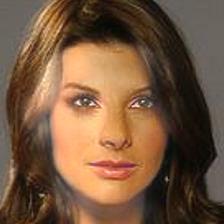} 
\end{tabular}
\vspace{-2mm}
\caption{\small Lighting transfer results. We transfer the lighting of source images (first row) to target images (first column).
We have similar performance compare to the state-of-the-art method of Shu~\etal~\cite{shu2018portrait} despite  being orders of magnitude faster ($150$ ms vs. $3$ min per image). }
\label{fig:editing_lighting}\figvspace \vspace{-2mm}
\end{center}
\end{figure}

\Paragraph{Relighting}
First we show an application to replacing the lighting of a target face image using lighting from a source face (Fig.~\ref{fig:editing_lighting}). 
After estimating the lighting parameters $\mathbf{L}_{\text{source}}$ of the source image, we render the transfer shading using  the target shape $\mathbf{S}_{\text{target}}$ and the source lighting $\mathbf{L}_{\text{source}}$. 
This transfer shading can be used to replace the original source shading. 
Alternatively, value of $\mathbf{L}_{\text{source}}$ can be arbitrarily chosen based on the SH lighting model, without the need of source images. 
Also, here we use the original texture instead of the output of our decoder to maintain image details. 

\begin{figure}[t!]
\begin{center}
\small
\setlength{\tabcolsep}{3pt}
\begin{tabular}{ @{}c@{\hskip 2.5mm}c@{\hskip .5mm}c@{\hskip .5mm}c@{\hskip .5mm}c@{}c@{}}
\includegraphics[width=\EditingFigWid]{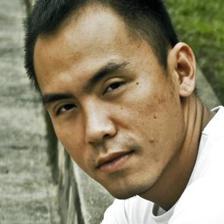} & 
\includegraphics[width=\EditingFigWid]{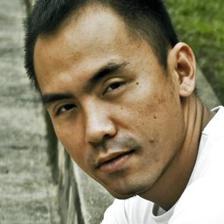} &
\includegraphics[width=\EditingFigWid]{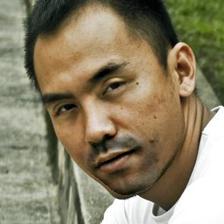} &
\includegraphics[width=\EditingFigWid]{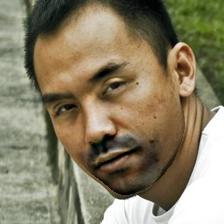} \\

\includegraphics[width=\EditingFigWid]{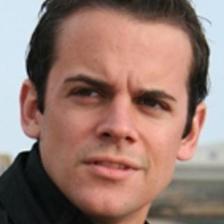} &
\includegraphics[width=\EditingFigWid]{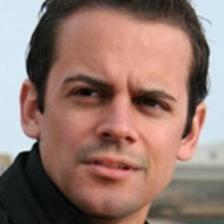} &
\includegraphics[width=\EditingFigWid]{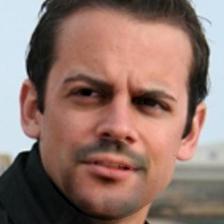} &
\includegraphics[width=\EditingFigWid]{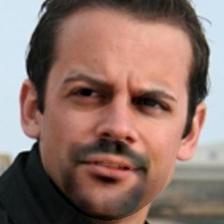} \\

\includegraphics[width=\EditingFigWid]{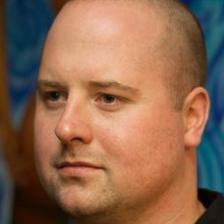} &
\includegraphics[width=\EditingFigWid]{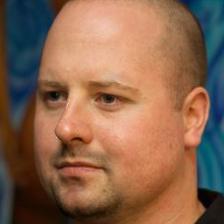} &
\includegraphics[width=\EditingFigWid]{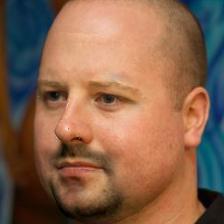} &
\includegraphics[width=\EditingFigWid]{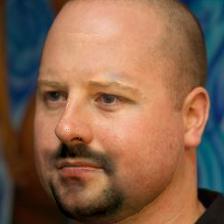} \\

\includegraphics[width=\EditingFigWid]{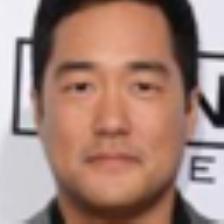} &
\includegraphics[width=\EditingFigWid]{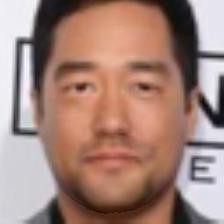} &
\includegraphics[width=\EditingFigWid]{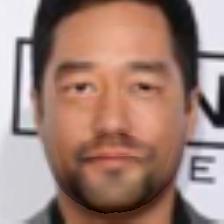} &
\includegraphics[width=\EditingFigWid]{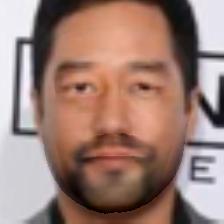} \\

\hdashline

\includegraphics[trim=0 0 0 -15,clip, width=\EditingFigWid]{img/Editing/Mustache/Shu/pred_img_b_00_i_02_in.jpg} &
\includegraphics[width=\EditingFigWid]{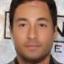} &
\includegraphics[width=\EditingFigWid]{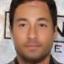} &
\includegraphics[width=\EditingFigWid]{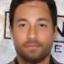} \\
\end{tabular}
\vspace{-2mm}
\caption{\small Growing mustache editing results. The first collumn shows original images, the following collumns show edited images with increasing magnitudes. Comparing to Shu~\etal~\cite{shu2017neural} results (last row), our edited images are more realistic and identity preserved.}
\label{fig:editing}\figvspace
\end{center}
\end{figure}

\Paragraph{Attribute Manipulation}
Given faces fitted by $3$DMM model, we can edit images by naive modifying one or more elements in the albedo or shape representation. 
More interestingly, we can even manipulate the semantic attribute, such as growing beard, smiling, etc. 
The approach is similar to learning attribute embedding in Sec.~\ref{sec:expressiveness}. Assuming, we would like to edit appearance only. 
For a given attribute, e.g., beard, we feed two sets of images with and without that attribute $\{\mathbf{I}^p_i\}_{i=1}^n$ and $\{\mathbf{I}^n_i\}_{i=1}^n$ into our encoder to obtain two average parameters $\mathbf{f}_A^p$ and $\mathbf{f}_A^n$. 
Their difference $\mathbf{\Delta f}_A=\mathbf{f}_A^p - \mathbf{f}_A^n$ is the direction to move from the distribution of negative images to positive ones.
By adding $\mathbf{\Delta f}_A$ with different magnitudes, we can generate modified images with different degree of changes.
To achieve high-quality editing with identity-preserved, the final editing result is obtained by adding the residual, the different between the modified image and our reconstruction, to the original input image.
This is a critical difference to Shu~\etal~\cite{shu2017neural} to improve results quality (Fig.~\ref{fig:editing}).

\section{Conclusions}

Since its debut in 1999, $3$DMM has became a cornerstone of facial analysis research 
with applications to many problems.
Despite its impact, it has drawbacks in requiring training data of $3$D scans, learning from controlled $2$D images, and limited representation power due to linear bases for both shape and texture.
These drawbacks could be formidable when fitting $3$DMM to unconstrained faces, or learning $3$DMM for generic objects such as shoes.
This paper demonstrates that there exists an alternative approach to $3$DMM learning, where a nonlinear $3$DMM can be learned from a large set of in-the-wild face images without collecting $3$D face scans.
Further, the model fitting algorithm can be learnt jointly with $3$DMM, in an end-to-end fashion.

Our experiments cover a diverse aspects of our learnt model, some of which might need the subjective judgment of the readers.
We hope that both the judgment and quantitative results could be viewed under the context that, unlike linear $3$DMM, no genuine $3$D scans are used in our learning.
Finally, we believe that unsupervisedly or weak-supervisedly learning $3$D models from large-scale in-the-wild $2$D images is one promising research direction. 
This work is one step along this direction.


%

%

\ifCLASSOPTIONcompsoc
\else
\fi


\ifCLASSOPTIONcaptionsoff
  \newpage
\fi



\bibliographystyle{IEEEtran}
\bibliography{nonlinear_3dmm}  
%

%

%
\vspace{-7mm}
\begin{IEEEbiography}
[{\includegraphics[width=1in,height=1.2in,clip,keepaspectratio]{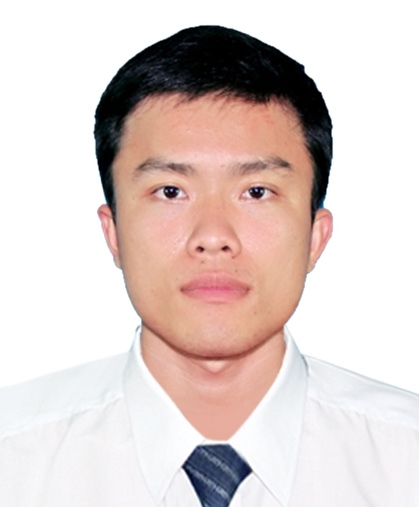}}]
{Luan Tran}
received his B.S. in Computer Science from Michigan State University
with High Hornors in $2015$. He is now pursuing his Ph.D. also at Michigan State University in the area of deep learning and computer vision. His research areas of interest include deep learning and computer vision, in particular, face modeling and face recognition. He is a member of the IEEE.
\end{IEEEbiography}

\vspace{-7mm}
\begin{IEEEbiography}[{\includegraphics[width=1in,height=1.2in,clip,keepaspectratio]{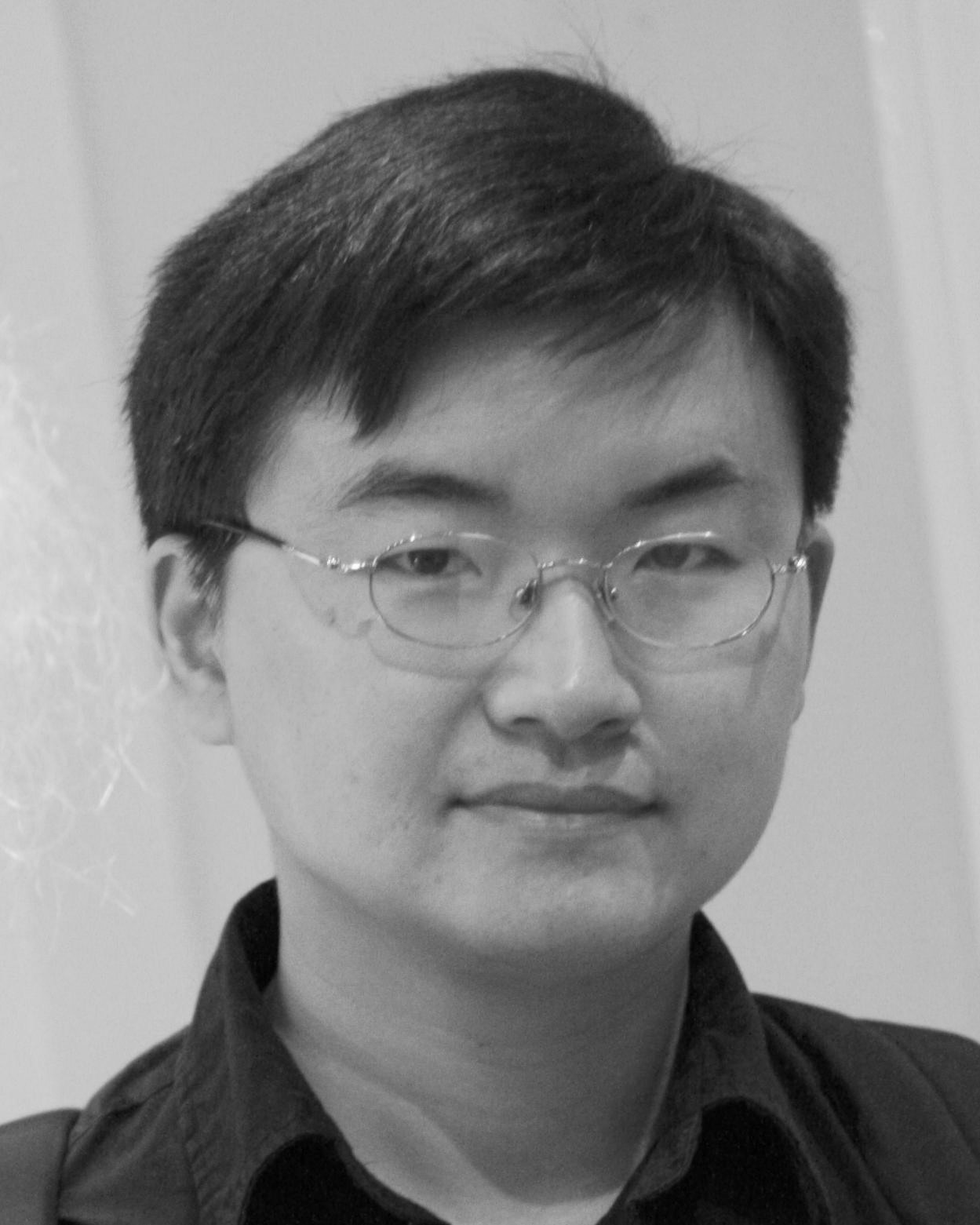}}]{Xiaoming Liu}is an Associate Professor at the Department of Computer Science and Engineering of Michigan State University. He received the Ph.D. degree in Electrical and Computer Engineering from Carnegie Mellon University in $2004$. Before joining MSU in Fall $2012$, he was a research scientist at General Electric (GE) Global Research. His research interests include computer vision, machine learning, and biometrics. As a co-author, he is a recipient of Best Industry Related Paper Award runner-up at ICPR $2014$, Best Student Paper Award at WACV $2012$ and $2014$, Best Poster Award at BMVC $2015$, and Michigan State University College of Engineering Withrow Endowed Distinguished Scholar Award. He has been the Area Chair for numerous conferences, including FG, ICPR, WACV, ICIP, ICCV, and CVPR. He is the program co-chair of WACV $2018$ and BTAS $2018$. He is an Associate Editor of Neurocomputing, Pattern Recognition Letters, and Pattern Recognition. He has authored more than $100$ scientific publications, and has filed $26$ U.S. patents. 
\end{IEEEbiography}




\end{document}